\RequirePackage{fix-cm}
\documentclass[smallcondensed]{svjour3}     
\smartqed  
\usepackage{graphicx}
\usepackage{url}
\usepackage{graphics}
\usepackage{amsmath}
\usepackage{amssymb}
\usepackage{microtype}
\usepackage{booktabs}

\usepackage{algorithm}
\usepackage{float}
\usepackage{natbib}
\usepackage{algorithmic}
\usepackage{subcaption}
\newtheorem{assumption}{Assumption}
\begin{document}

\title{Variance Reduction on General Adaptive Stochastic Mirror Descent}


\author{Wenjie Li \thanks{Correspondence to Wenjie Li}  \and
        Zhanyu Wang \and
        Yichen Zhang \and
        Guang Cheng 
}


\institute{Wenjie Li\at
              \email{li3549@purdue.edu}           
                Department of Statistics, Purdue University
           \and
           Zhanyu Wang\at
             \email{wang4094@purdue.edu}  
                Department of Statistics, Purdue University
           \and
           Yichen Zhang \at
              \email{zhang@purdue.edu}
              Krannert School of Management, Purdue University
           \and
           Guang Cheng \at
              \email{chengg@purdue.edu}
                Department of Statistics, Purdue University
}

\date{Received: date / Accepted: date}

\maketitle

\begin{abstract}
In this work, we investigate the idea of variance reduction by studying its properties with general adaptive mirror descent algorithms in the nonsmooth nonconvex finite-sum optimization problems. We propose a simple yet generalized framework for variance reduced adaptive mirror descent algorithms named SVRAMD and provide its convergence analysis in both the nonsmooth nonconvex problem and the P-L conditioned problem. We prove that variance reduction reduces the SFO complexity of adaptive mirror descent algorithms and thus accelerates their convergence. In particular, our general theory implies that variance reduction can be applied to algorithms using time-varying step sizes and self-adaptive algorithms such as AdaGrad and RMSProp. Moreover, the convergence rates of SVRAMD recover the best existing rates of non-adaptive variance reduced mirror descent algorithms without complicated algorithmic components. Extensive experiments in deep learning validate our theoretical findings.
\end{abstract}
\section{Introduction}
\label{sec: introduction}
In this work, we study the nonsmooth nonconvex finite sum problem
\begin{equation}
    \begin{aligned}
    \text{min}_{x \in \mathcal{X}} F(x) := f(x) + h(x) \nonumber
    \end{aligned}
\end{equation}
where $f(x) = \frac{1}{n} \sum_{i=1}^n f_i(x)$ and each $f_i$ is a smooth but possibly nonconvex function, and $h(x)$ is a nonsmooth but convex function, for example, the $L_1$ regularization. Recently, the smooth version of the problem has been thoroughly studied, i.e., when $h(x) = 0$. Since it is difficult to determine the global minimum of a nonconvex function, the convergence analyses of different algorithms have focused on the gradient complexity of finding the first order stationary point of the loss $F(x)$, i.e., $\|\nabla F(x) \|_2^2 \leq \epsilon$. To reduce the gradient complexity of gradient descent and stochastic gradient descent (SGD), the famous Stochastic Variance Reduced Gradient method (SVRG) \citep{Johnson2013Accelerating} and its popular variants have been proposed, such as SAGA \citep{Defazio2014SAGA}, SCSG \citep{lei2017Nonconvex}, SNVRG \citep{Zhou2018Stochastic}, SPIDER \citep{Cong2018SPIDER}, stablized SVRG \citep{ge2019stabilized}, and Natasha momentum variants \citep{allenzhu2017natasha,allenzhu2017natasha2}. These algorithms are proven to accelerate the convergence of SGD substantially. 

When it comes to the nonsmooth case, a few algorithms based on the mirror descent algorithm \citep{Beck2003Mirror, Duchi2010Composite} have been studied recently. For example, \citet{ghadimi2016Minibatch} provided the convergence rate of Proximal Gradient Descent (ProxGD), Proximal SGD(ProxSGD), and Stochastic Mirror Descent (SMD) when the sample size was sufficiently large. The gradient complexity of these algorithms were shown to match the original algorithms without the proximal operation \citep{ghadimi2016Minibatch}. \citet{Reddi2016Proximal} proposed ProxSVRG and ProxSAGA and analyzed their convergence rates, which were the proximal variants of SVRG and SAGA respectively. \citet{li2018Simple} proposed ProxSVRG+ and obtained even faster convergence than ProxSVRG with favorable constant minibatch sizes. 

However, all the above extensions of SGD and SVRG, both for the smooth and the nonsmooth optimization problems, do not consider the case when the algorithm becomes adaptive, for example, when the step sizes are not fixed or even when the proximal functions in mirror descent are not fixed. Adaptive step sizes, such as linear decay, warm up \citep{goyal2017accurate}, and restart \citep{loshchilov2016sgdr} are frequently used in the training of neural networks in practice, but recent papers of variance reduced algorithms only use constant step sizes in their analyses \citep{li2018Simple, Zhou2018Stochastic}. Moreover, adaptive algorithms such as AdaGrad \citep{Duchi2011Adaptive}, Adam \citep{Kingma2015Adam} and AMSGrad \citep{Reddi2018On} have become popular in application, but their proximal functions change with time. No theoretical guarantees of applying variance reduction to such adaptive algorithms have been shown before.

Therefore, instead of trying to create even faster algorithms in the nonsmooth setting with special components, this work addresses a more important and  interesting question: \textit{Can the simplest variance reduction technique accelerate the convergence of all or most of the adaptive stochastic mirror descent algorithms?} We give an affirmative answer to this question, with the only additional mild requirement that there is a lower bound for the strong convexity of the proximal functions in the mirror descent algorithms.

In particular, we highlight the following contributions:

\begin{enumerate}
    \item We propose a simple and general variance reduction algorithm framework for most adaptive mirror descent algorithms named SVRAMD. We prove that in both the nonsmooth nonconvex problems and the gradient dominated problems (P-L condition, see Section \ref{subsec: SVRAMD_convergence_PL}), the SVRAMD algorithm converges faster than the original adaptive mirror descent algorithms with a mild assumption on the convexity of the proximal functions.
    
    \item Our general theory implies many interesting results. For example, we claim that time-varying step sizes are allowed for ProxSVRG+ (and many other variance reduced algorithms). As long as the step sizes are upper bounded by the inverse of the smoothness constant $1/L$, ProxSVRG+ still converges faster than ProxSGD under the same step size schedule. Also, we claim that variance reduction can work well with self-adaptive algorithms, such as AdaGrad \citep{Duchi2011Adaptive} and RMSProp  \citep{Tieleman2012RMSProp} (zero momentum version of Adam) and make their convergence faster. Moreover, our choices of the hyper-parameters in the theorem provide a general intuition that larger batch sizes are needed when using variance reduction on adaptive mirror descent algorithms with weaker convexity, which can be helpful when tuning the batch sizes in practice.
    \item We examine the correctness of our claims thoroughly using popular deep learning models on the MNIST \citep{MNIST}, the CIFAR-10 and the CIFAR-100 \citep{Cifar} datasets. All the experimental results are consistent with our theoretical findings.
\end{enumerate}

\section{Additional Related Work}
\label{sec: related_work}
Due to the huge amount of work in optimization algorithms and their analyses, we are not able to review all of them in this paper. Instead, we choose to review the two additional lines of research that are related to this paper.

\subsection{Self-adaptive Algorithms and Their Analysis}

Since the creation of AdaGrad \citep{Duchi2011Adaptive}, self-adaptive algorithms have become popular choices for optimization, especially in deep learning topics such as  Natural Language Processing (NLP) and training generative adversarial models \citep{ghadimi2016Minibatch}. \citet{Kingma2015Adam} proposed Adam that combined momentum with RMSProp \citep{Tieleman2012RMSProp} to further improve the convergence speed of AdaGrad, but \citet{Reddi2016Proximal} proved that Adam could diverge even in convex problems. \citet{wilson2017marginal} also showed that the generalization performance of adaptive algorithms was even worse than SGD. Therefore, several complicated variants of Adam have been proposed recently, such as AdaBound \citep{Luo2019Adaptive}, NosAdam \citep{Huang2019Nostalgic}, Radam \citep{liu2019on}, AdaBelief \citep{zhuang2020adabelief} and AdaX \citep{Li2020AdaX} to further improve the convergence and the generalization performance of Adam. However, all the aforementioned papers only proved their fast convergence in convex problems and with sparse gradients. Recently, \citet{Chen2019On} and \citet{zhou2020convergence} proved the convergence rates of some self-adaptive algorithms (AMSGrad, AdaGrad) in the nonconvex smooth problem, but their convergence rates are still the same as that of SGD.

\subsection{Combining Variance Reduction with Mirror Descent Algorithms}

Combining variance reduction with mirror descent algorithms has become another heated topic. Recently, \citet{lei2019adaptivity} further extended SCSG to the general mirror descent form with some additional assumptions on the (fixed) proximal function, but they only considered a convex problem setting. \citet{liu2020adam} proposed Adam+ and claimed that variance reduction could improve the convergence rate of a special variant of Adam. \citet{duboistaine2021svrg} proposed to use AdaGrad in the inner loop of SVRG to make it robust to different step size choices.

\section{Problem Setup and Notations}
\label{sec: prelim}

In this section, we present the preliminary notations and the concepts used for the convergence analysis throughout this paper.
 
\subsection{Notations}
We first present the notations we use. For two matrices $A, B \in \mathbb{R}^{d \times d}$, we use $A\succeq B$ to denote that the matrix $A - B$ is positive semi-definite. For two real numbers $a,b \in \mathbb{R}$, we use $a \wedge b, a \vee b$ as short-hands for $\text{min}(a, b)$ and $\text{max}(a, b)$. We use $\lfloor a \rfloor$ to denote the largest integer that is smaller than $a$. We use $\widetilde{O}(\cdot)$ to hide logarithm factors in big-$O$ notations. Moreover, for an integer $n\in \mathbb{N}$, we frequently use the notation $[n]$ to represent the set $\{1,2,\cdots, n\}$. For  the loss function $F(x)$, we denote the global minimum value of $F(x)$ to be $F^*$, and define $\Delta_F = F (x_1) - F^*$, where $x_1$ is the initialization point of the algorithm. 

\subsection{Assumptions and Definitions for Nonconvex Nonsmooth Analysis}
We recall the update rule of the general stochastic mirror descent (SMD) algorithm with  \textit{adaptive proximal functions} $\psi_{t}(x)$ as follows
\begin{equation}
\label{eqn: composite_mirror_descent}
    x_{t+1} = \text{argmin}_x  \left \{\alpha_t \langle g_{t}, x\rangle + \alpha_t h(x)+ B_{\psi_t}(x, x_{t}) \right \}
\end{equation}
where $\alpha_t$ is the step size, $g_t = \nabla f_{\mathcal{I}_j} (x_t)$ is the gradient from a random data batch $\mathcal{I}_j$, and $h(x)$ is the regularization function on the dual space. $B_{\psi_t}(x, x_t)$ is the Bregman divergence with respect to the proximal function $\psi_t(x)$, defined as 
\begin{equation*}
B_{\psi_t}(x, y) = \psi_t(x) - \psi_t(y) - \langle \nabla \psi_t(y), x- y 
\rangle\end{equation*}

Different $\psi_t(x)$  would generate different Bregman divergences. For example, the update rule for ProxSGD is
\begin{equation*}
\nonumber
    x_{t+1} = \text{argmin}_x \left \{\alpha_t \langle g_{t}, x\rangle + \alpha_t h(x)+ \frac{1}{2}\|x-x_t\|_2^2)  \right \}
\end{equation*}
In this case, $B_{\psi_t}(x, y) = \frac{1}{2}\|x-y\|_2^2$  and it is generated by the proximal function $\psi_t(x) = \frac{1}{2}\|x\|_2^2$. A very important property for Bregman divergence is that $\psi_t(x)$ is $m$-strongly convex if and only if $B_{\psi_t}(x, y) \geq \frac{m}{2}\|y-x\|_2^2$. In this work, we consider general adaptive mirror descent algorithms whose proximal functions are undetermined and can vary with respect to time. However, we require that they are all $m$-strongly convex for some real constant $m > 0$ ({Assumption \textbf{\ref{assumption: 1}}}) , 

\begin{assumption}
\label{assumption: 1}
The proximal functions $\psi_{t}(x)$ are all $m$-strongly convex with respect to $\|\cdot\|_2$, i.e.,
\begin{equation}
    \begin{aligned}
    \label{def: strongly_convex}
        \psi_{t}(y) \geq \psi_{t}(x) + \langle \nabla \psi_{t}(x), y-x\rangle + \frac{m}{2}\|y-x\|_2^2, \forall t > 0 \nonumber
        \end{aligned}
\end{equation}
\end{assumption}

The constant $m$ can be viewed as a lower bound of the strong convexity of all the proximal functions $\{\psi_{t}(x)\}_{t=1}^T$, where $T$ is the total number of iterations, and therefore Assumption \textbf{\ref{assumption: 1}} is mild. Here we provide a few examples of different proximal functions that satisfy Assumption \textbf{\ref{assumption: 1}} and the corresponding lower bound to show the cases where our general theory can be applied.

\begin{example}
\label{example: 1}
$\psi_{t}(x) = \phi_{t}(x) + \frac{c}{2}\|x\|_2^2, c > 0$,  where each $\phi_{t}(x)$ is an arbitrary convex differentiable function, then $m = c$. Note that this case also reduces to the ProxSGD algorithm when $\phi_t(x) = 0$ and $c = 1$.
\end{example}

\begin{example}
\label{example: 2}
$\psi_{t}(x) = \frac{c_t}{2}\|x\|_2^2$, where $c_t \geq c > 0, \forall t \in [T]$, then $m = c$. This case is equivalent to using time-varying step sizes in ProxSGD even when $\alpha_t$ is fixed, as we can divide all terms in Eqn. (\ref{eqn: composite_mirror_descent}) by $c_t$ simultaneously. In other words, we only require the time-varying step sizes ($\alpha_t /c_t$) to be upper bounded to satisfy Assumption \textbf{\ref{assumption: 1}}.
\end{example}

\begin{example}
\label{example: 3}
$\psi_{t}(x) = \frac{1}{2}\langle x, H_{t} x \rangle$, where $H_t \in \mathbb{R}^{d\times d}$ and $H_{t} \succeq c I, \forall  t \in [T] $, then $m = c$. This case covers all the adaptive algorithms with a lower bound for the matrix $H_t$. Such a lower bound can be achieved by adding  $m I$ to $H_t$ when it is a non-negative diagonal matrix or by assuming a lower bound of the gradient sizes. More details will be discussed in Section \ref{sec: algorithm}.
\end{example}

For the functions $\{f_i\}_{i=1}^n$, we assume that their gradients are all $L$-smooth, unbiased with mean $\nabla f$, and have bounded variance $\sigma^2$, which are all standard assumptions in the nonconvex optimization analysis literature \citep{ghadimi2016Minibatch, Reddi2016Proximal, li2018Simple}.

\begin{assumption}
\label{assumption: 2}
Each function $f_i$ is $L$-smooth, i.e., 
    \begin{equation}
    \begin{aligned}
    \label{def: smoothness}
        \|\nabla f_i(x) - \nabla f_i(y)\|_2 \leq L\|x-y\|_2 
        \nonumber
        \end{aligned}
    \end{equation}
\end{assumption}

\begin{assumption}
\label{assumption: 3}
Each $f_i(x)$ has unbiased stochastic gradients with bounded variance $\sigma^2$, i.e.,
\begin{equation}
    \begin{aligned}
    \label{def: bounded_variance}
        &\mathbb{E}_{i \sim [n]} \left[\nabla f_i(x) \right] =  \nabla f(x),  \mathbb{E}_{i \sim [n]} \left [\|\nabla f_i(x) - \nabla f(x)\|_2^2 \right] \leq \sigma^2 \nonumber
        \end{aligned}
\end{equation}
\end{assumption}

The convergence rates of different algorithms in the nonconvex optimization problem is usually measured by the stationarity of the gradient $\nabla f(x)$, i.e., $\mathbb{E}[\|\nabla f(x) \|^2] \leq \epsilon$ \footnote{Some recent works such as \citet{Zhou2018Stochastic}   use $\mathbb{E}[\|\nabla f(x) \|^2] \leq \epsilon^2$ instead of our choice. Simply replacing all the $\epsilon$ in our results by $\epsilon^2$ can generate the conclusions in their settings. }. However, due to the existence of $h(x)$ in the nonsmooth setting, such a definition is no longer intuitive as it does not indicate that the algorithm is stationary anymore. Instead, we use a more general definition of the generalized gradient and the related convergence criterion. 
\begin{definition}
       \label{def: generalized_gradient_deterministic}
Given the $x_t$ generated by Eqn. (\ref{eqn: composite_mirror_descent}), the generalized gradient ${g}_{X,t}$ is defined as 
\begin{equation}
    \begin{aligned}
    {g}_{X,t} = \frac{1}{\alpha_t}(x_t - x_{t+1}^+), \text{    where } 
     x_{t+1}^+ &= \text{argmin}_{x} \left\{\alpha_t \langle \nabla f(x_t), x \rangle + \alpha_t h(x) +  B_{\psi_t}(x, x_t) \right\} \nonumber
        \end{aligned}
\end{equation}
\end{definition}
Correspondingly, we change the convergence criterion into the stationarity of the generalized gradient $ \mathbb{E}[\|{g}_{X, t^*}\|^2] \leq \epsilon \nonumber$. 
In Definition \ref{def: generalized_gradient_deterministic}, we replace the stochastic gradient $g_t$ in Eqn. (\ref{eqn: composite_mirror_descent}) by the full-batch gradient $\nabla f(x_t)$. In other words, $x_{t+1}^+$ is the next-step parameter if we use the full batch gradient $\nabla f(x_t)$ for mirror descent at time $t$, and $g_{X,t}$ is the difference between $x_t$ and $x_{t+1}^+$ scaled by the step size. Therefore, the generalized gradient is small when $x_{t+1}^+$ and $x_t$ are close enough and thus the algorithm converges. We emphasize that Definition \ref{def: generalized_gradient_deterministic} and the convergence criterion are commonly observed in the mirror descent algorithm literature such as \citet{ghadimi2016Minibatch, li2018Simple}. If $\psi_t(x) = \frac{1}{2}\|x\|^2_2$, then the definition is equivalent to the generalized gradient in \citet{li2018Simple}. If we further assume $h(x)$ is a constant, then Definition \ref{def: generalized_gradient_deterministic} reduces to the original full batch gradient $\nabla f(x)$. When  $h(x)$ is a constant  and $\psi_t(x) =  \frac{1}{2}\langle x, H_t x\rangle$, the generalized gradient is equivalent to $H_t^{-1} \nabla f(x)$, which is the update 
rule of self-adaptive algorithms.

\subsection{Assumptions and Definitions for P-L condition}
Next, we present the additional assumptions and definitions needed for the convergence analysis under the generalized Polyak-Lojasiewicz (P-L) condition \citep{Polyak1963Gradient}, which is also known as the gradient dominant condition. The original P-L condition  (in smooth problems $h(x) = 0$) is defined as
\begin{equation}
    \begin{aligned}
       \nonumber
     \exists \mu > 0, ~s.t.~ \|\nabla F(x)\|^2 \geq 2\mu(F(x) - F^*), ~\forall x \in \mathbb{R}^d
        \end{aligned}
\end{equation}
which is even weaker than restricted strong convexity and it has been studied in many nonconvex convergence analyses such as \citet{reddi2016stochastic, Zhou2018Stochastic, lei2017Nonconvex}. Note that P-L condition implies all stationary points are global minimums. Therefore the original convergence criterion $(\mathbb{E}[\|\nabla F(x)\|^2] \leq \epsilon)$ is equivalent to
\begin{equation}
    \begin{aligned}
       \nonumber
      2\mu \left[{\mathbb{E}[F(x)] -F^*} \right]\leq \epsilon
        \end{aligned}
\end{equation}
However, because of the existence of $h(x)$ in nonsmooth problems, we utilize the definition of the generalized gradient $g_{X,t}$ to define the generalized P-L condition as follows.

\begin{definition}
\label{def: pl_condition}
The loss function $F(x)$ satisfies the generalized P-L condition if
\begin{equation}
    \begin{aligned}
    \nonumber
     \exists \mu > 0, ~ s.t. ~\|g_{X,t}\|^2 \geq 2\mu \left(F(x) - F^* \right), ~ \forall x \in \mathbb{R}^d
        \end{aligned}
\end{equation}
where $g_{X,t}$ is the generalized gradient defined in Definition \ref{def: generalized_gradient_deterministic}.
\end{definition}

\citet{li2018Simple} used a similar definition of the general P-L condition, and ours is a natural extension since the proximal functions in Eqn. \ref{eqn: composite_mirror_descent} are undetermined. The above definition reduces to the P-L condition by \citet{li2018Simple} when $\psi_t(x) = \frac{1}{2}\|x\|_2^2$. If we further assume that $h(x)$ is a constant, then Definition \ref{def: pl_condition} is the same as the original P-L condition. For simplicity, we further assume that the constant $\mu$ satisfies $L/(m^2\mu) \geq \sqrt{n}$, which is similar to what \citet{li2018Simple, Reddi2016Proximal} assumed in their papers. The condition is similar to the "high condition number regime" for strongly convex optimization problems and it is assumed only because we want to use the same step size $\alpha_t = m/L$ in all the theorems (especially for Theorem \ref{Thm: convergence of SMD PL} and Theorem \ref{Thm: convergence of SMD+VR PL}) in this paper. If it is not satisfied, we can simply use a more complicated step size setting as in the Appendix A.2 in \citet{li2018Simple}. 

\subsection{Comparison Measure} We use the stochastic first-order oracle (SFO) complexity to compare the convergence rates of different algorithms. When given the parameter $x$, SFO returns one stochastic gradient $\nabla f_i (x)$. In other words, the SFO complexity measures the number of gradient computations in the optimization process. We summarize the SFO complexity of a few mirror descent algorithms in both the nonconvex nonsmooth setting and the generalized P-L condition in Table \ref{tab: complexity}. 

\begin{table}[t]
\caption{Comparisons of the SFO complexity of different algorithms to reach $\epsilon$-stationary point of the generalized gradient. $n$ is the total number of samples and $b$ is the mini-batch size. $\widetilde{O}$ notation omits the logarithm term $\log  \frac{1}{\epsilon}$}.
\vspace{1pt}
\centering
\small
    \begin{tabular}{ l l l }
\hline
Algorithms & Nonconvex Nonsmooth & P-L condition \\
 \hline
\vspace{2pt}
ProxGD \citep{ghadimi2016Minibatch}& $O\left(\frac{n}{\epsilon}\right)$ & $\widetilde{O} \left(\frac{n}{\mu} \right)$ \\
\vspace{2pt}
ProxSVRG \citep{Reddi2016Proximal}& $O\left(\frac{n}{\epsilon \sqrt{b}}+ n\right)$ & $\widetilde{O} \left(\frac{n}{\mu\sqrt{b}}+ n \right)$  \\
\vspace{2pt}
SCSG \citep{lei2017Nonconvex} & $O\left(\frac{n^{2/3} b^{1/3}}{\epsilon} \wedge \frac{b^{1/3}}{\epsilon^{5/3}}\right)$  &  $\widetilde{O}\left(\frac{nb^{1/3}}{\mu} \wedge \frac{b^{1/3}}{\mu^{5/3} \epsilon^{2/3}} + n \wedge \frac{1}{\mu\epsilon}\right)$\\
\vspace{2pt}
ProxSVRG+ \citep{li2018Simple} &$O\left(\frac{n}{\epsilon \sqrt{b}} \wedge \frac{1}{\epsilon^2\sqrt{b}} + \frac{b}{\epsilon}\right) $ & $\widetilde{O}\left((n \wedge \frac{1}{\mu \epsilon}) \frac{1}{\mu\sqrt{b}} +  \frac{b}{\mu}\right)$ \\
\vspace{2pt}
{Adaptive SMD (Algorithm \ref{alg: SMD algorithm})} &  $O\left(\frac{n}{\epsilon} \wedge \frac{1}{\epsilon^2} \right)$ & $\widetilde{O}\left(\frac{n}{\mu} \wedge \frac{1}{\mu^2\epsilon} \right)$ \\
\vspace{2pt}
\textbf{SVRAMD (Algorithm \ref{alg: ASMDVR algorithm})}& $O\left(\frac{n}{\epsilon\sqrt{b}} \wedge \frac{1}{\epsilon^2\sqrt{b}} + \frac{b}{\epsilon} \right)$ & $\widetilde{O}\left((n \wedge \frac{1}{\mu \epsilon}) \frac{1}{\mu\sqrt{b}} + \frac{b}{\mu}\right)$\\
\hline
\end{tabular}
\label{tab: complexity}
\end{table}

\section{Algorithm and Convergence}
\label{sec: algorithm}

In this section, we present our main theoretical claims. In subsection \ref{subsec: SMD_convergence}, we provide the convergence rate of the Adaptive SMD algorithm as a competitive baseline, which matches the  best existing rates of non-adaptive mirror descent algorithms. In subsections \ref{subsec: SVRAMD_algorithm}, \ref{subsec: SVRAMD_convergence}, and \ref{subsec: SVRAMD_convergence_PL}, we present our generalized variance reduced mirror descent algorithm framework SVRAMD and provide its convergence analysis to show that variance reduction can accelerate the convergence of most adaptive mirror descent algorithms in both the nonconvex nonsmooth problem and the generalized P-L conditioned problem.

\subsection{The Adaptive Mirror Descent Algorithm and Its Convergence}
\label{subsec: SMD_convergence} 

We first establish the the SFO complexity of the general adaptive SMD algorithm (Algorithm \ref{alg: SMD algorithm}) as the baseline, which is to our best knowledge, a new result in literature. We prove in both the nonconvex nonsmooth case and the P-L conditioned case, the convergence rates of Algorithm \ref{alg: SMD algorithm} are similar to those of the non-adaptive SMD algorithm \citep{ghadimi2016Minibatch}, \citep{karimi2016linear}. The proof of these SFO complexities are provided in Appendix \ref{sec: appendixA}.

\begin{algorithm}[ht]
   \caption{General Adaptive SMD Algorithm}
   \label{alg: SMD algorithm}
\begin{algorithmic}[1]
   \STATE \textbf{Input:} Number of stages $T$, initial $x_1$, step sizes $\{\alpha_t\}_{t=1}^T$, proximal functions $\{\psi_t\}_{t=1}^T$
   \FOR{$ t= 1 $ \textbf{to} $T$}
   \STATE Randomly sample a batch $\mathcal{I}_t$ with size $b$
    \STATE $g_t = \nabla f_{\mathcal{I}_t}({x}_{t})$
   \STATE $x_{t+1} = \text{argmin}_x\{\alpha_t \langle g_{t}, x\rangle + \alpha_t h(x)+ B_{\psi_t}(x, x_{t}) \}$
   \ENDFOR
\STATE \textbf{Return} Uniformly sample $t^*$ from $\{t\}_{t=1}^T$ and ouput $x_{t^*}$
\end{algorithmic}
\end{algorithm}

\begin{theorem}
\label{Thm: convergence of SMD}
Suppose that $\psi_t(x)$ satisfies the $m$-strong convexity assumption (\textbf{1}), and $f$ satisfies the Lipschitz gradients and bounded variance assumptions (\textbf{2, 3}). Further assume that the learning rate and the mini batch sizes are set to be $\alpha_t = m/L,  b =n \wedge ({12\sigma^2}/(m^2\epsilon))$. Then the output of Algorithm \ref{alg: SMD algorithm} converges with gradient computations
\begin{equation}
\nonumber
    O \left(\frac{n}{\epsilon} \wedge \frac{\sigma^2}{\epsilon^2} + n \wedge \frac{\sigma^2}{\epsilon} \right)
\end{equation}
\end{theorem}

\textbf{Remark}. The above complexity can be treated as $O(n\epsilon^{-1} \wedge \epsilon^{-2})$ and it is the same as the complexity of non-adaptive mirror descent algorithms by \citet{ghadimi2016Minibatch}. Algorithm \ref{alg: SMD algorithm} needs a relatively large batch size ($O(\epsilon^{-1})$) to obtain a convergence rate close to that of GD ($O(n\epsilon^{-1})$) and SGD ($O(\epsilon^{-2})$) \citep{reddi2016stochastic}. However, it is still only asymptotically as fast as one of them, depending on the relationship between $O(\epsilon^{-1})$ and the sample size $n$.

We now provide the convergence rate of Algorithm \ref{alg: SMD algorithm} in the generalized P-L conditioned problem.

\begin{theorem}
\label{Thm: convergence of SMD PL}
Suppose that $\psi_{tk}(x)$ satisfies the $m$-strong convexity assumption (\textbf{1}), and $f$ satisfies the Lipschitz gradients and the bounded variance assumptions (\textbf{2, 3}). Further assume that the P-L condition (Definition \ref{def: pl_condition}) is satisfied. The learning rate and the mini-batch sizes are set to be $\alpha_t = m/L, b_t = n \wedge ({2(1+m^2)\sigma^2}/(\epsilon m^2\mu))$. Then the output of Algorithm \ref{alg: SMD algorithm} converges with gradient computations
\begin{equation}
\nonumber
  O \left((\frac{n}{\mu} \wedge \frac{\sigma^2}{\mu^2\epsilon})\log \frac{1}{\epsilon} \right) 
\end{equation}
\end{theorem}

\textbf{Remarks}. The above result is $\widetilde{O}({n}{\mu}^{-1} \wedge{\mu^{-2}\epsilon^{-1}})$ if we hide the logarithm term. Similar to Theorem \ref{Thm: convergence of SMD}, the SFO complexity matches the smaller complexity of ProxSGD and ProxGD in the P-L conditioned problem \citep{karimi2016linear}. We emphasize that since the general Algorithm \ref{alg: SMD algorithm} covers ProxSGD ($b=1$) and ProxGD ($b=n$) with $\psi_t(x) = \frac{1}{2}\|x\|_2^2$, we believe that our convergence rates in Theorem \ref{Thm: convergence of SMD} and Theorem \ref{Thm: convergence of SMD PL} are the best rates achievable.

\subsection{The General Variance Reduced Algorithm-SVRAMD}
\label{subsec: SVRAMD_algorithm}

We now present the Stochastic Variance Reduced Adaptive Mirror Descent (SVRAMD) algorithm, which is a simple and generalized variance reduction framework for any mirror descent algorithm. We provide the pseudo-code in Algorithm \ref{alg: ASMDVR algorithm}. The input parameters $B_t$ and $b_t$ are called the batch sizes and the mini-batch sizes. The algorithm consists of two loops: the outer loop has $T$ iterations while the inner loop has $K$ iterations. At each iteration of the outer loop (line 4), we estimate a large batch gradient $g_t = \nabla f_{\mathcal{I}_t}({x}_{t})$ with batch size $B_t$ at the reference point $x_t$, both of which will be stored during the inner loop. At each iteration of the inner loop (line 7), the large batch gradient is used to reduce the variance of the small batch gradient $\nabla f_{\widetilde{\mathcal{I}}_t}(y_{k}^t)$ so that the convergence becomes more stable.
Note that in line 8, the proximal function is an undetermined $\psi_{tk}(x)$ which can change over time, therefore Algorithm \ref{alg: ASMDVR algorithm} covers ProxSVRG+ and a lot more algorithms with different proximal functions. For example, if $\psi_{tk}(x) = \frac{1}{2}\|x\|_2^2$, then $B_{tk}(y, y_k^t) = \frac{1}{2}\|y - y_k^t\|_2^2$ and SVRAMD reduces to ProxSVRG+ \citep{li2018Simple}. If $\psi_{tk}(x) = \frac{1}{2} \langle x, H_{tk}x \rangle$, where $H_{tk} \succeq mI$, then SVRAMD reduces to variance reduced self-adaptive algorithms for different matrices $H_{tk}$, e.g., VR-AdaGrad in Section \ref{subsec: extension_to_self_adaptive}.

\begin{algorithm}[!ht]
   \caption{Stochastic Variance Reduced Adaptive Mirror Descent (SVRAMD)}
   \label{alg: ASMDVR algorithm}
\begin{algorithmic}[1]
   \STATE \textbf{Input:} Number of rounds $T$, initial $x_1 \in \mathbb{R}^d$, step sizes $\{\alpha_t\}_{t=1}^T$, batch, mini-batch sizes $\{B_t, b_t\}_{t=1}^T$, inner-loop iterations $K$, proximal functions $\{\psi_{tk}\}_{t=1, k=1}^{T,K}$
   \FOR{$ t= 1 $ \textbf{to} $T$}
   \STATE Randomly sample a batch $\mathcal{I}_t$ with size $B_t$
   \STATE $g_t = \nabla f_{\mathcal{I}_t}({x}_{t}); \quad y_1^t = x_{t}$
     \FOR{$ k= 1 $ \textbf{to} $K$}
        \STATE Randomly pick sample $\widetilde{\mathcal{I}}_t$ of size $b_t$ 
        \STATE $v_{k}^t = \nabla f_{\widetilde{\mathcal{I}}_t}(y_{k}^t) - \nabla f_{\widetilde{\mathcal{I}}_t}(y_{1}^t)+ g_t$
        \STATE $y_{k+1}^t = \text{argmin}_y\{\alpha_t \langle v_{k}^t, y\rangle + \alpha_t h(x)+ B_{\psi_{tk}}(y, y_{k}^t)\}$
     \ENDFOR
   \STATE $x_{t+1} = y^t_{K+1}$
   \ENDFOR
    \STATE \textbf{Return}  $x_{t^*}$ where $t^*$ is determined by\\
    \quad (Nonsmooth case) Uniformly sample $t^*$ from $[T]$; \\ 
    \quad (P-L condition case) ${t^*} = {T+1}$.
    \end{algorithmic}
\end{algorithm}

\subsection{Nonconvex Nonsmooth Convergence}
\label{subsec: SVRAMD_convergence}
Now given the general form of Algorithm \ref{alg: ASMDVR algorithm}, we provide its SFO complexity in the nonconvex nonsmooth problem in the following theorem. 

\begin{theorem}
\label{Thm: convergence of SMD+VR}
Suppose that $\psi_{tk}(x)$ satisfies the $m$-strong convexity assumption (\textbf{1}) and  $f$ satisfies the Lipschitz gradients and bounded variance assumptions (\textbf{2, 3}). Further assume that the learning rate, the batch sizes, the mini-batch sizes, and the number of inner-loop iterations are set to be $\alpha_t = m/L, ~ B_t = n \wedge ({20\sigma^2}/m^2\epsilon), ~b_t = b, ~K = \left \lfloor{\sqrt{b/20}}\right \rfloor \vee 1$. Then the output of Algorithm \ref{alg: ASMDVR algorithm} converges with SFO complexity
\begin{equation}
   O \left(\frac{n}{\epsilon\sqrt{b}} \wedge \frac{\sigma^2}{\epsilon^2\sqrt{b}} + \frac{b}{\epsilon} \right)\nonumber
\end{equation}
\end{theorem}

\textbf{Remarks}. The proof is provided in Appendix \ref{sec: appendixB}. Theorem \ref{Thm: convergence of SMD+VR} essentially claims that with Assumptions \textbf{\ref{assumption: 1}, \ref{assumption: 2}, \ref{assumption: 3}}, we can guarantee the SFO complexity of Algorithm \ref{alg: ASMDVR algorithm} is $ O \left(\frac{n}{\epsilon\sqrt{b}} \wedge \frac{\sigma^2}{\epsilon^2\sqrt{b}} + \frac{b}{\epsilon} \right)$,  with an undetermined mini-batch size $b$ to be tuned later.

Although $\alpha_t$ is fixed in the theorem, $\psi_{tk}$ can change over time and hence results in the adaptivity in the algorithm. For example, our theory indicates that using different designs of time-varying step sizes is allowed (Example \ref{example: 2} in Section \ref{sec: prelim}). When we take the proximal function to be $\psi_{tk}(x) = \frac{c_{tk}}{2}\|x\|_2^2, c_{tk} \geq m$, Algorithm \ref{alg: ASMDVR algorithm} reduces to ProxSVRG+ with time-varying effective step size $\alpha_t/c_{tk}$ (i.e., $\eta_t$ in \citet{li2018Simple}). As long as the effective step sizes $\alpha_t/c_{tk}$ are upper bounded by a constant $(m/L)/m = 1/L$, Algorithm \ref{alg: ASMDVR algorithm} still converges with the same complexity. Such a constraint is easy to satisfy for many step size schedules such as (linearly) decreasing step sizes, cyclic step sizes and warm up. Besides, $\psi_{tk}$ can be even more complicated, such as Example \ref{example: 1} and \ref{example: 3} we have mentioned in Section \ref{sec: prelim}, which will be discussed later. Another interesting result observed in our theorem is that when $m$ is small, we require a  relatively larger batch size $B_t$ to guarantee the fast convergence rate. We will later show that this intuition is actually supported by our experiments in Section \ref{sec: experiments}. 

Similar to ProxSVRG+, when SCSG \citep{lei2017Nonconvex} and ProxSVRG \citep{Reddi2016Proximal} achieve their best convergence rate at either a too small or a too large mini-batch size, i.e., $b =1$ or $b=n^{2/3}$, our algorithm achieves its best performance using a moderate mini-batch size $\epsilon^{- 2/3}$. We provide the following corollary.

\begin{corollary}
\label{Cor: convergence of SMD+VR}
Under all the assumptions and parameter settings in Theorem \ref{Thm: convergence of SMD+VR}, further assume that $b = \epsilon^{- 2/3}$, where $\epsilon^{- 2/3} \leq n$. Then the output of Algorithm \ref{alg: ASMDVR algorithm} converges with SFO complexity
\begin{equation}
\nonumber
   O \left(\frac{n}{\epsilon^{2/3}} \wedge \frac{1}{\epsilon^{5/3}} + \frac{1}{\epsilon^{5/3}} \right)    
\end{equation}
\end{corollary}

\textbf{Remarks}. The above SFO complexity is the same as the best SFO complexity of ProxSVRG+, and it is provably better than the SFO complexity of adaptive SMD in Theorem \ref{Thm: convergence of SMD}.  Note that the number of samples is usually very large in modern datasets, such as $n = 10^6$, thus $b = \epsilon^{- 2/3}\leq n$ can be easily satisfied with a constant mini-batch size and a small enough $\epsilon$, such as $b=10^2, \epsilon = 10^{-3}$. If the above best complexity cannot be achieved, meaning that either $n$ or $\epsilon$ is too small, some sub-optimal solutions are still available for fast convergence. For example, setting
$b = n^{2/3}$ would generate $O(n^{2/3}\epsilon^{-1})$ SFO complexity, which is also smaller than the SFO complexity of Algorithm \ref{alg: SMD algorithm}. Therefore, we conclude that if the parameters are carefully chosen, variance reduction can reduce the SFO complexity and accelerate the convergence of any adaptive SMD algorithm that satisfies Assumption \textbf{\ref{assumption: 1}} in the nonsmooth nonconvex problem. We summarize the SFO complexity generated by a few different choices of $b$ in the nonconvex nonsmooth problem in Table \ref{tab: batch_size}.

\begin{table}[t]
\caption{Summary of the SFO complexity of SVRAMD given different mini-batch sizes $b$. All notations follow Table \ref{tab: complexity}. SVRAMD is much faster than adaptive SMD in the middle three cases, i.e., $b = \epsilon^{-2/3}$, $b = \epsilon^{-1}$ and  $b = n^{2/3}$. It is asymptotically at most as fast as adaptive SMD when the mini-batch size is either too small or too big, i.e.,  $b=1$ and $b=n$ . 
}
\vspace{1pt}
\centering
\small
    \begin{tabular}{l l l }
\hline
\vspace{2pt}
Mini-batch & Nonconvex Nonsmooth & P-L condition \\
 \hline
\vspace{2pt}
$b = 1$ & $O \left(\frac{n}{\epsilon} \wedge \frac{1}{\epsilon^2} +  \frac{n}{\epsilon} \right)$ & $ \widetilde{O} \left((\frac{n}{\mu} \wedge \frac{1}{\mu^2 \epsilon}) \right) $  \\
\vspace{2pt}
$b = \epsilon^{-2/3} $ & $O \left(\frac{n}{\epsilon^{2/3}} \wedge \frac{1}{\epsilon^{5/3}} + \frac{1}{\epsilon^{5/3}} \right)$ & $ \widetilde{O} \left((\frac{n\epsilon^{1/3}}{\mu^{2/3}} \wedge \frac{\epsilon^{-2/3}}{\mu^{5/3}} + \frac{\epsilon^{-2/3}}{\mu^{5/3}}) \right)   $   \\
\vspace{2pt}
$b = \epsilon^{-1} $ & $   O\left(\frac{n}{\epsilon^{1/2}} \wedge \frac{1}{\epsilon^{3/2}} + \frac{1}{\epsilon^{2}}\right)$ & $ \widetilde{O} \left( \frac{n \epsilon^{1/2}}{\mu} \wedge \frac{1}{\mu^2 \epsilon^{1/2}} + \frac{1}{\mu \epsilon} \right)    $   \\
\vspace{2pt}
$b = n^{2/3}$ &  $O \left(\frac{n^{2/3}}{\epsilon} \wedge \frac{1}{\epsilon^{2}n^{1/3}} + \frac{n^{2/3}}{\epsilon}\right)$ & $ \widetilde{O} \left((\frac{n^{2/3}}{\mu}  \wedge \frac{1}{\mu^2 \epsilon})  + \frac{n^{2/3}}{\mu} \right)  $ \\
\vspace{2pt}
$b = n $ &$O \left(\frac{\sqrt{n}}{\epsilon } \wedge \frac{1}{\epsilon^2\sqrt{n}} + \frac{n}{\epsilon} \right) $ & $\widetilde{O} \left(( \frac{\sqrt{n}}{\mu} \wedge \frac{1}{\mu^2 \epsilon}) +  \frac{n}{\mu} \right)$   \\
\hline
\end{tabular}
\label{tab: batch_size}
\end{table}

\subsection{Convergence under the Generalized P-L Condition}
\label{subsec: SVRAMD_convergence_PL}
Now we provide the convergence of Algorithm \ref{alg: ASMDVR algorithm} under the generalized P-L condition (Definition \ref{def: pl_condition}). Note that the only difference in Algorithm \ref{alg: ASMDVR algorithm} in such problems is the final output, where we directly use the last $x_{T+1}$ rather than randomly sample from the historical $x_t$s. Therefore, we preserve the simplicity of our algorithm in the nonconvex nonsmooth case, i.e., there is no additional algorithmic components, such as the restart process in ProxSVRG \citep{Reddi2016Proximal}. Now we present the SFO complexity of our variance reduced algorithm under the generalized P-L condition and show its linear convergence rate.

\begin{theorem}
\label{Thm: convergence of SMD+VR PL}
Suppose that $\psi_{tk}(x)$ satisfies the $m$-strong convexity assumption (\textbf{1}) and  $f$ satisfies the Lipschitz gradients and bounded variance assumptions (\textbf{2, 3}). Further assume that the P-L condition (\ref{def: pl_condition}) is satisfied. The learning rate, the batch sizes, the mini-batch sizes, and  the number of inner loop iterations are set to be $\alpha_t = m/L, ~B_t = n \wedge ({10\sigma^2}/({\epsilon m^2\mu})), ~b_t = b, ~K = \lfloor{\sqrt{b/32}} \rfloor \vee 1$. Then the output of Algorithm \ref{alg: ASMDVR algorithm} converges with SFO complexity
\begin{equation}
\nonumber
   O \left((n \wedge \frac{\sigma^2}{\mu \epsilon}){ \frac{1}{\mu\sqrt{b}}\log \frac{1}{\epsilon}} + { \frac{b}{\mu}\log \frac{1}{\epsilon}} \right)    
\end{equation}
\end{theorem}

\textbf{Remarks}. The proof is provided in Appendix \ref{sec: appendixC}.  The above SFO complexity is of size $\widetilde{O}((n \wedge (\mu \epsilon)^{-1})({\mu\sqrt{b}})^{-1} + {b}{\mu}^{-1})$ if we hide the logarithm terms. Compared with the complexity of the original adaptive SMD algorithm, our complexity can be arguably smaller when we choose an appropriate mini-batch size $b$, which further proves our conclusion that variance reduction can be applied to most adaptive SMD algorithms to accelerate the convergence. We provide the following corollary for the best choice of $b$ to show its effectiveness.

\begin{corollary}
\label{Cor: convergence of SMD+VR PL}
Under all the assumptions and parameter settings in Theorem \ref{Thm: convergence of SMD+VR PL}, further assume that $b = (\mu\epsilon)^{- 2/3} \leq n$. Then the output of Algorithm \ref{alg: ASMDVR algorithm} converges with SFO complexity
\begin{equation}
\nonumber
   O \left((\frac{n\epsilon^{1/3}}{\mu^{2/3}} \wedge \frac{\epsilon^{-2/3}}{\mu^{5/3}} + \frac{\epsilon^{-2/3}}{\mu^{5/3}})\log \frac{1}{\epsilon} \right)    
\end{equation}
\end{corollary}

\textbf{Remarks}. The above complexity is the same as the best complexity of ProxSVRG+ and it is smaller than the complexity of adaptive SMD in Theorem \ref{Thm: convergence of SMD PL}. Moreover, it generalizes the best results of ProxSVRG/ProxSAGA and SCSG, without the need to perform any restart or use the stochastically controlled iterations trick. Therefore, Algorithm \ref{alg: ASMDVR algorithm} automatically switches to fast convergence in regions satisfying the generalized P-L condition. We also provide the SFO complexity of some other choices of $b$ in Table \ref{tab: batch_size} for a clearer comparison. Similarly, if $b = (\mu\epsilon)^{- 2/3}$ is impractical, using $b = n^{2/3}$ also generates faster convergence than adaptive SMD.

\subsection{Extensions to Self-Adaptive Algorithms}
\label{subsec: extension_to_self_adaptive}

\begin{algorithm}[!ht]
   \caption{Variance Reduced AdaGrad Algorithm}
   \label{alg: VR_AdaGrad}
    \begin{algorithmic}[1]
   \STATE \textbf{Input:} Number of stages $T$, initial $x_1$, step sizes $\{\alpha_t\}_{t=1}^T$, batch sizes $\{B_t\}_{t=1}^T$, mini-batch sizes $\{b_t\}_{t=1}^T$, constant $m$
   \FOR{$ t= 1 $ \textbf{to} $T$}
   \STATE Randomly sample a batch $\mathcal{I}_t$ with size $B_t$
   \STATE $g_t = \nabla f_{\mathcal{I}_t}({x}_{t})$
   \STATE $y_1^t = x_{t}$
     \FOR{$ k= 1 $ \textbf{to} $K$}
        \STATE Randomly pick sample $\Tilde{\mathcal{I}}_t$ of size $b_t$ 
        \STATE $v_{k}^t = \nabla f_{\Tilde{\mathcal{I}}_t}(y_{k}^t) - \nabla f_{\Tilde{\mathcal{I}}_t}(y_{1}^t)+ g_t$
        \STATE $H_{tk} = \text{diag}(\sqrt{\sum_{\tau=1}^{t-1}\sum_{i=1}^K v_i^{\tau 2} + \sum_{i=1}^k v_i^{t2}}+m)$
        \STATE $y_{k+1}^t = \text{argmin}_y\{\alpha_t \langle v_{k}^t, y\rangle + \alpha_t h(x)+ \frac{1}{2}\langle y-y_{k}^t, H_{tk} (y-y_{k}^t) \rangle\}$
     \ENDFOR
   \STATE $x_{t+1} = y^t_{K+1}$
   \ENDFOR
    \STATE \textbf{Return} (Smooth case) Uniformly sample $x_{t^*}$ from $\{y_{k}^t\}_{k=1,t=1}^{K,T}$; (P-L case) $x_{t^*} = x_{T+1}$
    \end{algorithmic}
\end{algorithm}

As we have mentioned in Section \ref{sec: prelim}, self-adaptive algorithms such as AdaGrad are special cases of the general Algorithm \ref{alg: SMD algorithm} and thus we can use Algorithm \ref{alg: ASMDVR algorithm} to accelerate them. The proximal functions of most adaptive algorithms have the form of $\psi_t(x) = \frac{1}{2}\langle x, H_t x \rangle$, where $H_t \in \mathbb{R}^{d\times d}$ is often designed to be a diagonal matrix. 

Assumption \textbf{\ref{assumption: 1}} is satisfied if we consistently add a constant matrix $m I$ to $H_t$, with $I$ being the identity matrix. We remark that such an operation is commonly used in self-adaptive algorithms to avoid division by zero \citep{Duchi2010Composite, Kingma2015Adam}, and thus variance reduction can be applied. Assumption \textbf{\ref{assumption: 1}} can also be satisfied for many special designs of $H_t$ with some mild assumptions. For example, if $H_{t,ii} = \sqrt{g_{1,i}^2 + g_{2,i}^2 + \cdots g_{t,i}^2}, \forall i \in [d]$ (the original AdaGrad algorithm), then $H_{t} \succeq m I$ if there exists one $\tau_i \in [t]$ on each dimension $i$ such that $g_{\tau_i, i} \geq m$, meaning that there is at least one derivative larger than $m$ on each dimension in the whole optimization process. If such a mild condition holds, then the conclusions in Theorem \ref{Thm: convergence of SMD+VR}, \ref{Thm: convergence of SMD+VR PL} and Corollary \ref{Cor: convergence of SMD+VR}, \ref{Cor: convergence of SMD+VR PL} still hold.

However, notice that the strong convexity lower bounds of these algorithms are relatively smaller ($m$ is often set as 1e-3 or even smaller in real experiments), Theorem \ref{Thm: convergence of SMD+VR} implies that the batch size $B_t$ needs to be sufficiently large for these algorithm to converge fast. If variance reduction can work with such algorithms with small $m$, then we should expect good performances with the other algorithms that have large $m$'s.  We provide the implementation for Variance Reduced AdaGrad (VR-AdaGrad) in Algorithm \ref{alg: AdaGrad+VR} and Variance Reduced RMSProp (VR-RMSProp) is similar. 

One interesting question here is whether VR-AdaGrad and VR-RMSProp can converge even faster than ProxSVRG+, since AdaGrad and RMSProp are known to converge faster than SGD in convex problems. Our conjecture is that they cannot do so, at least not in the nonconvex case because their rapid convergence relies on not only the convexity of the problem, but also the sparsity of the gradients \citep{Duchi2011Adaptive}. Neither of these two assumptions is appropriate in the nonconvex analysis framework.

\section{Experiments}
\label{sec: experiments}

In this section, we present several experiments on neural networks to show the effectiveness of variance reduction in the adaptive SMD algorithms.  We train a two-layer fully connected network on the MNIST dataset, the LeNet \citep{LeCun1998Gradient} on the CIFAR-10 dataset, and the ResNet-20 model on the CIFAR-100 dataset \citep{He2016Deep}. 

\textbf{Implementations}. In general, we follow the settings by \citet{Zhou2018Stochastic} on SNVRG for all our experiments.  Except for normalization, we do not perform any additional data transformation or augmentation techniques such as rotation, flipping, and cropping on the images.  For the batch sizes and mini batch sizes $B_t$ and $b_t$ in the algorithms, we follow \citet{Zhou2018Stochastic} to use a slightly different but equivalent notation of batch size ratio $ r = B_t/b_t$. Note that increasing this ratio when $b_t$ is fixed would be the same as increasing the batch size $B_t$. Our code is based on the publicly released PyTorch code by \citet{SVRGcode}. All experiments here are run on Nvidia V100 GPUs. We have thoroughly tuned the hyper-parameters for all the algorithms on the different datasets. More details of our experiments can be found in Appendix \ref{sec: appendixD}.

\begin{figure}[ht]
     \centering
     \begin{subfigure}[b]{0.24\linewidth}
        \centering
        \includegraphics[width=\linewidth]{  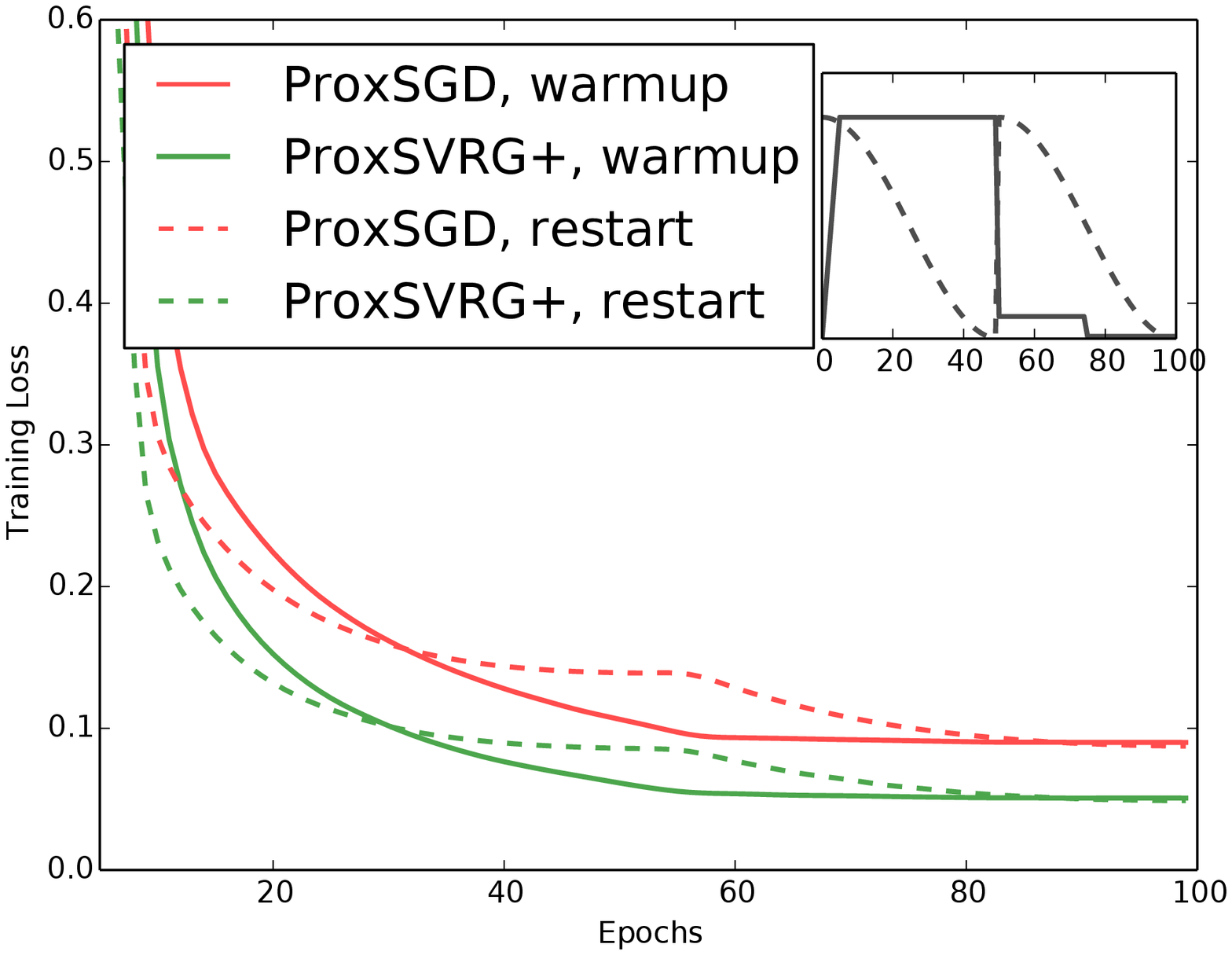}
        \caption{MNIST Training Loss}
         \label{fig: MNIST_lrschedule_train}
     \end{subfigure}
     \hfill
     \begin{subfigure}[b]{0.24\linewidth}
         \centering
  \includegraphics[width=\linewidth]{  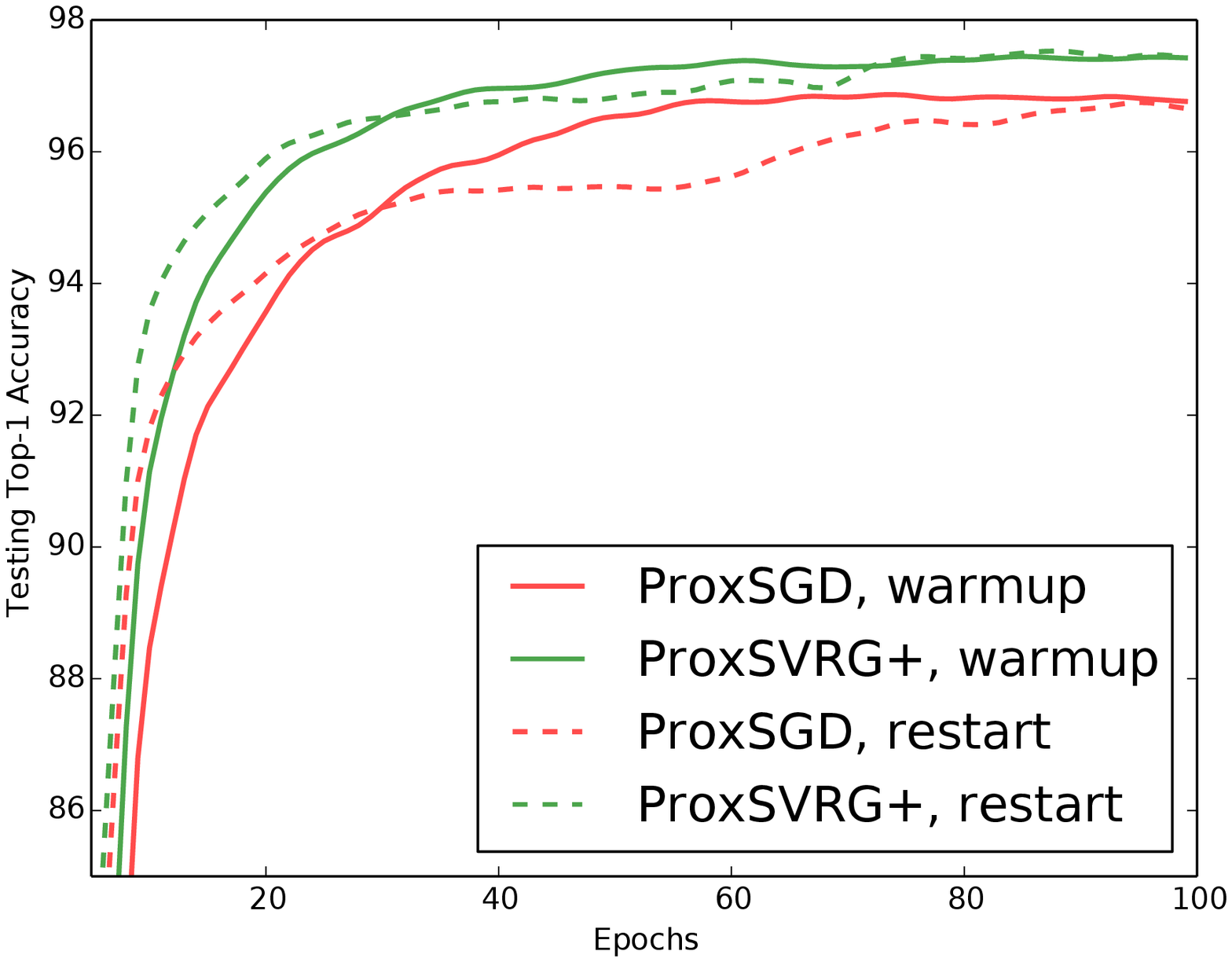}
    \caption{MNIST Testing Acc.}
    \label{fig: MNIST_lrschedule_test}
     \end{subfigure}
     \hfill
     \begin{subfigure}[b]{0.24\linewidth}
        \centering
  \includegraphics[width=\linewidth]{  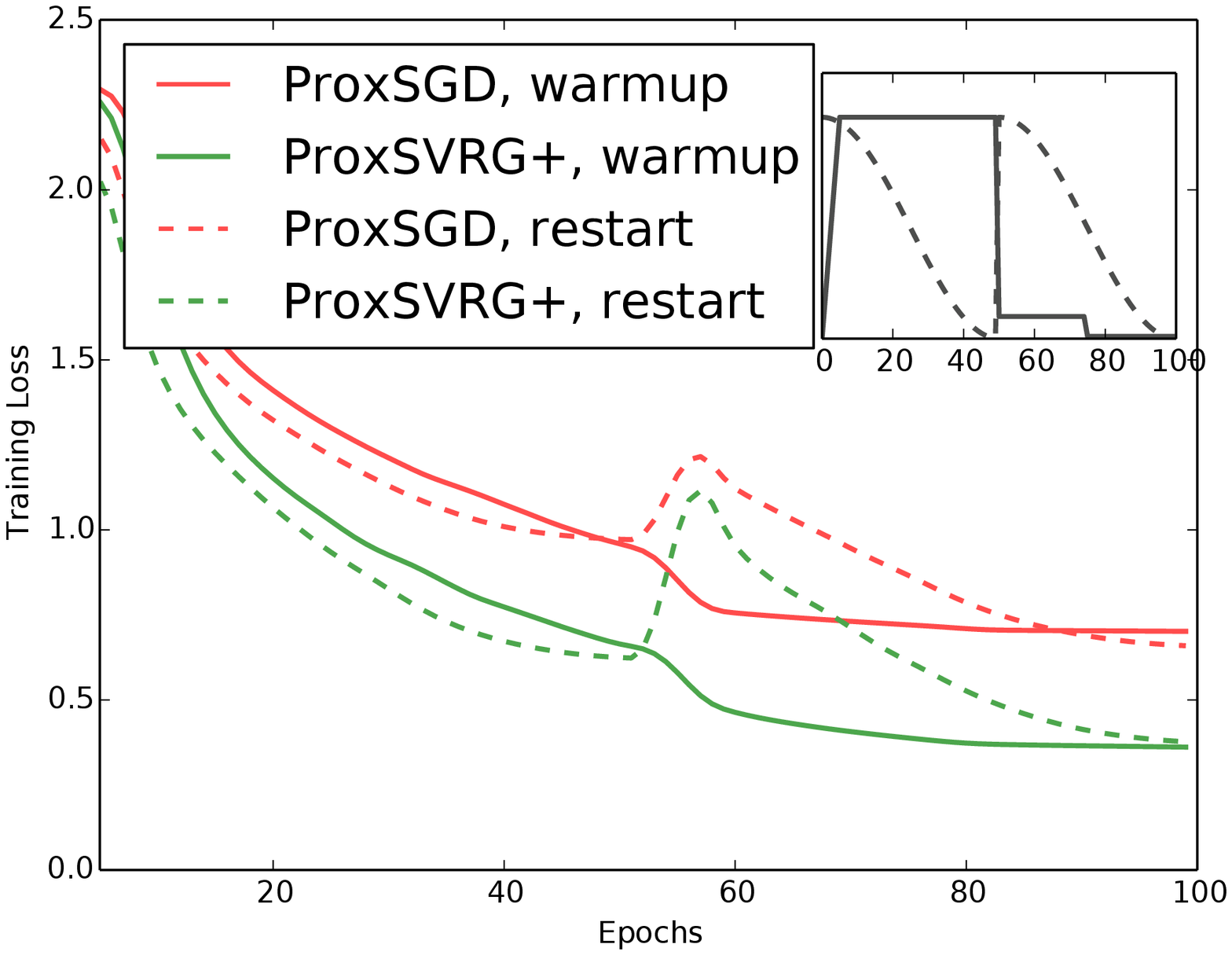}
  \caption{  CIFAR10 Training Loss.}
  \label{fig: Cifar10_lrschedule_train}
     \end{subfigure}
     \hfill
     \begin{subfigure}[b]{0.24\linewidth}
         \centering
  \includegraphics[width=\linewidth]{  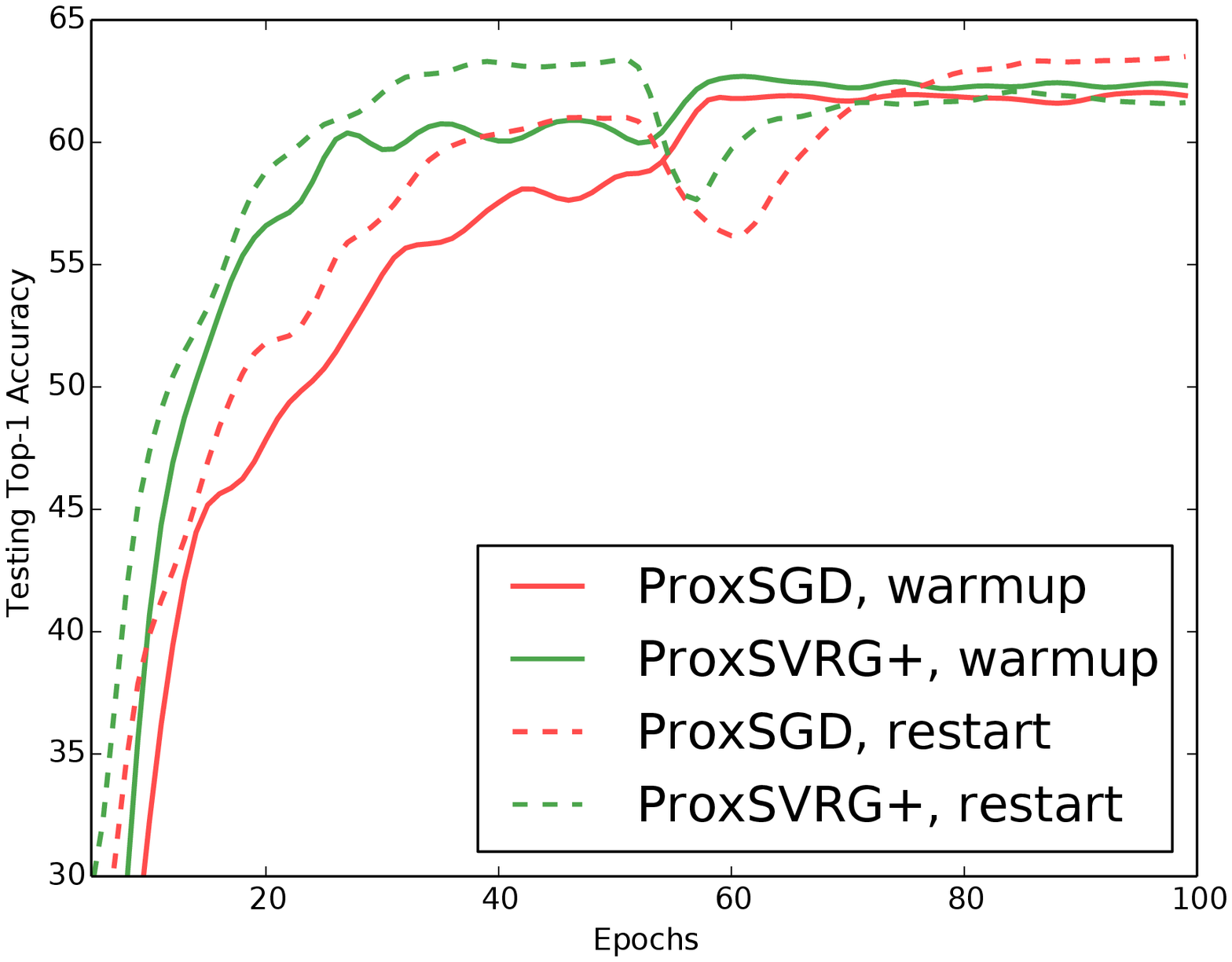}
  \caption{CIFAR10 Testing Acc.}
      \label{fig: Cifar10_lrschedule_test}
     \end{subfigure}
     \caption{ Performance of ProxSGD and ProxSVRG+ with different step size schedules. The solid lines are trained with the warmup schedule (warmup for 5 epochs then step decay) and the dashed lines are trained with the warm restart schedule. We plot the two schedules at the top-right corners of the training loss subfigures for the readers' reference. \textbf{(a) and (b)}: training loss and testing accuracy using fully connected network on MNIST. \textbf{(c) and (d)}: training loss and testing accuracy using LeNet on CIFAR-10. The results were averaged over 5 runs.}
     \label{fig: LRschedule}
\end{figure}

\begin{figure}[ht]
     \centering
     \begin{subfigure}[b]{0.33\linewidth}
        \centering
        \includegraphics[width=\linewidth]{  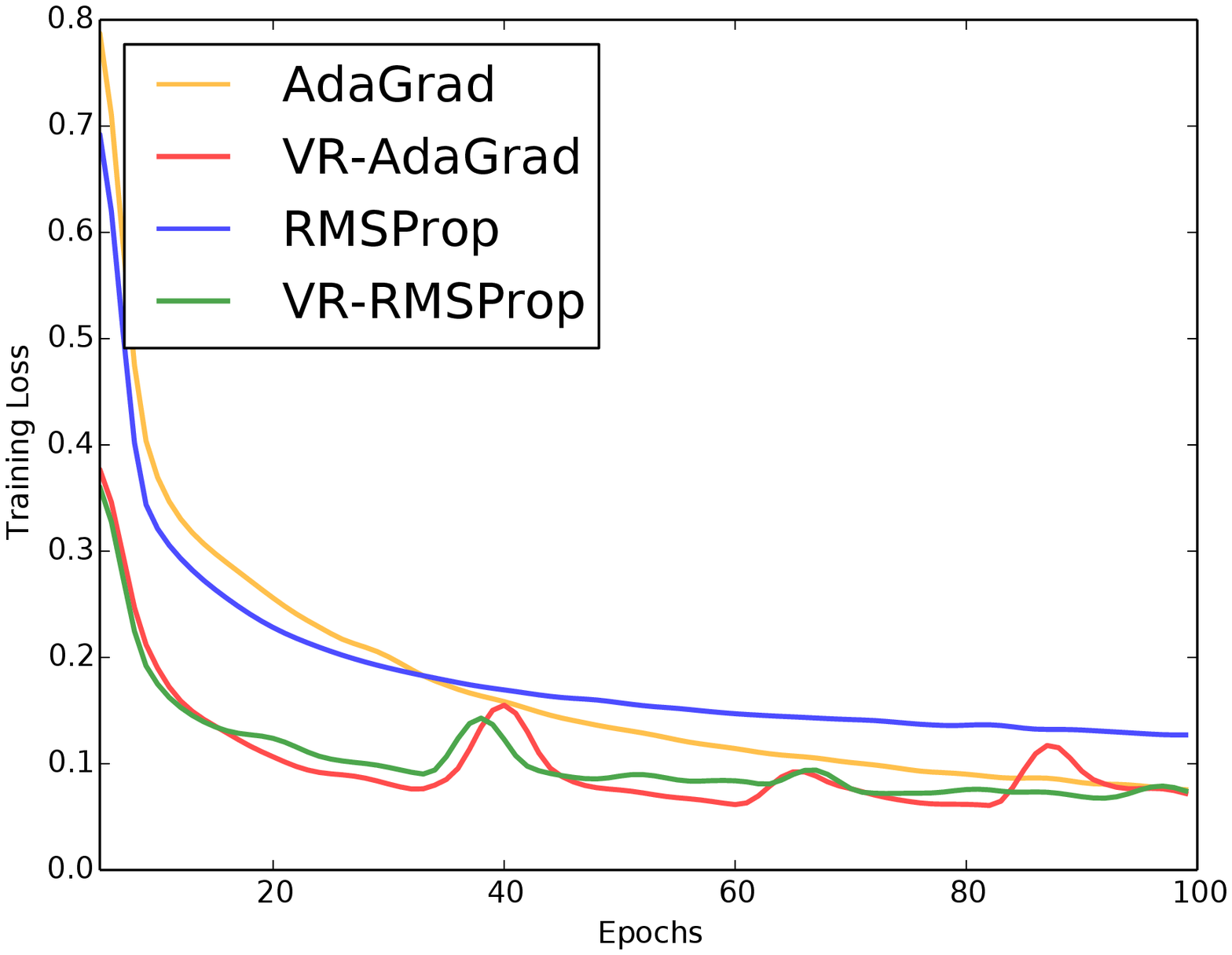}
        \caption{MNIST Training Loss}
         \label{fig: MNIST_1}
     \end{subfigure}
     \hspace*{-1em}
     \begin{subfigure}[b]{0.33\linewidth}
         \centering
        \includegraphics[width=\linewidth]{  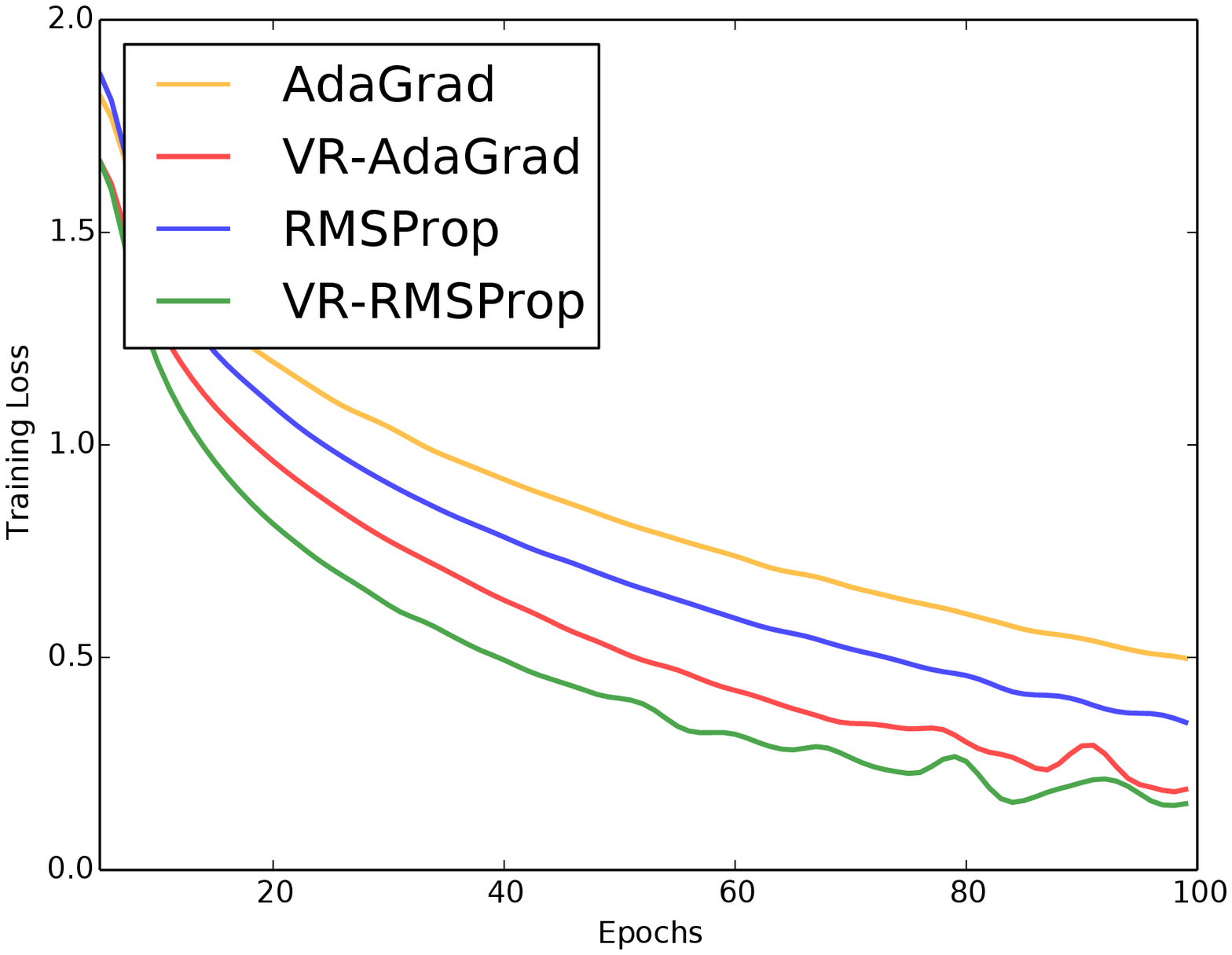}
        \caption{CIFAR10 Training Loss.}
        \label{fig: Cifar10_1}
     \end{subfigure}
    \hspace*{-1em}
     \begin{subfigure}[b]{0.33\linewidth}
        \centering
        \includegraphics[width=\linewidth]{  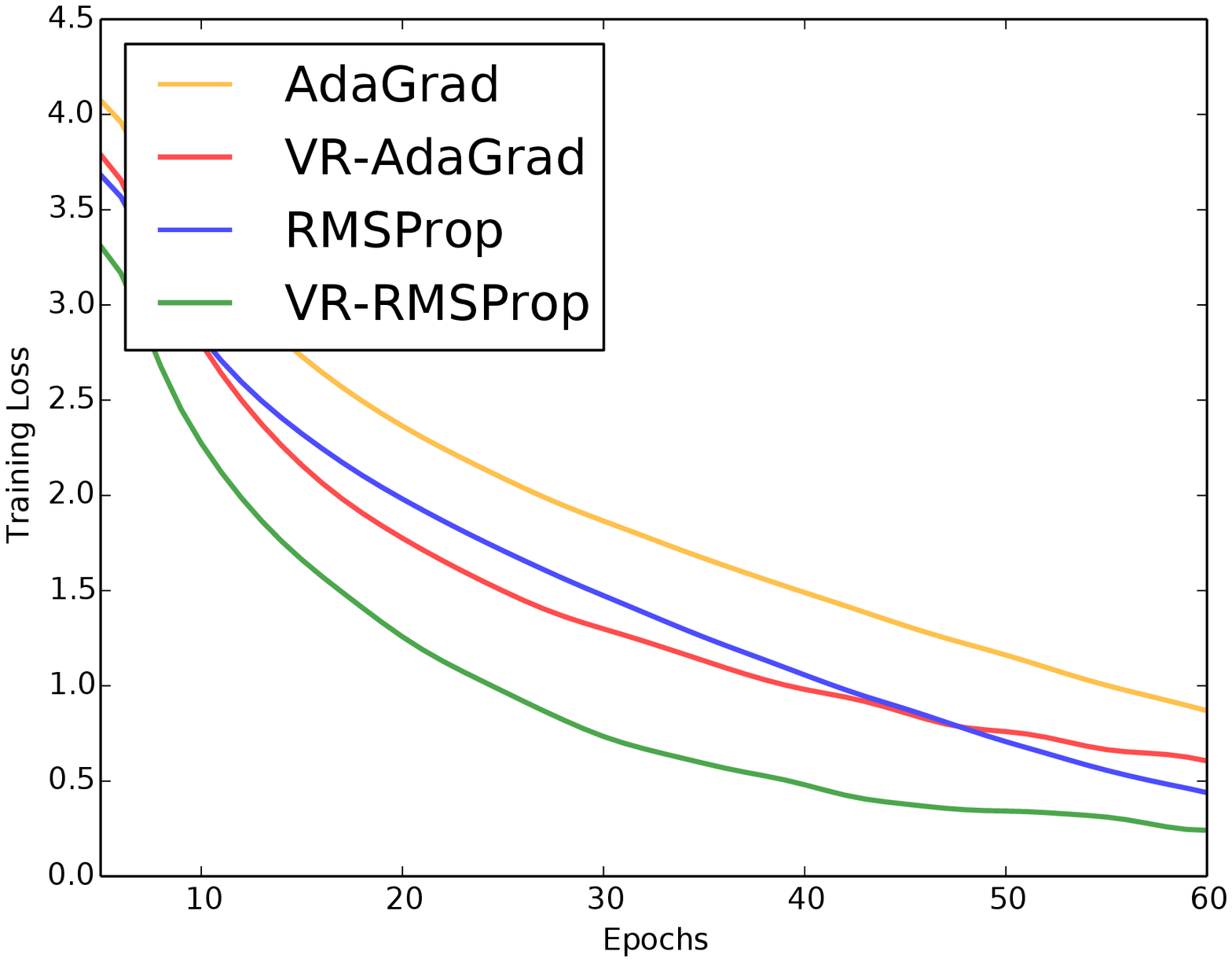}
        \caption{  CIFAR100 Training Loss.}
        \label{fig: Cifar100_1}
     \end{subfigure}
     \hfill
     \begin{subfigure}[b]{0.33\linewidth}
         \centering
        \includegraphics[width=\linewidth]{  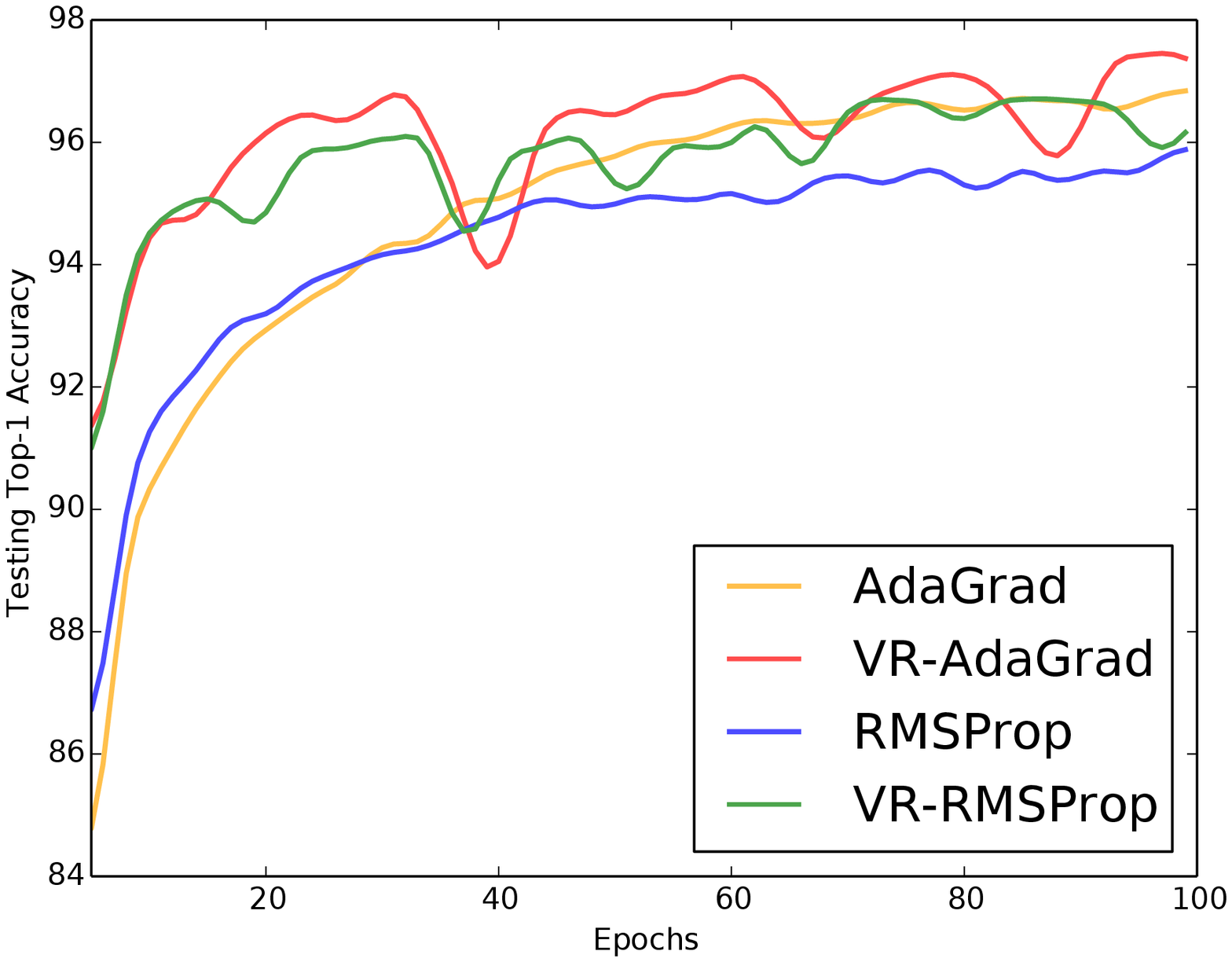}
        \caption{MNIST Testing Acc.}
          \label{fig: MNIST_2}
     \end{subfigure}
          \hspace*{-1em}
      \begin{subfigure}[b]{0.33\linewidth}
            \centering
             \includegraphics[width=\linewidth]{  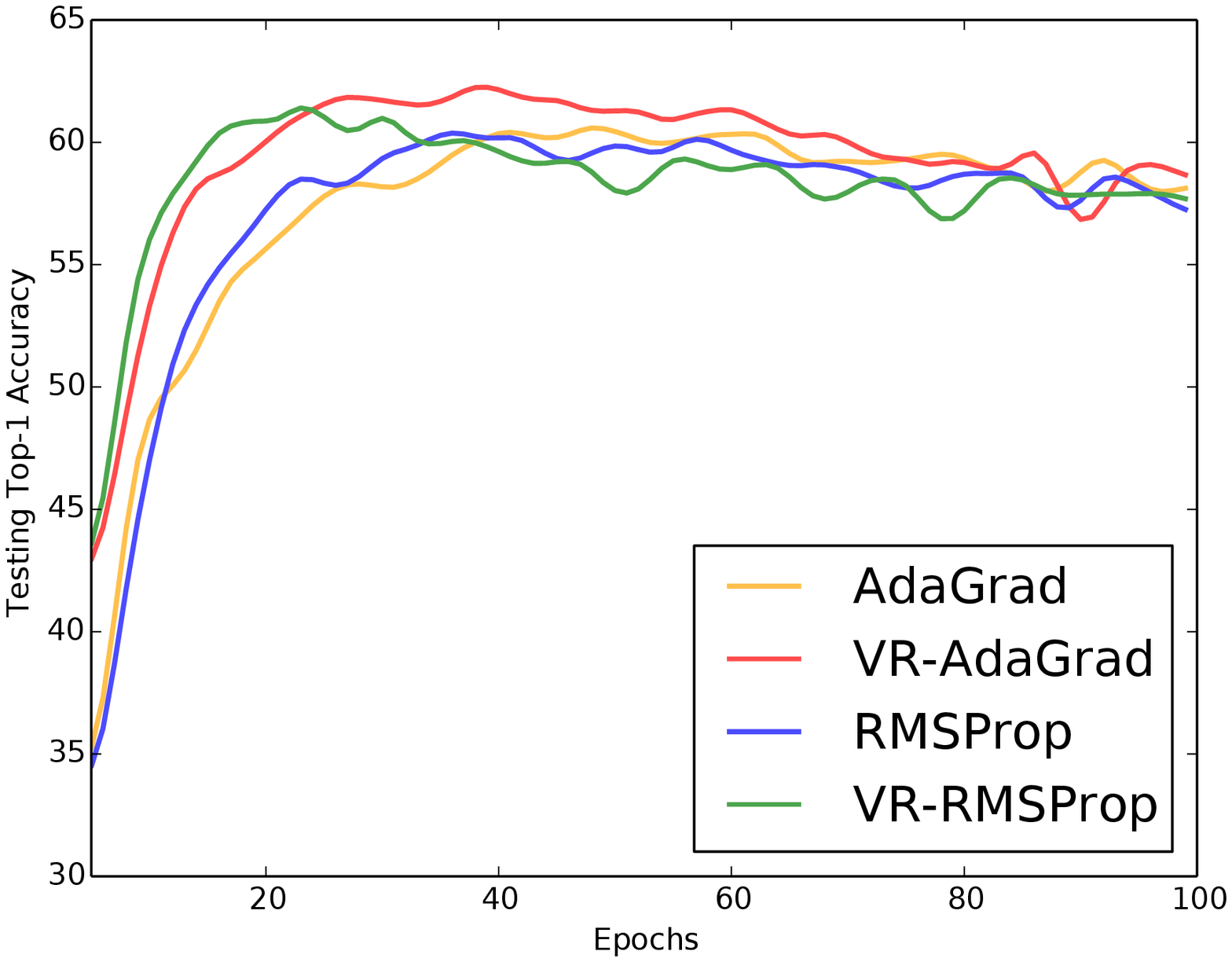}
            \caption{CIFAR10 Testing Acc.}
                  \label{fig: Cifar10_2}
     \end{subfigure}
          \hspace*{-1em}
     \begin{subfigure}[b]{0.33\linewidth}
             \centering
        \includegraphics[width=\linewidth]{  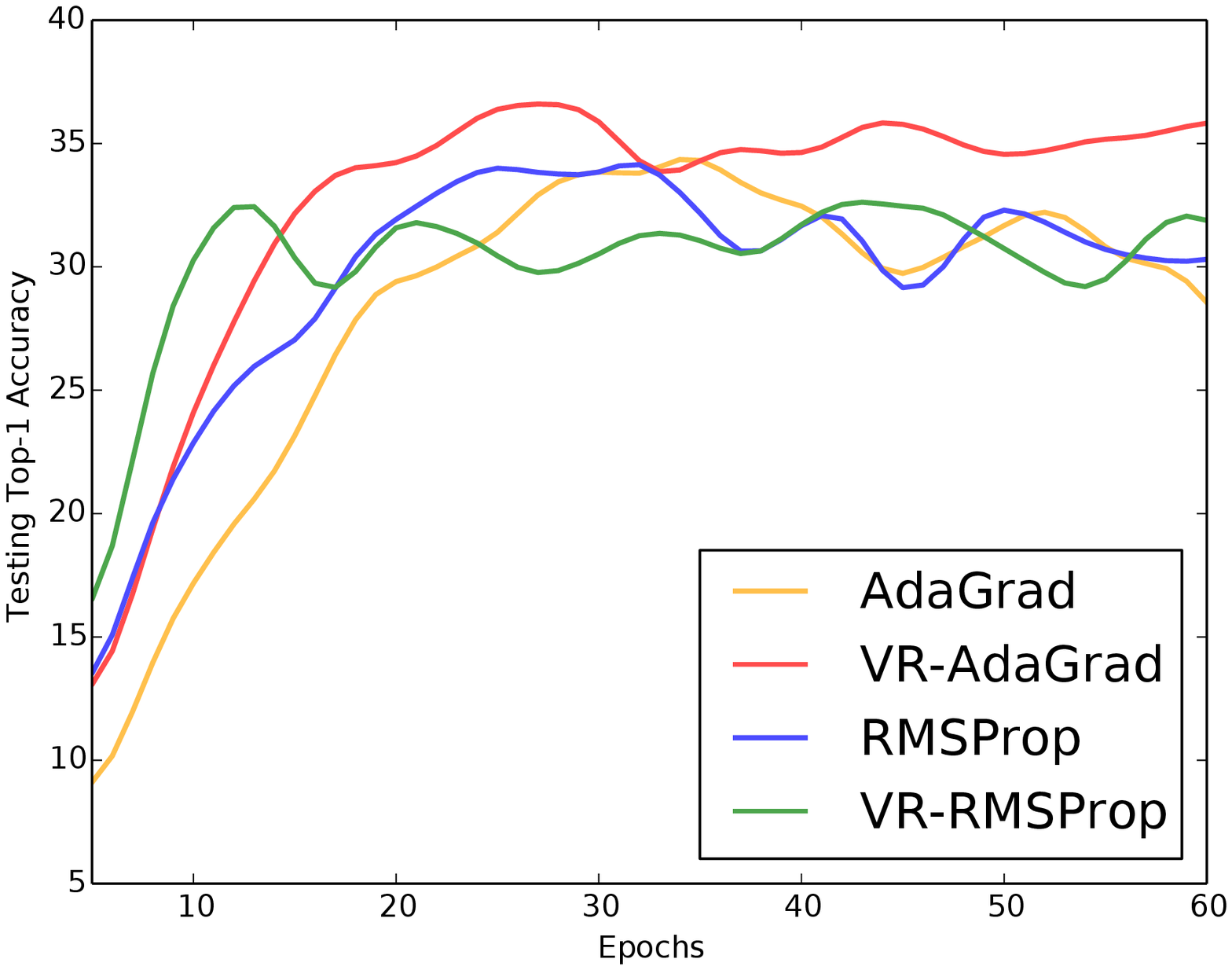}
        \caption{CIFAR100 Testing Acc.}
      \label{fig: Cifar100_2}
     \end{subfigure}
    \caption{{(a) and (d)}: training loss and testing accuracy on MNIST. {(b) and (e)}: training loss and testing accuracy on CIFAR10. (c) and (f):  training loss and testing accuracy on CIFAR100. The results are averaged over 5 runs.    }
    \label{fig: Training and Testing}
\end{figure}

\begin{figure}[ht]
     \centering
     \begin{subfigure}[b]{0.33\linewidth}
        \centering
  \includegraphics[width=\linewidth]{  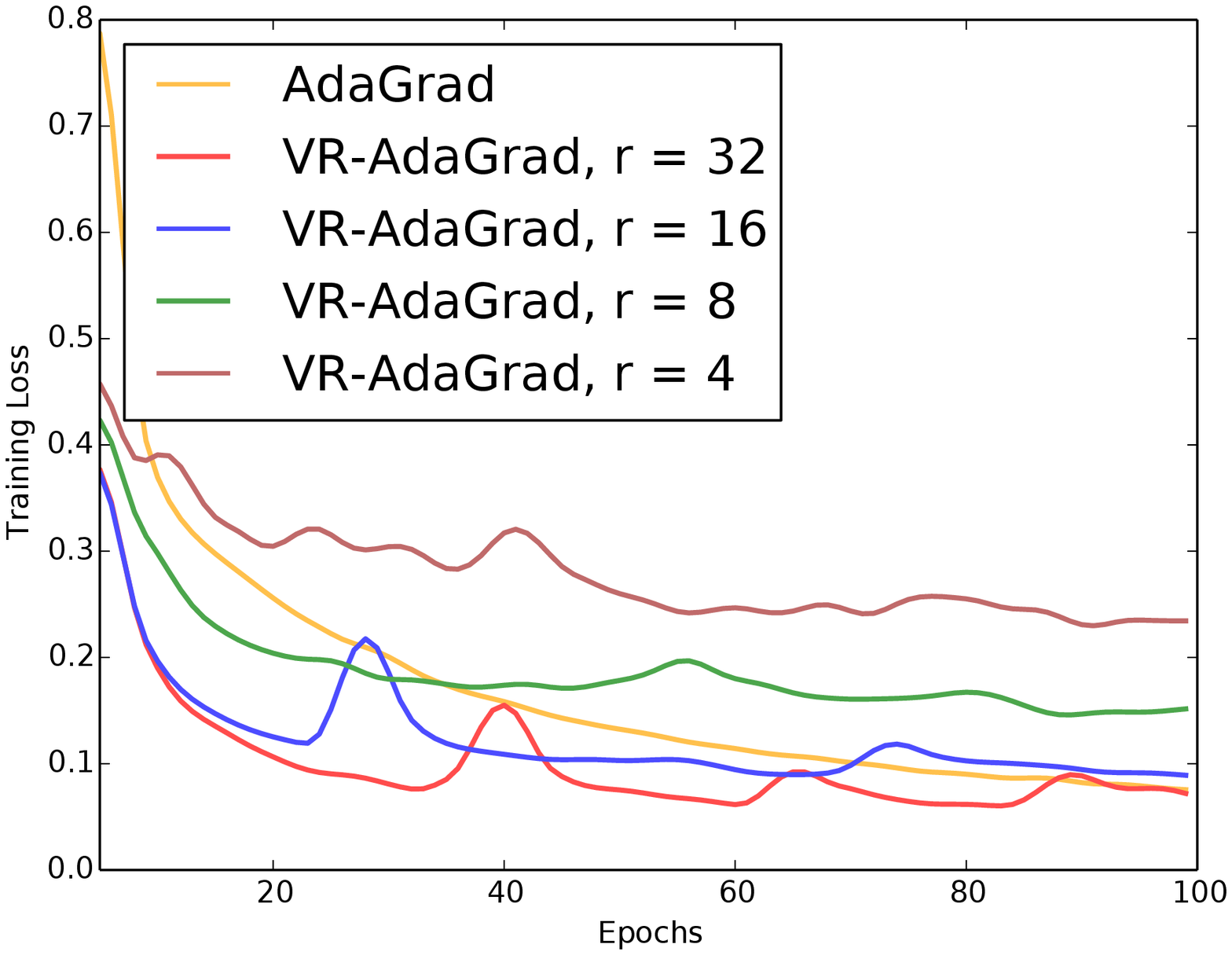}
  \caption{VR-AdaGrad on MNIST}
  \label{fig: MNIST_AdaGrad_batch}
     \end{subfigure}
     \hspace*{-0.5em}
     \begin{subfigure}[b]{0.33\linewidth}
         \centering
  \includegraphics[width=\linewidth]{  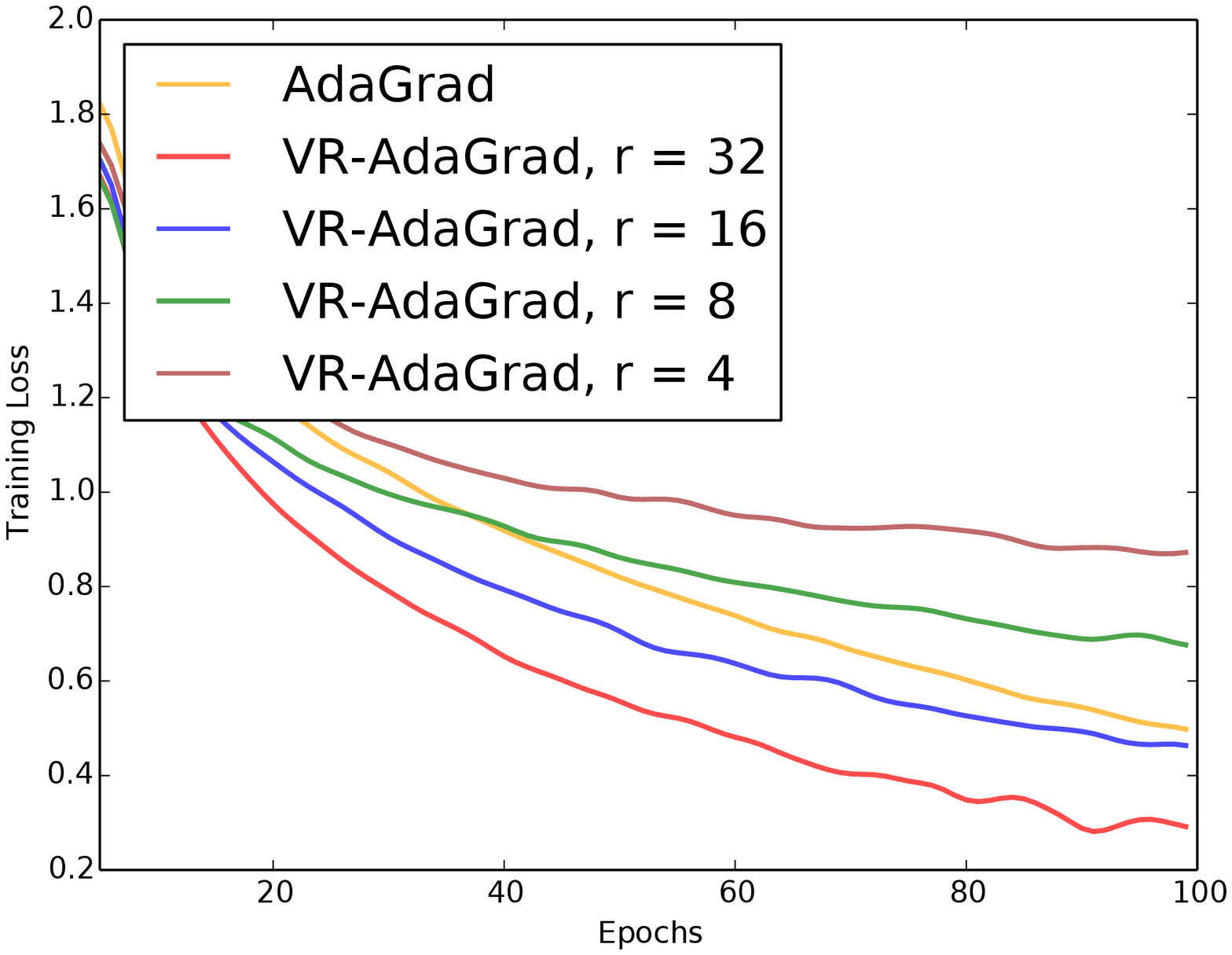}
    \caption{VR-AdaGrad on CIFAR10}
  \label{fig: Cifar10_AdaGrad_batch}
     \end{subfigure}
    \hspace*{-0.5em}
     \begin{subfigure}[b]{0.33\linewidth}
        \centering
         \includegraphics[width=\linewidth]{  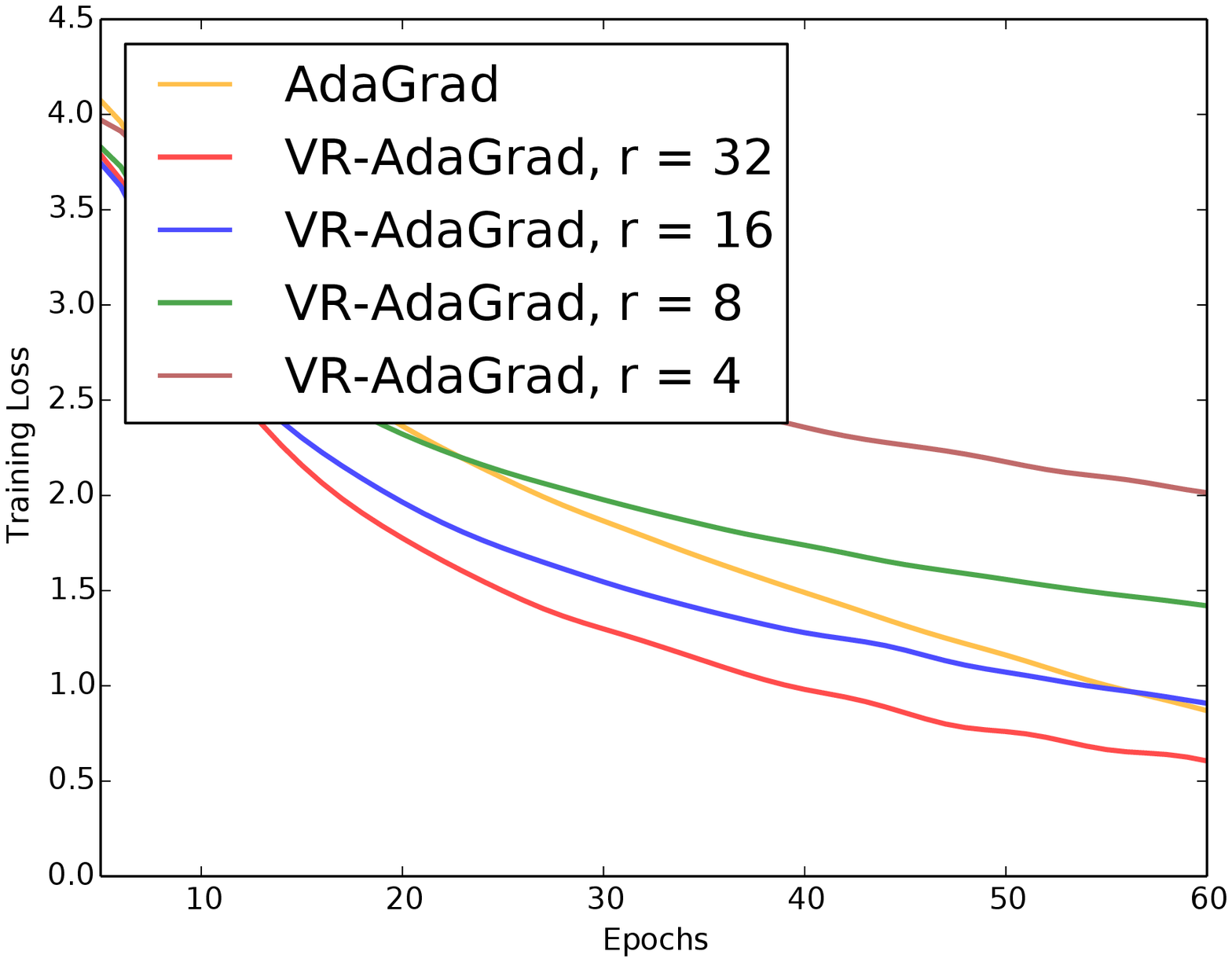}
            \caption{VR-AdaGrad on CIFAR100}
            \label{fig: Cifar100_AdaGrad_batch}
     \end{subfigure}
         \hspace*{-0.5em}
     \begin{subfigure}[b]{0.33\linewidth}
         \centering
            \includegraphics[width=\linewidth]{  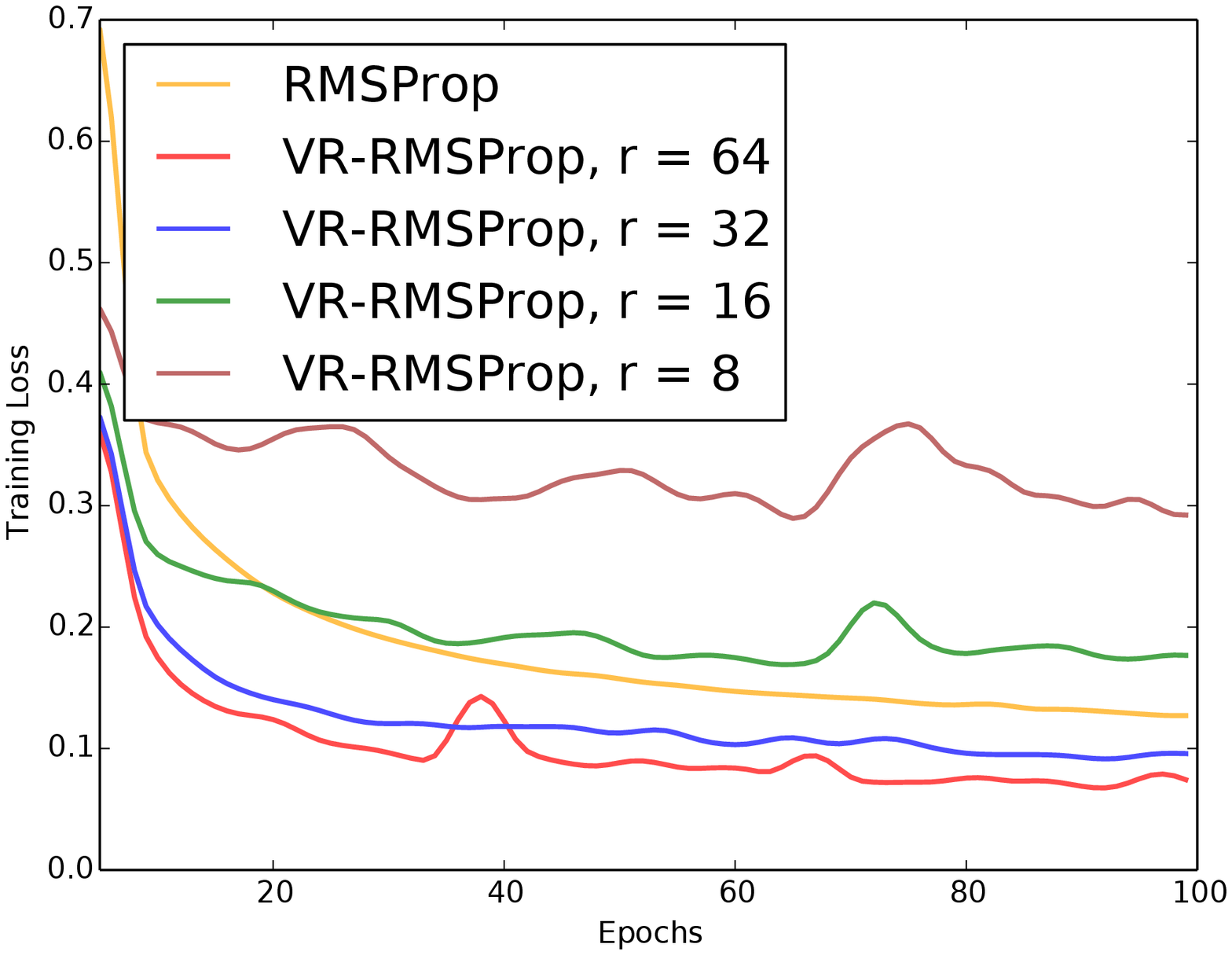}
            \caption{VR-RMSProp on MNIST}
            \label{fig: MNIST_RMSProp_batch}
     \end{subfigure}
          \hspace*{-0.5em}
      \begin{subfigure}[b]{0.33\linewidth}
            \centering
             \includegraphics[width=\linewidth]{  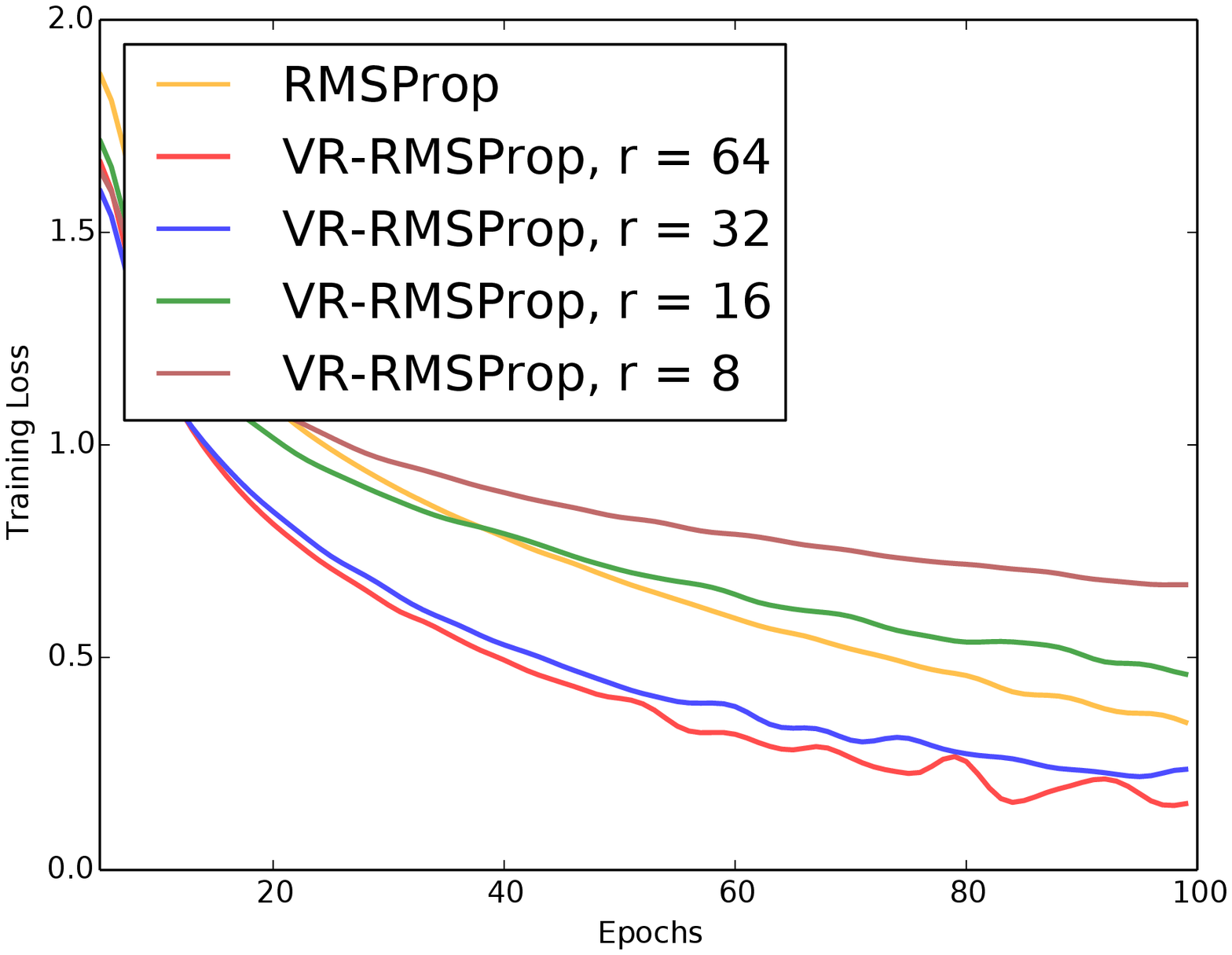}
               \caption{VR-RMSProp on CIFAR10}
                \label{fig: Cifar10_RMSProp_batch}
     \end{subfigure}
          \hspace*{-0.5em}
     \begin{subfigure}[b]{0.33\linewidth}
             \centering
        \includegraphics[width=\linewidth]{  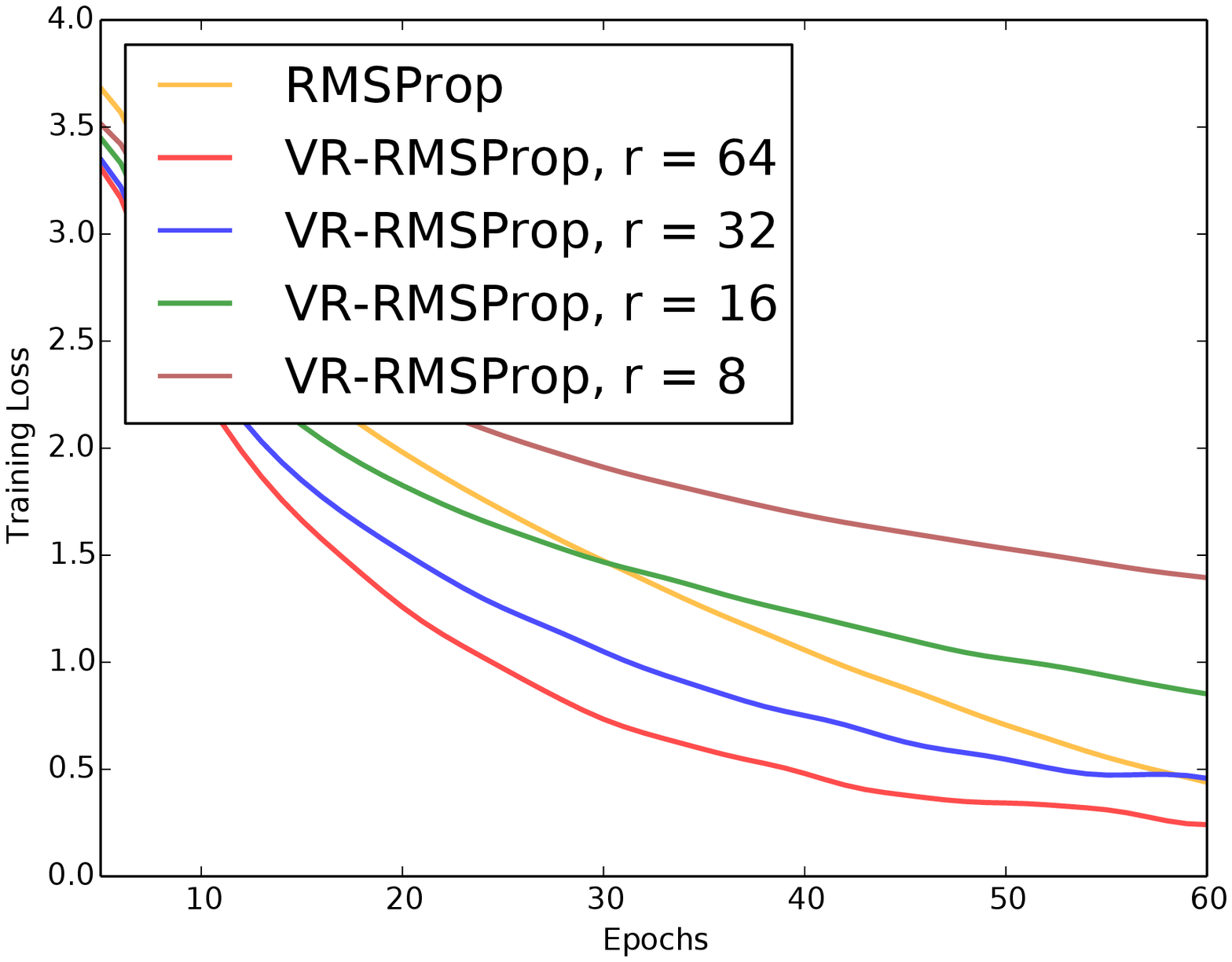}
          \caption{VR-RMSProp on CIFAR100}
      \label{fig: Cifar100_RMSProp_batch}
     \end{subfigure}
    \caption{{(a) and (d)}: training loss with different $r$ on  MNIST. {(b) and (e)}: training loss with different $r$ on CIFAR10 . (c) and (f): training loss with different ratio $r$ on CIFAR100.}
\label{fig: Batch_Size}
\end{figure}

We first validate our claims in Section \ref{sec: algorithm} by showing that variance reduction can work well with different step size schedules. We choose two special schedules that are both complicated enough and popular nowadays, warm up and warm restart. We choose to compare ProxSGD and ProxSVRG+ because they are special cases of Algorithm \ref{alg: SMD algorithm} and Algorithm \ref{alg: ASMDVR algorithm} with proximal function $\frac{1}{2\alpha_t}\|x\|_2^2$, where $\alpha_t$ is the step size. For warm up, we increase the step size from zero to the best step size of ProxSGD/ProxSVRG+ linearly for the initial 5 epochs and then multiply the step size by 0.1 at the 50-th and the 75-th epoch. For warm restart, we decrease from the best step size to zero with cosine annealing in 50 epochs and then restart the decay. The training loss, the testing accuracy and the step size schedules on 
MNIST and CIFAR-10 are plotted in Figure \ref{fig: LRschedule}. As can be seen in the figures, ProxSVRG+ converges faster than ProxSGD with both of the two special step size schedules, proving our claim that as long as the step sizes are bounded, variance reduction can still improve the convergence rate of ProxSGD with time-varying step sizes.

We then choose VR-AdaGrad and VR-RMSProp as the other two special examples of our general algorithm because they have relatively smaller lower bound of the strong convexity and their proximal functions are even more complicated. We use the constant step size schedule for these algorithms as they are self-adaptive. The training loss and the testing Top-1 accuracy of VR-AdaGrad, VR-RMSProp, and their original algorithms on all the three datasets are plotted in Figure \ref{fig: Training and Testing}. As can be observed, the variance reduced algorithms converge faster than their original algorithms and their best testing Top-1 accuracy is also comparable or even higher, proving the effectiveness of variance reduction. We emphasize that the experiments are not designed to pursue the state-of-the-art performances, but to show that variance reduction can work well with any adaptive proximal functions and contribute to faster training, even if the proximal functions have very small strong convexity lower bound.

Finally, we also show that algorithms with weaker convexity need a larger batch size $B_t$ to converge fast, which corresponds to our claims in Section \ref{subsec: SVRAMD_convergence} and \ref{subsec: SVRAMD_convergence_PL}. We fix the mini batch sizes $b_t$ in VR-AdaGrad and VR-RMSProp to be the same as in the previous experiments and gradually increase the batch size ratio $r$. The baseline ratios of ProxSVRG+, which has a large strong convexity lower bound, are provided in Appendix \ref{sec: appendixD} for the readers' reference and the training performances of VR-AdaGrad and VR-RMSProp with different $r$ are shown in Figure \ref{fig: Batch_Size}. Note that ProxSVRG+ only needs a small ratio ($r=4$ or $r=8$) to be faster than SGD \citep{li2018Simple}, but for  VR-RMSProp, even when $r = 16$, the algorithms still do not converge faster than RMSProp, proving that algorithms with weaker convexity need larger batch sizes $B_t$ to show the effectiveness of variance reduction.

\section{Conclusions}
\label{sec: conclusion}

In this work, we generalize the convergence results of variance reduced algorithms to almost all adaptive stochastic mirror descent algorithms by proposing a general  adaptive extension of ProxSVRG+. We prove that the variance reduction technique contributes to the same level of improvement in the convergence rates of adaptive mirror descent algorithms under a very mild assumption, both in the general nonsmooth nonconvex problems and the PL-conditioned problems. In particular, our theory can be applied to proving the soundness of using time-varying step sizes in the training process of ProxSVRG+, and the effectiveness of applying variance reduction to adaptive algorithms such as AdaGrad and RMSProp. A potential future work direction is whether our Assumption \ref{assumption: 1} can still be replaced by weaker conditions, such as the CHEF conditions \citet{lei2019adaptivity} have mentioned in the convex case. Another interesting direction is whether the SFO complexity can be further improved with the help of additional algorithmic components, for example, the nested variance reduction technique by \citet{Zhou2018Stochastic} or momentum accelerations.

\section{Declarations}

The authors have no conflicts of interest to declare that are relevant to the content of this article.

\bibliographystyle{spbasic}      

\clearpage
\bibliography{my_ref}

\newpage
\onecolumn
\appendix
\textsc{\Large Appendix}
\section{Convergence of the Adaptive SMD Algorithm}
\label{sec: appendixA}

\begin{definition}
    \label{def: generalized_gradient_stochastic}
We define the generalized stochastic gradient at $t$ as
\begin{equation}
    \begin{aligned}
    \nonumber
    \Tilde{g}_{X,t} = \frac{1}{\alpha_t}(x_t - x_{t+1})
        \end{aligned}
\end{equation}
\end{definition}

\subsection{Auxiliary Lemmas}
\begin{lemma}
\label{lem: inner_product}
[\textbf{Lemma 1 in \citet{ghadimi2016Minibatch}}]. Let $g_t$ be the stochastic gradient in Algorithm \ref{alg: SMD algorithm} obtained at $t$ and $\Tilde{g}_{X,t}$ be defined as in Definition \ref{def: generalized_gradient_stochastic}, then
\begin{equation}
    \begin{aligned}
    \langle g_t, \Tilde{g}_{X,t} \rangle \geq m\|\Tilde{g}_{X,t}\|^2 + \frac{1}{\alpha_t}[h(x_{t+1}) - h(x_t)]
        \end{aligned}
\end{equation}
\end{lemma}
\textbf{Proof.}
By the optimality of the mirror descent update rule, it implies for any $x \in \mathcal{X}$ and $\nabla h(x_{t+1}) \in \partial h(x_{t+1})$
\begin{equation}
    \begin{aligned}
   \langle g_t + \frac{1}{\alpha_t}(\nabla \psi_t(x_{t+1})- \nabla \psi_t(x_{t})) + \nabla h(x_{t+1}), x_t- x_{t+1} \rangle \geq 0
        \end{aligned}
\end{equation}

Let $x = x_t$ in the above in equality, we get

\begin{equation}
    \begin{aligned}
   \langle g_t, x_t- x_{t+1} \rangle 
   &\geq  \frac{1}{\alpha_t} \langle \nabla \psi_t(x_{t+1})- \nabla \psi_t(x_{t}), x_{t+1} -x_t \rangle + \langle \nabla h(x_{t+1}), x_{t+1} -x_t \rangle\\
   & \geq \frac{m}{\alpha_t} \|x_{t+1} - x_t\|_2^2 + h(x_{t+1}) - h(x)
        \end{aligned}
\end{equation}

where the second inequality is due to the strong convexity of the function $\psi_t(x)$ and the convexity of $h(x)$, by noting that $x_{t} - x_{t+1} = \alpha_t \Tilde{g}_{X,t}$ , the inequality follows.
\begin{lemma}
\label{lem: diff_generalized_gradient}
Let ${g}_{X,t}, \Tilde{g}_{X,t}$ be defined as in Definition \ref{def: generalized_gradient_stochastic} and Definition \ref{def: generalized_gradient_deterministic} respectively, then
\begin{equation}
    \begin{aligned}
    \|g_{X,t} - \Tilde{g}_{X,t}\|_2\leq \frac{1}{m} \|\nabla f(x_t) - g_t\|_2
    \end{aligned}
\end{equation}
\end{lemma}
\textbf{Proof.}
By the definition of $g_{X_t}$ and $\Tilde{g}_{X,t}$,
\begin{equation}
    \begin{aligned}
    \|g_{X,t} - \Tilde{g}_{X,t}\|_2 = \frac{1}{\alpha_t}\|(x_t-x_{t+1}^+) - (x_t-x_{t+1})\|_2 = \frac{1}{\alpha_t}\|x_{t+1} - x_{t+1}^+ \|_2
    \end{aligned}
\end{equation}

Similar to Lemma \ref{lem: inner_product}, by the optimality of the mirror descent update rule, we have the following two inequalities 
\begin{equation}
    \begin{aligned}
   &\langle g_t + \frac{1}{\alpha_t}(\nabla \psi_t(x_{t+1})- \nabla \psi_t(x_{t})) +\nabla h(x_{t+1}), x- x_{t+1} \rangle \geq 0, \forall x \in \mathcal{X}, \nabla h(x_{t+1}) \in \partial h(x_{t+1})  \\
   &\langle \nabla f(x_t) + \frac{1}{\alpha_t}(\nabla \psi_t(x_{t+1}^+)- \nabla \psi_t(x_{t})) + \nabla h(x_{t+1}^+), x- x_{t+1}^+ \rangle \geq 0, \forall x \in \mathcal{X}, \nabla h(x_{t+1}^+) \in \partial h(x_{t+1}^+) 
        \end{aligned}
\end{equation}

Take $x = x_{t+1}^+$ in the first inequality and $x = x_{t+1}$ in the second one, we can get 

\begin{equation}
    \begin{aligned}
   &\langle - g_t, x_{t+1}-  x_{t+1}^+ \rangle 
   \geq  \frac{1}{\alpha_t} \langle \nabla \psi_t(x_{t+1})- \nabla \psi_t(x_{t}), x_{t+1} -x_{t+1}^+ \rangle + h(x_{t+1}) - h(x_{t+1}^+) \\
    &\langle \nabla f(x_t), x_{t+1}- x_{t+1}^+ \rangle 
   \geq  \frac{1}{\alpha_t} \langle \nabla \psi_t(x_{t+1}^+)- \nabla \psi_t(x_{t}), x_{t+1}^+ -x_{t+1} \rangle + h(x_{t+1}^+) - h(x_{t+1}) \\
        \end{aligned}
\end{equation}

Summing up the above inequalities, we can get

\begin{equation}
    \begin{aligned}
   &\langle \nabla f(x_t) - g_t, x_{t+1} - x_{t+1}^+ \rangle \\
   &\geq \frac{1}{\alpha_t} (\langle \nabla \psi_t(x_{t+1})- \nabla \psi_t(x_{t}), x_{t+1} - x_{t+1}^+ \rangle + \langle \nabla \psi_t(x_{t+1}^+)- \nabla \psi_t(x_{t}), x_{t+1}^+ -x_{t+1} \rangle) \\
   & = \frac{1}{\alpha_t} (\langle \nabla \psi_t(x_{t+1})- \nabla \psi_t(x_{t+1}^+), x_{t+1} -x_{t+1}^+\rangle )\\
   & \geq \frac{m}{\alpha_t}\|x_{t+1} - x_{t+1}^+\|_2^2
        \end{aligned}
\end{equation}

Therefore by Cauchy Schwarz inequality,

\begin{equation}
    \begin{aligned}
   \|g_t - \nabla f(x_t)\|_2 \geq \frac{m}{\alpha_t}\|x_{t+1} - x_{t+1}^+\|_2
        \end{aligned}
\end{equation}

Hence the inequality in the lemma follows.

\begin{lemma}
\label{lem: bounded_variance}
[\textbf{Lemma A.1 in \citet{lei2017Nonconvex}}]. Let $x_1, \cdots, x_M \in \mathbb{R}^d$ be an arbitrary population of $M$ vectors with the condition that
\begin{equation}
    \begin{aligned}
    \sum_{i=1}^M x_j = 0
    \end{aligned}
\end{equation}
Further let $J$ be a uniform random subset of $\{1, \cdots, M\}$ with size $m$, then
\begin{equation}
    \begin{aligned}
    \mathbb{E}[ \|\frac{1}{m} \sum_{j\in J} x_j\|^2] \leq \frac{I(m <M)}{mM}\sum_{j=1}^M \|x_j\|^2
    \end{aligned}
\end{equation}
\end{lemma}

Proof of the above general lemma can be found in \citet{lei2017Nonconvex}.

\subsection{Convergence of the Adaptive SMD Algorithm in the Nonconvex Nonsmooth Problem}
\label{subsec: convergence_adaptive_smd}
\textbf{Proof of Theorem \ref{Thm: convergence of SMD}.}
From the $L$-Lipshitz gradients and Lemma \ref{lem: inner_product}, we know that

\begin{equation}
    \begin{aligned}
        f(x_{t+1}) &\leq f(x_t) + \langle \nabla f(x_t), x_{t+1} - x_t \rangle + \frac{L}{2}\|x_{t+1} - x_t\|^2 \\
                    &= f(x_t) - \alpha_t\langle \nabla f(x_t), \Tilde{g}_{X,t} \rangle + \frac{L}{2} \alpha_t^2\|\Tilde{g}_{X,t}\|_2^2 \\
                    &= f(x_t) - \alpha_t\langle g_t, \Tilde{g}_{X,t} \rangle + \frac{L}{2}\alpha_t^2\|\Tilde{g}_{X,t}\|_2^2 + \alpha_t \langle  g_t - \nabla f(x_t), \Tilde{g}_{X,t} \rangle\\
                    &\leq f(x_t) +  \frac{L}{2}\alpha_t^2\|\Tilde{g}_{X,t}\|_2^2 - \alpha_t m \|\Tilde{g}_{X,t}\|_2^2 -[h(x_{t+1}) - h(x_t)]+  \alpha_t\langle  g_t - \nabla f(x_t), \Tilde{g}_{X,t} \rangle\\
    \end{aligned}
\end{equation}

Therefore since $F(x) = f(x) + h(x)$,  we get 
\begin{equation}
    \begin{aligned}
            F(x_{t+1})&\leq F(x_t)  - (\alpha_t m -\frac{L}{2}\alpha_t^2)\|\Tilde{g}_{X,t}\|_2^2 + \alpha_t \langle  g_t - \nabla f(x_t), \Tilde{g}_{X,t} \rangle + \alpha_t\langle  g_t - \nabla f(x_t), \Tilde{g}_{X,t} - g_{X,t}\rangle \\
                    &\leq F(x_t)  - (\alpha_t m -\frac{L}{2}\alpha_t^2)\|\Tilde{g}_{X,t}\|_2^2 +  \alpha_t \langle  g_t - \nabla f(x_t), \Tilde{g}_{X,t} \rangle + \alpha_t \| \nabla f(x_t)-g_t\|_2\|\Tilde{g}_{X,t} - g_{X,t}\|_2 \\
                    &\leq  F(x_t)  - (\alpha_t m -\frac{L}{2}\alpha_t^2)\|\Tilde{g}_{X,t}\|_2^2 +  \alpha_t \langle  g_t - \nabla f(x_t), \Tilde{g}_{X,t} \rangle + \frac{\alpha_t}{m} \| \nabla f(x_t)-g_t\|_2^2
    \end{aligned}
\end{equation}

where the second last one is a direct result from Cauchy-Schwarz inequality and the last inequality is from Lemma \ref{lem: diff_generalized_gradient}. Rearrange the above inequalities and sum up from 1 to $T$, we get

\begin{equation}
    \begin{aligned}
        \sum_{t=1}^T (\alpha_t m -\frac{L}{2}\alpha_t^2)\|\Tilde{g}_{X,t}\|_2^2 
        &\leq  \sum_{t=1}^T [F(x_t) - F(x_{t+1})] + \sum_{t=1}^T [\alpha_t \langle  g_t - \nabla f(x_t), \Tilde{g}_{X,t} \rangle +  \frac{\alpha_t}{m}\| \nabla f(x_t)-g_t\|_2^2] \\
        &= F(x_1) - F(x_{T+1})+\sum_{t=1}^T [\alpha_t \langle  g_t - \nabla f(x_t), \Tilde{g}_{X,t} \rangle +  \frac{\alpha_t}{m} \| \nabla f(x_t)-g_t\|^2] \\
        & \leq F(x_1) - F^* +\sum_{t=1}^T [\alpha_t \langle  g_t - \nabla f(x_t), \Tilde{g}_{X,t} \rangle +  \frac{\alpha_t}{m} \| \nabla f(x_t)-g_t\|^2] \\
    \end{aligned}
\end{equation}

where the last inequality is due to $F^* \leq F(x), \forall x$. Define the filtration $\mathcal{F}_t = \sigma(x_1, \cdots, x_t)$. Note that we suppose $g_t$ is an unbiased estimate of $\nabla f(x_t)$, hence $\mathbb{E}[\langle \nabla f(x_t) -g_t, g_{X,t} \rangle | \mathcal{F}_t]= 0$. Moreover, since the sampled gradients has bounded variance $\sigma^2$, hence by applying Lemma \ref{lem: bounded_variance} with $x_i = \nabla_{i \in \mathcal{I}_j} f_i(x_t) - \nabla f(x_t)$

\begin{equation}
    \begin{aligned}
        \mathbb{E}[\|\nabla f(x_t) -g_t\|^2] \leq \frac{\sigma^2}{b_t} I(b_t < n)
    \end{aligned}
\end{equation}
where $I$ is the indicator function. Since the final $x_{t^*}$ is uniformly sampled from all $\{x_{t}\}_{t=1}^T$, therefore
\begin{equation}
    \begin{aligned}
    \mathbb{E} [\|\Tilde{g}_{X,t^*}\|_2^2] = \mathbb{E}[\mathbb{E}[\|\Tilde{g}_{X,t^*}\|_2^2|t^*]] = \frac{1}{T}\sum_{t=1}^T \mathbb{E}[\|\Tilde{g}_{X,t}\|_2^2]
    \end{aligned}
\end{equation}
Therefore when $\alpha_t, b_t$ are constants, the average can be found as

\begin{equation}
    \begin{aligned}
        T (\alpha_t m -\frac{L}{2}\alpha_t^2) \mathbb{E} (\|\Tilde{g}_{X,t^*}\|_2^2)
        & \leq F(x_1) - F^*+\sum_{t=1}^T \frac{\alpha_t}{m} \mathbb{E} [\|\nabla f(x_t) -g_t\|_2^2]\\
        &= \Delta_F + T \frac{\alpha_t\sigma^2}{mb_t} I(b_t < n)
    \end{aligned}
\end{equation}

where we define $\Delta_F =  F(x_1) - F^*$. Take $\alpha_t = \frac{m}{L}$, then $\alpha_t m - \frac{L}{2}\alpha_t^2 = \frac{m^2}{2L}$ and 

\begin{equation}
    \begin{aligned}
        \mathbb{E} (\|\Tilde{g}_{X,t^*}\|_2^2)
        &\leq \frac{2\Delta_FL}{m^2 T} + \frac{2\sigma^2}{b_t m^2} I(b_t < n)
    \end{aligned}
\end{equation}

Also by Lemma \ref{lem: diff_generalized_gradient}, the difference between $g_{X,t^*}$ and $\Tilde{g}_{X, t^*}$ are bounded, hence

\begin{equation}
    \begin{aligned}
     \mathbb{E} [\|{g}_{X,t^*}\|_2^2] &\leq 2\mathbb{E} [\|\Tilde{g}_{X,t^*}\|_2^2] + 2\mathbb{E} [\|g_{X,t^*} - \Tilde{g}_{X,t^*}\|_2^2]\\
     &\leq \frac{4\Delta_FL}{m^2 T} + \frac{4\sigma^2}{b_t m^2} I(b_t < n) + \frac{2\sigma^2}{ b_t m^2} I(b_t < n) \\
     &= \frac{4\Delta_FL}{m^2 T} + \frac{6\sigma^2}{b_t m^2} I(b_t < n)
    \end{aligned}
\end{equation}

Take $b_t = n \wedge ({12\sigma^2}/m^2\epsilon), T = 1 \vee (8\Delta_F L / m^2 \epsilon)$ as in the theorem, the expectation is 

\begin{equation}
    \begin{aligned}
     \mathbb{E} [\|{g}_{X,t^*}\|_2^2] 
     &\leq \frac{4\Delta_FL}{m^2 T} + \frac{6\sigma^2}{b_t m^2} I(b_t < n) \\
     &\leq \frac{\epsilon}{2} + \frac{\epsilon}{2}  = \epsilon
    \end{aligned}
\end{equation}

Therefore since one iteration takes $b_t$ stochastic gradient computations, the total number of stochastic gradient computations is 

\begin{equation}
    \begin{aligned}
     Tb_t \leq \frac{8\Delta_F L}{m^2\epsilon}b_t  + b_t 
           = O \left(\frac{n}{\epsilon} \wedge \frac{\sigma^2}{\epsilon^2} + n \wedge \frac{\sigma^2}{\epsilon}\right)
    \end{aligned}
\end{equation}

\subsection{Convergence of the Adaptive SMD Algorithm under the P-L condition}

\textbf{Proof of Theorem \ref{Thm: convergence of SMD PL}}
By the proof in \ref{subsec: convergence_adaptive_smd}, we have 

\begin{equation}
    \begin{aligned}
            F(x_{t+1}) 
            &\leq  F(x_t)  - (\alpha_t m -\frac{L}{2}\alpha_t^2)\|\Tilde{g}_{X,t}\|_2^2 +  \alpha_t \langle  g_t - \nabla f(x_t), \Tilde{g}_{X,t} \rangle + \frac{\alpha_t}{m} \| \nabla f(x_t)-g_t\|_2^2 \\
    \end{aligned}
\end{equation}

Take expectation on both sides, we know that

\begin{equation}
    \begin{aligned}
        \mathbb{E} [F(x_{t+1})] 
            &\leq  \mathbb{E} [F(x_t)]  - (\alpha_t m -\frac{L}{2}\alpha_t^2) \mathbb{E} [\|\Tilde{g}_{X,t}\|_2^2] + \frac{\alpha_t}{m} \mathbb{E} [\| \nabla f(x_t)-g_t\|_2^2] \\
    \end{aligned}
\end{equation}

Since

\begin{equation}
    \begin{aligned}
     \mathbb{E} [\|{g}_{X,t^*}\|_2^2] &\leq 2\mathbb{E} [\|\Tilde{g}_{X,t^*}\|_2^2] + 2\mathbb{E} [\|g_{X,t^*} - \Tilde{g}_{X,t^*}\|_2^2]
    \end{aligned}
\end{equation}

Hence the inequality becomes

\begin{equation}
    \begin{aligned}
        \mathbb{E} [F(x_{t+1})] 
            &\leq  \mathbb{E} [F(x_t)]  - (\alpha_t m -\frac{L}{2}\alpha_t^2) (\frac{1}{2}\mathbb{E} [\|{g}_{X,t}\|_2^2]-  \mathbb{E} [\| \nabla f(x_t)-g_t\|_2^2]) + \frac{\alpha_t}{m} \mathbb{E} [\| \nabla f(x_t)-g_t\|_2^2] \\
            &\leq  \mathbb{E} [F(x_t)]  - (\frac{\alpha_t m}{2} -\frac{L}{4}\alpha_t^2) \mathbb{E} [\|{g}_{X,t}\|_2^2] + (\frac{\alpha_t}{m}+\alpha_t m -\frac{L}{2}\alpha_t^2) \mathbb{E} [\| \nabla f(x_t)-g_t\|_2^2] \\
            &\leq  \mathbb{E} [F(x_t)]  - \mu({\alpha_t m} -\frac{L}{2}\alpha_t^2) ( \mathbb{E} [F(x_t)] - F^*) + (\frac{\alpha_t}{m}+\alpha_t m -\frac{L}{2}\alpha_t^2) \mathbb{E} [\| \nabla f(x_t)-g_t\|_2^2] \\
    \end{aligned}
\end{equation}

Take $\alpha_t = m/L$ and minus $F(x)^*$ on both sides, we get
\begin{equation}
    \begin{aligned}
        \mathbb{E} [F(x_{t+1})] - F^*
            &\leq  (1 - \mu({\alpha_t m} -\frac{L}{2}\alpha_t^2) ( \mathbb{E} [F(x_t)] - F^*) + (\frac{\alpha_t}{m}+\alpha_t m -\frac{L}{2}\alpha_t^2) \mathbb{E} [\| \nabla f(x_t)-g_t\|_2^2] \\
            &=  (1 - \mu \frac{m^2}{2L}) ( \mathbb{E} [F(x_t)] - F^*) + (\frac{1}{L}+\frac{m^2}{2L}) \mathbb{E} [\| \nabla f(x_t)-g_t\|_2^2] \\
            &=  (1 - \mu \frac{m^2}{2L}) ( \mathbb{E} [F(x_t)] - F^*) + (\frac{1}{L}+\frac{m^2}{2L}) \frac{\sigma^2}{b_t} I(b_t < n)\\
    \end{aligned}
\end{equation}

Let $\gamma = 1 - \frac{\mu m^2}{2L}$, since $m^2\mu / L \leq \frac{1}{\sqrt{n}}$, $\gamma \in (0, 1)$, divide by $\gamma^{t+1}$ on both sides, we get
\begin{equation}
    \begin{aligned}
        \frac{\mathbb{E} [F(x_{t+1})] - F^*}{\gamma^{t+1}}
            &\leq  \frac{ \mathbb{E} [F(x_t)] - F^*}{\gamma^{t}} +  \frac{(\frac{1}{L}+\frac{m^2}{2L})}{\gamma^{t+1}} \frac{\sigma^2}{b_t} I(b_t < n)\\
    \end{aligned}
\end{equation}

Take summation with respect to the loop parameter $t$ from $t=1$ to $t = T$,  assume that $b_t$ is a constant, the inequality becomes

\begin{equation}
    \begin{aligned}
        {\mathbb{E} [F(x_{T+1})] - F^*}
            &\leq  \gamma^T \Delta_F +  \gamma^{T} \sum_{t=1}^T \frac{(\frac{1}{L}+\frac{m^2}{2L})}{\gamma^{t}} \frac{\sigma^2}{b_t} I(b_t < n)\\
            &\leq \gamma^T \Delta_F +   (\frac{1}{L}+\frac{m^2}{2L})\frac{1-\gamma^T}{1-\gamma}\frac{\sigma^2}{b_t} I(b_t < n)\\
            &\leq \gamma^T \Delta_F +   (\frac{1}{L}+\frac{m^2}{2L})\frac{2L}{\mu m^2}\frac{\sigma^2}{b_t} I(b_t < n)\\
            &= \gamma^T \Delta_F +   (\frac{1}{m^2}+1)\frac{1}{\mu}\frac{\sigma^2}{b_t} I(b_t < n)\\
    \end{aligned}
\end{equation}

Therefore when taking $ T = 1 \vee (\log \frac{2\Delta_F}{\epsilon})/({\log \frac{1}{\gamma}}) = O(\log \frac{2\Delta_F}{\epsilon}/\mu)$, $b_t = n \wedge \frac{2(1+m^2)\sigma^2}{\epsilon m^2\mu}$. Then the total number of stochastic gradient computations is

\begin{equation}
    \begin{aligned}
    Tb &= O((n \wedge \frac{\sigma^2}{\mu\epsilon})({ \frac{1}{\mu}\log \frac{1}{\epsilon}}))\\
       &= O \left((\frac{n}{\mu} \wedge \frac{\sigma^2}{\mu^2\epsilon})\log \frac{1}{\epsilon} \right) 
    \end{aligned}
\end{equation}

\section{Convergence of SVRAMD in the nonconvex nonsmooth problem}
\label{sec: appendixB}

Recall the Algorithm \ref{alg: ASMDVR algorithm} in the algorithm section, similarly define 
\begin{definition}
    We define the variance reduced generalized gradient mapping as
    \label{def: VR_generalized_gradient_stochastic}
    \begin{equation}
    \begin{aligned}
    \Tilde{g}_{Y,k}^t = \frac{1}{\alpha_t}(y_k^t - y_{k+1}^t) \nonumber
        \end{aligned}
\end{equation}
\end{definition}

\begin{definition}
       \label{def: VR_generalized_gradient_deterministic}
   We also define its corresponding term when the algorithm uses non-stochastic full batch gradient
\begin{equation}
    \begin{aligned}
    {g}_{Y,k}^t = \frac{1}{\alpha_t}(y_k^t - y_{k+1}^{t+}), \text{ when }  y_{k+1}^{t+} = \text{argmin}_{y}\{\alpha_t \langle \nabla f(y_k^t), y \rangle + \alpha_t h(x)+ B_{\psi_{tk}}(y, y_k^t)\}
        \end{aligned}
\end{equation}
\end{definition}


\subsection{Auxiliary Lemmas}
\begin{lemma}
\label{lem: VR_inner_product}
Let $v^t_k$ be defined as in Algorithm \ref{alg: ASMDVR algorithm} and $\Tilde{g}_{Y,k}^t$ be defined as in Definition \ref{def: VR_generalized_gradient_stochastic}, then we have
\begin{equation}
    \begin{aligned}
    \langle v_k^t, \Tilde{g}_{Y,k}^t \rangle \geq m\| \Tilde{g}_{Y,k}^t\|^2 + \frac{1}{\alpha_t}[h(y^t_{k+1}) - h(y^t_{k})]
        \end{aligned}
\end{equation}
\end{lemma}
\textbf{Proof.} 
The proof of this inequality is similar to that of Lemma \ref{lem: inner_product}. By the optimality of the mirror descent update rule, it implies for any $y \in \mathcal{X}, \nabla h(y_{k+1}^t) \in \partial h(y_{k+1}^t)$,
\begin{equation}
    \begin{aligned}
   \langle v_k^t + \frac{1}{\alpha_t}(\nabla \psi_{tk}(y_{k+1}^t)- \nabla \psi_{tk}(y_{k}^t)) + \nabla h(y_{k+1}^t), y- y^t_{k+1} \rangle \geq 0
        \end{aligned}
\end{equation}

Let $x = y_{k}^t$ in the above in equality, we get

\begin{equation}
    \begin{aligned}
   \langle v_k^t, y^t_{k} - y^t_{k+1} \rangle 
   &\geq  \frac{1}{\alpha_t} \langle \nabla \psi_{tk}(y_{k+1}^t)- \nabla \psi_{tk}(y_{k}^t), y^t_{k+1} - y^t_{k} \rangle + \langle \nabla h(y_{k+1}^t), y^t_{k+1} - y^t_{k} \rangle \\
   & \geq \frac{m}{\alpha_t} \|y^t_{k+1} - y^t_{k}\|_2^2 + [h(y^t_{k+1}) - h(y^t_{k})]
        \end{aligned}
\end{equation}

where the second inequality is due to the $m$-strong convexity of the function $\psi_{tk}(x)$ and the convexity of $h$. Note from the definition that $ y^t_{k} - y^t_{k+1}= \alpha_t \Tilde{g}_{Y,k}^t$ , the inequality follows.

\begin{lemma}
\label{lem: VR_diff_generalized_gradient}
Let ${g}_{Y,k}^t, \Tilde{g}_{Y,k}^t$ be defined as in Definition \ref{def: VR_generalized_gradient_stochastic} and Definition \ref{def: VR_generalized_gradient_deterministic} respectively, then we have
\begin{equation}
    \begin{aligned}
    \|\Tilde{g}_{Y,k}^t - {g}_{Y,k}^t\|_2 \leq \frac{1}{m} \|\nabla f(y_k^t)-v_k^t\|_2
    \end{aligned}
\end{equation}
\end{lemma}
\textbf{Proof.}
The proof is similar to Lemma \ref{lem: diff_generalized_gradient}. By definition of $\Tilde{g}_{Y,k}^t$ and ${g}_{Y,k}^t$,
\begin{equation}
    \begin{aligned}
    \|\Tilde{g}_{Y,k}^t - {g}_{Y,k}^t\|_2 = \frac{1}{\alpha_t}\|(y_k^t -y_{k+1}^{t})  - (y_k^t -y_{k+1}^{t+}) \|_2 = \frac{1}{\alpha_t}\|y_k^t - y_{k+1}^{t+} \|_2
    \end{aligned}
\end{equation}

As in Lemma \ref{lem: VR_inner_product}, by the optimality of the mirror descent update rule, we have the following two inequalities 
\begin{equation}
    \begin{aligned}
   &\langle v_k^t + \frac{1}{\alpha_t}(\nabla \psi_{tk}(y_{k+1}^t)- \nabla \psi_{tk}(y_{k}^t)) +\nabla h(y_{k+1}^t) , y- y^t_{k+1} \rangle \geq 0, \forall y \in \mathcal{X}, \nabla h(y_{k+1}^t) \in \partial h(y_{k+1}^t)\\
   &\langle \nabla f(y_k^t) + \frac{1}{\alpha_t}(\nabla \psi_{tk}(y^{t+}_{k+1})- \nabla \psi_{tk}(y_k^t)) + \nabla h(y_{k+1}^{t+}) , y- y^{t+}_{k+1} \rangle \geq 0, \forall y \in \mathcal{X}, \nabla h(y_{k+1}^{t+1}) \in \partial h(y_{k+1}^{t+})
        \end{aligned}
\end{equation}

Take $y = y_{k+1}^{t+}$ in the first inequality and $y = y_{k+1}^{t}$ in the second one, we can get 

\begin{equation}
    \begin{aligned}
   &\langle v_k^t, y_{k+1}^{t+}- y_{k+1}^{t} \rangle 
   \geq  \frac{1}{\alpha_t} \langle \nabla \psi_{tk}(y_{k+1}^t)- \nabla \psi_{tk}(y_{k}^t), y_{k+1}^{t}- y_{k+1}^{t+}\rangle + h(y_{k+1}^t) - h(y_{k+1}^{t+})  \\
    &\langle \nabla f(y_{k}^t), y_{k+1}^{t}- y_{k+1}^{t+}\rangle 
   \geq  \frac{1}{\alpha_t} \langle \nabla \psi_{tk}(y_{k}^t)- \nabla \psi_{tk}(y_{k+1}^{t+}), y_{k+1}^{t+}- y_{k+1}^{t}\rangle  + h(y_{k+1}^{t+}) - h(y_{k+1}^{t})\\
        \end{aligned}
\end{equation}

Summing up the above inequalities, we can get

\begin{equation}
    \begin{aligned}
   &\langle v_k^t -  \nabla f(y_{k}^t), y_{k+1}^{t+}- y_{k+1}^{t} \rangle \\
   &\geq \frac{1}{\alpha_t} \langle \nabla \psi_{tk}(y_{k+1}^t)- \nabla \psi_{tk}(y_{k}^t), y_{k+1}^{t}- y_{k+1}^{t+}\rangle + \frac{1}{\alpha_t} \langle \nabla \psi_{tk}(y_{k}^t)- \nabla \psi_{tk}(y_{k+1}^{t+}), y_{k+1}^{t+}- y_{k+1}^{t}\rangle \\
   & = \frac{1}{\alpha_t} (\langle \nabla \psi_{tk}(y_{k+1}^t)- \nabla \psi_{tk}(y_{k+1}^{t+}), y_{k+1}^{t}- y_{k+1}^{t+}\rangle )\\
   & \geq \frac{m}{\alpha_t}\|y_{k+1}^{t}- y_{k+1}^{t+}\|_2^2
        \end{aligned}
\end{equation}

where the last inequality is due to the strong convexity of $\psi_{tk}(x)$. Therefore by Cauchy Schwarz inequality,

\begin{equation}
    \begin{aligned}
    \frac{1}{m} \|\nabla f(y_k^t)-v_k^t\|_2 \geq  \frac{1}{\alpha_t}\|y_{k+1}^{t}- y_{k+1}^{t+}\|_2 \geq \|\Tilde{g}_{Y,k}^t - {g}_{Y,k}^t\|_2
        \end{aligned}
\end{equation}

Hence the inequality in the lemma follows.

\begin{lemma}
\label{lem: VR_bounded_variance}
Let $\nabla f(y_{k}^t), v_{k}^t$ be the full batch gradient and the , then
\begin{equation}
    \begin{aligned}
  \mathbb{E}[\| \nabla f(y_k^t)-v_k^t\|_2^2] \leq   \frac{L^2}{b_t}  \mathbb{E}[ \|y_k^t -  x_t\|^2] + 
   \frac{I(B_t<n)\sigma^2}{B_t}
    \end{aligned}
\end{equation}
\end{lemma}
\textbf{Proof}. Note that the large batch $\mathcal{I}_j$ and the mini-batch $\Tilde{\mathcal{I}}_j$ are independent, hence

\begin{equation}
    \begin{aligned}
  &\mathbb{E}[\| \nabla f(y_k^t)-v_k^t\|_2^2] \\
  &= \mathbb{E}[\|  \frac{1}{b_t}\sum_{i \in \Tilde{\mathcal{I}}_k} (\nabla f_i(y_k^t) -  \nabla f_i(x_t)) - (\nabla f(y_k^t) - g_t)\|_2^2] \\
  &= \mathbb{E}[\|  \frac{1}{b_t}\sum_{i \in \Tilde{\mathcal{I}}_k} (\nabla f_i(y_k^t) -  \nabla f_i(x_t)) - (\nabla f(y_k^t) - \frac{1}{B_t}\sum_{i \in {\mathcal{I}}_t} \nabla  f_i(x_t) )\|_2^2] \\
  &= \mathbb{E}[\|  \frac{1}{b_t}\sum_{i \in \Tilde{\mathcal{I}}_k} (\nabla f_i(y_k^t) -  \nabla f_i(x_t)) - \nabla f(y_k^t) + \nabla  f(x_t) + \frac{1}{B_t}\sum_{i \in {\mathcal{I}}_t} (\nabla  f_i(x_t) -  \nabla  f(x_t)) \|_2^2] \\
  &= \mathbb{E}[\|  \frac{1}{b_t}\sum_{i \in \Tilde{\mathcal{I}}_k} (\nabla f_i(y_k^t) -  \nabla f_i(x_t)) - \nabla f(y_k^t) + \nabla  f(x_t)\|_2^2 + \mathbb{E} \|\frac{1}{B_t}\sum_{i \in {\mathcal{I}}_t} (\nabla  f_i(x_t) -  \nabla  f(x_t)) \|_2^2] \\
  &=  \mathbb{E}[\|  \frac{1}{b_t}\sum_{i \in \Tilde{\mathcal{I}}_k} (\nabla f_i(y_k^t) - \nabla f(y_k^t) ) - ( \nabla f_i(x_t) - \nabla  f(x_t))\|_2^2 + \mathbb{E} \|\frac{1}{B_t}\sum_{i \in {\mathcal{I}}_t} (\nabla  f_i(x_t) -  \nabla  f(x_t)) \|_2^2] \\
   & \leq \mathbb{E}[\|  \frac{1}{b_t}\sum_{i \in \Tilde{\mathcal{I}}_k} (\nabla f_i(y_k^t) - \nabla f(y_k^t) )-( \nabla f_i(x_t) - \nabla  f(x_t))\|_2^2 + 
   \frac{I(B_t<n)\sigma^2}{B_t} \\
   &=  \frac{1}{b_t^2} \mathbb{E}[\sum_{i \in \Tilde{\mathcal{I}}_k} \|\nabla f_i(y_k^t) -  \nabla f_i(x_t)) - \nabla f(y_k^t) + \nabla  f(x_t) \|^2] + 
   \frac{I(B_t<n)\sigma^2}{B_t} \\
   & \leq  \frac{1}{b_t^2}  \mathbb{E}[\sum_{i \in \Tilde{\mathcal{I}}_k} \|\nabla f_i(y_k^t) -  \nabla f_i(x_t))\|^2] + 
   \frac{I(B_t<n)\sigma^2}{B_t} \\ 
   & \leq  \frac{L^2}{b_t}  \mathbb{E}[ \|y_k^t -  x_t\|^2] + 
   \frac{I(B_t<n)\sigma^2}{B_t} \\ 
    \end{aligned}
\end{equation}

where the fourth equality is because of the independence between $\mathcal{I}_j$ and $\Tilde{\mathcal{I}}_j$. The first and the second inequalities are by Lemma \ref{lem: bounded_variance}. The third inequality follows from $\mathbb{E}[\|x - \mathbb{E}(x)\|^2] = \mathbb{E}[\|x\|^2] $ and the last inequality follows from the $L$-smoothness of $f(x)$

\subsection{Main Proof}
\textbf{Proof of Theorem \ref{Thm: convergence of SMD+VR}}.
From the $L$-Lipshitz gradients and Lemma \ref{lem: VR_inner_product}, we know that

\begin{equation}
    \begin{aligned}
        f(y_{k+1}^t) &\leq f(y_k^t) + \langle \nabla f(y_k^t), y_{k+1}^t - y_k^t  \rangle + \frac{L}{2}\|y_{k+1}^t - y_k^t\|^2 \\
                    &= f(y_k^t) - \alpha_t\langle \nabla f(y_k^t), \Tilde{g}_{Y,k}^t \rangle + \frac{L}{2} \alpha_t^2\|\Tilde{g}_{Y,k}^t\|_2^2 \\
                    &= f(y_k^t) - \alpha_t\langle v_k^t, \Tilde{g}_{Y,k}^t \rangle + \frac{L}{2}\alpha_t^2\|\Tilde{g}_{Y,k}^t \|_2^2 + \alpha_t \langle v_k^t-\nabla f(y_k^t), \Tilde{g}_{Y,k}^t \rangle\\
                    &\leq f(y_k^t) +  \frac{L}{2}\alpha_t^2\|\Tilde{g}_{Y,k}^t\|_2^2 - \alpha_t m \|\Tilde{g}_{Y,k}^t\|_2^2 +  \alpha_t \langle v_k^t-\nabla f(y_k^t), \Tilde{g}_{Y,k}^t \rangle - [h(y^t_{k+1}) - h(y^t_{k})]\\
      \end{aligned}
\end{equation}
Since $F(x) = f(x) + h(x)$, we can get
\begin{equation}
    \begin{aligned}
                F(y_{k+1}^t)    
                &= F(y_k^t) - (\alpha_t m -\frac{L}{2}\alpha_t^2)\|\Tilde{g}_{Y,k}^t\|_2^2 + \alpha_t \langle v_k^t -  \nabla f(y_k^t), {g}_{Y,k}^t \rangle + \alpha_t \langle v_k^t - \nabla f(y_k^t), \Tilde{g}_{Y,k}^t - {g}_{Y,k}^t\rangle \\
                &\leq F(y_k^t) - (\alpha_t m -\frac{L}{2}\alpha_t^2)\|\Tilde{g}_{Y,k}^t\|_2^2 + \alpha_t \langle v_k^t -  \nabla f(y_k^t), {g}_{Y,k}^t \rangle + \alpha_t \| \nabla f(y_k^t)-v_k^t\|_2\|\Tilde{g}_{Y,k}^t - {g}_{Y,k}^t\|_2 \\
                &\leq  F(y_k^t) - (\alpha_t m -\frac{L}{2}\alpha_t^2)\|\Tilde{g}_{Y,k}^t\|_2^2 + \alpha_t \langle  v_k^t -  \nabla f(y_k^t), {g}_{Y,k}^t \rangle + \frac{\alpha_t}{m} \| \nabla f(y_k^t)-v_k^t\|_2^2 \\
    \end{aligned}
\end{equation}

where the second last inequality is from Cauchy Schwartz inequality and the last inequality is from Lemma \ref{lem: VR_diff_generalized_gradient}. Define the filtration $\mathcal{F}_{k}^t = \sigma(y_1^1, \cdots y_{K+1}^1, y_1^2, \cdots, y_{K+1}^2, \cdots, y_1^t, \cdots, y_{k}^t)$. Note that $\mathbb{E}[\langle \nabla f(y_k^t)-v_k^t, {g}_{Y,k}^t \rangle|F_{k}^t] = 0$. Take expectation on both sides and use Lemma \ref{lem: VR_bounded_variance}, we get 
\begin{equation}
    \begin{aligned}
    \label{eqn: main_inequality}
            \mathbb{E}[F(y_{k+1}^t)]    
                &\leq \mathbb{E}[F(y_k^t)] - ( \frac{m}{\alpha_t} -\frac{L}{2}) \mathbb{E}[\|y_{k+1}^t - y_k^t\|_2^2] + \frac{L^2 \alpha_t}{b_t m}  \mathbb{E}[ \|y_k^t -  x_t\|^2] + \frac{\alpha_t I(B_t<n)\sigma^2}{mB_t}\\
                &\leq \mathbb{E}[F(y_k^t)] - (\frac{ m}{2\alpha_t} -\frac{L}{4}) \mathbb{E}[\|y_{k+1}^t - y_k^t\|_2^2] + \frac{L^2 \alpha_t}{b_t m}  \mathbb{E}[ \|y_k^t -  x_t\|^2] + \frac{\alpha_t I(B_t<n)\sigma^2}{mB_t}\\
                & \qquad  + (\frac{ m}{2\alpha_t} -\frac{L}{4}) \frac{\alpha_t^2L^2}{m^2b_t}\mathbb{E}[ \|y_k^t -  x_t\|^2]  + (\frac{ m}{2\alpha_t} -\frac{L}{4}) \frac{\alpha_t^2 I(B_t<n)\sigma^2}{m^2B_t} - (\frac{ m}{4\alpha_t} -\frac{L}{8}) \mathbb{E}[\|y_{k+1}^{t+} - y_k^t\|_2^2] \\
                &= \mathbb{E}[F(y_k^t)] - (\frac{ m}{2\alpha_t} -\frac{L}{4}) \mathbb{E}[\|y_{k+1}^t - y_k^t\|_2^2] - (\frac{ m}{4\alpha_t} -\frac{L}{8}) \mathbb{E}[\|y_{k+1}^{t+} - y_k^t\|_2^2] \\
                & \qquad +   (\frac{3L^2 \alpha_t}{2b_t m}- \frac{\alpha_t^2 L^3}{4m^2b_t})\mathbb{E}[ \|y_k^t -  x_t\|^2] + (\frac{3\alpha_t}{2m}- \frac{\alpha_t^2L}{4m^2})\frac{ I(B_t<n)\sigma^2}{B_t} \\
                &= \mathbb{E}[F(y_k^t)] - (\frac{ m}{2\alpha_t} -\frac{L}{4}) \mathbb{E}[\|y_{k+1}^t - y_k^t\|_2^2] - (\frac{ m\alpha_t}{4} -\frac{L\alpha_t^2}{8}) \mathbb{E}[\|g^t_{Y,k}\|_2^2] \\
                & \qquad +   (\frac{3L^2 \alpha_t}{2b_t m}- \frac{\alpha_t^2 L^3}{4m^2b_t})\mathbb{E}[ \|y_k^t -  x_t\|^2] + (\frac{3\alpha_t}{2m}- \frac{\alpha_t^2L}{4m^2})\frac{ I(B_t<n)\sigma^2}{B_t}
                \\
    \end{aligned}
\end{equation}

where the second inequality uses the fact that by Lemma \ref{lem: VR_bounded_variance}
\begin{equation}
    \begin{aligned}
 \mathbb{E}[\|y_{k+1}^{t+} - y_k^t\|_2^2] &= \alpha_t ^2 \mathbb{E} [\|{g}_{Y,k}^t\|_2^2] \leq 2\alpha_t ^2 \mathbb{E} [\|\Tilde{g}_{Y,k}^t\|_2^2] + 2\alpha_t ^2 \mathbb{E} [\|{g}_{Y,k}^t - \Tilde{g}_{Y,k}^t\|_2^2] \\
&\leq 2\alpha_t ^2 \mathbb{E} [\|\Tilde{g}_{Y,k}^t\|_2^2] + \frac{2\alpha_t ^2 }{m^2}\mathbb{E} [\|\nabla f(y_k^t)-v_k^t\|_2^2] \\
&\leq 2\mathbb{E} [\|y_{k+1}^t - y_k^t\|_2^2] + \frac{2\alpha_t ^2 }{m^2}(\frac{L^2}{b_t}  \mathbb{E}[ \|y_k^t -  x_t\|^2] + \frac{I(B_t<n)\sigma^2}{B_t})\\
&= 2\mathbb{E} [\|y_{k+1}^t - y_k^t\|_2^2] + \frac{2\alpha_t ^2 L^2 }{m^2b_t}  \mathbb{E}[ \|y_k^t -  x_t\|^2] + \frac{2I(B_t<n)\sigma^2\alpha_t^2}{B_tm^2}\\
    \end{aligned}
\end{equation}

Since by Young's inequality, we know that 

\begin{equation}
    \begin{aligned}
    \| y_{k+1}^t - x_t \| \leq (1 + \frac{1}{p})\| y_k^t - x_t\|_2^2 +  (1 + p)\|y_{k+1}^{t} - y_k^t\|_2^2, \forall p \in \mathbb{R}
    \end{aligned}
\end{equation}

Hence substitute into equation \ref{eqn: main_inequality}, we can get 

\begin{equation}
    \begin{aligned}
            \mathbb{E}[F(y_{k+1}^t)]    
                &\leq \mathbb{E}[F(y_k^t)] - (\frac{ m}{2\alpha_t} -\frac{L}{4})\mathbb{E}(\frac{\|y^t_{k+1} - x_t\|_2^2}{1+p}- \frac{\|y^t_{k} - x_t\|_2^2}{p}) - (\frac{ m\alpha_t}{4} -\frac{L\alpha_t^2}{8}) \mathbb{E}[\|g^t_{Y,k}\|_2^2] \\
                & \qquad +   (\frac{3L^2 \alpha_t}{2b_t m}- \frac{\alpha_t^2 L^3}{4m^2b_t})\mathbb{E}[ \|y_k^t -  x_t\|^2] + (\frac{3\alpha_t}{2m}- \frac{\alpha_t^2L}{4m^2})\frac{ I(B_t<n)\sigma^2}{B_t}\\
               &\leq \mathbb{E}[F(y_k^t)] - (\frac{ m}{2\alpha_t} -\frac{L}{4})\mathbb{E}(\frac{\|y^t_{k+1} - x_t\|_2^2}{1+p}) - (\frac{ m\alpha_t}{4} -\frac{L\alpha_t^2}{8}) \mathbb{E}[\|g^t_{Y,k}\|_2^2] \\
               &\qquad  +   (\frac{3L^2 \alpha_t}{2b_t m}- \frac{\alpha_t^2 L^3}{4m^2b_t} + \frac{ m}{2\alpha_tp} -\frac{L}{4p})\mathbb{E}[ \|y_k^t -  x_t\|^2] + (\frac{3\alpha_t}{2m}- \frac{\alpha_t^2L}{4m^2})\frac{ I(B_t<n)\sigma^2}{B_t}
                \\
    \end{aligned}
\end{equation}

Let $p = 2k-1$ and take summation with respect to the inner loop parameter $k$, we can get

\begin{equation}
    \begin{aligned}
            &\mathbb{E}[F(x^{t+1})]    \\
               &\leq \mathbb{E}[F(x^t)] - \sum_{k=1}^K (\frac{ m}{2\alpha_t (2k)} -\frac{L}{4(2k)})\mathbb{E}({\|y^t_{k+1} - x_t\|_2^2}) - \sum_{k=1}^K (\frac{ m\alpha_t}{4} -\frac{L\alpha_t^2}{8}) \mathbb{E}[\|g^t_{Y,k}\|_2^2] \\
               & \qquad +  \sum_{k=1}^K (\frac{3L^2 \alpha_t}{2b_t m}- \frac{\alpha_t^2 L^3}{4m^2b_t} + \frac{ m}{2\alpha_t (2k-1)} -\frac{L}{4(2k-1)})\mathbb{E}[ \|y_k^t -  x_t\|^2] + \sum_{k=1}^K (\frac{3\alpha_t}{2m}- \frac{\alpha_t^2L}{4m^2})\frac{ I(B_t<n)\sigma^2}{B_t}        \\
               &\leq \mathbb{E}[F(x^t)] - \sum_{k=1}^{K-1} (\frac{ m}{2\alpha_t(2k)} -\frac{L}{4(2k)})\mathbb{E}({\|y^t_{k+1} - x_t\|_2^2}) - \sum_{k=1}^K (\frac{ m\alpha_t}{4} -\frac{L\alpha_t^2}{8}) \mathbb{E}[\|g^t_{Y,k}\|_2^2] \\
               & \qquad  +  \sum_{k=2}^K (\frac{3L^2 \alpha_t}{2b_t m}- \frac{\alpha_t^2 L^3}{4m^2b_t} + \frac{ m}{2\alpha_t (2k-1)} -\frac{L}{4(2k-1)})\mathbb{E}[ \|y_k^t -  x_t\|^2] + \sum_{k=1}^K (\frac{3\alpha_t}{2m}- \frac{\alpha_t^2L}{4m^2})\frac{ I(B_t<n)\sigma^2}{B_t}      \\
              &\leq \mathbb{E}[F(x^t)] - \sum_{k=1}^K (\frac{ m\alpha_t}{4} -\frac{L\alpha_t^2}{8}) \mathbb{E}[\|g^t_{Y,k}\|_2^2]  +  \sum_{k=1}^K (\frac{3\alpha_t}{2m}- \frac{\alpha_t^2L}{4m^2})\frac{ I(B_t<n)\sigma^2}{B_t}\\ 
              & \qquad +  \sum_{k=1}^{K-1} (\frac{3L^2 \alpha_t}{2b_t m}- \frac{\alpha_t^2 L^3}{4m^2b_t} + \frac{ m}{2\alpha_t(2k+1)} -\frac{L}{4(2k+1)} - (\frac{ m}{2\alpha_t(2k)} -\frac{L}{4(2k)}))\mathbb{E}[ \|y_k^t -  x_t\|^2] \\
              &= \mathbb{E}[F(x^t)] - \sum_{k=1}^K (\frac{ m\alpha_t}{4} -\frac{L\alpha_t^2}{8}) \mathbb{E}[\|g^t_{Y,k}\|_2^2]  +  \sum_{k=1}^K (\frac{3\alpha_t}{2m}- \frac{\alpha_t^2L}{4m^2})\frac{ I(B_t<n)\sigma^2}{B_t}\\ 
              & \qquad +  \sum_{k=1}^{K-1} (\frac{3L^2 \alpha_t}{2b_t m}- \frac{\alpha_t^2 L^3}{4m^2b_t} + (\frac{L}{4} - \frac{m}{2\alpha_t}) (\frac{1}{2k(2k+1)}))\mathbb{E}[ \|y_k^t -  x_t\|^2] \\
    \end{aligned}
\end{equation}

where the second inequality is due to the fact that $x_t = y_1^t$ and $\|x_{t+1} - x_t\| > 0$. Take $\alpha_t = {m}/{L}$

\begin{equation}
    \begin{aligned}
            \mathbb{E}[F(x^{t+1})]    
              &\leq \mathbb{E}[F(x^t)] - \sum_{k=1}^K \frac{ m^2}{8L}  \mathbb{E}[\|g^t_{Y,k}\|_2^2]  +  \sum_{k=1}^K (\frac{5}{4L})\frac{ I(B_t<n)\sigma^2}{B_t}
               +  \sum_{k=1}^{K-1} (\frac{5L}{4b_t} - \frac{L}{8k(2k+1)})\mathbb{E}[ \|y_k^t -  x_t\|^2] \\
              &\leq \mathbb{E}[F(x^t)] - \sum_{k=1}^K \frac{ m^2}{8L}  \mathbb{E}[\|g^t_{Y,k}\|_2^2]  +  \sum_{k=1}^K (\frac{5}{4L})\frac{ I(B_t<n)\sigma^2}{B_t}\\ 
    \end{aligned}
\end{equation}

where the last inequality follows from the setting $ K \leq \left \lfloor{\sqrt{b_t/20}}\right \rfloor $ and therefore 
\begin{equation}
    \begin{aligned}
            \frac{5L}{4b_t} - \frac{L}{8(K-1)(2(K-1)+1)} \leq \frac{5L}{4b_t} - \frac{L}{16 K^2} \leq 0 
    \end{aligned}
\end{equation}

Take sum with respect to the outer loop parameter $t$ and re-arrange the inequality

\begin{equation}
    \begin{aligned}
            \sum_{t=1}^T\sum_{k=1}^K \frac{ m^2}{8L}  \mathbb{E}[\|g^t_{Y,k}\|_2^2] 
            &\leq \mathbb{E}[F(x^1) - F(x^{T+1}) ] +   \sum_{t=1}^T\sum_{k=1}^K (\frac{5}{4L})\frac{ I(B_t<n)\sigma^2}{B_t} \\
            &\leq \Delta_F +   TK(\frac{5}{4L})\frac{ I(B_t<n)\sigma^2}{B_t}
    \end{aligned}
\end{equation}

Therefore when taking $B_t = n \wedge 20\sigma^2/(m^2\epsilon)$, $T = 1 \vee 16\Delta_F L/(m^2\epsilon K)$
\begin{equation}
    \begin{aligned}
       \mathbb{E}[\|g_{X,t^*}\|_2^2] 
            \leq \frac{8\Delta_FL}{m^2 TK} + \frac{ 10 I(B_t<n)\sigma^2}{B_tm^2} \leq \frac{\epsilon}{2} + \frac{\epsilon}{2} \leq \epsilon
    \end{aligned}
\end{equation}
The total number of stochastic gradient computations is 
\begin{equation}
    \begin{aligned}
    TB + TKb &= O \left((n \wedge \frac{\sigma^2}{\epsilon} + b \sqrt{b})(1 + \frac{1}{\epsilon \sqrt{b}})\right) \\
             &= O\left(n \wedge \frac{\sigma^2}{\epsilon} + b \sqrt{b} + \frac{n}{\epsilon\sqrt{b}} \wedge \frac{\sigma^2}{\epsilon^2\sqrt{b}} + \frac{b}{\epsilon}\right) \\
             &= O \left(\frac{n}{\epsilon\sqrt{b}} \wedge \frac{\sigma^2}{\epsilon^2\sqrt{b}} + \frac{b}{\epsilon}\right) 
    \end{aligned}
\end{equation}
where the last inequality is because $b^2\leq \epsilon^{-2}$ when $b \leq \epsilon^{-1}$ and $\sqrt{b} \leq \epsilon^{-1}$ when $ b \leq \epsilon^{-2}$. However,we will never let $b$ to be as large as $\epsilon^{-2}$ as it is even larger than the batch size $B_t$ and doing so will make the number of gradient computations $O(\epsilon^{-3})$, which is undesirable.

\section{Convergence of SVRAMD under the P-L Condition}
\label{sec: appendixC}
\textbf{Proof of Theorem \ref{Thm: convergence of SMD+VR PL}}. Recall the definition of the PL condition and modify the notations a little bit, we get 
\begin{equation}
    \begin{aligned}
     \exists \mu > 0, s.t. \|g_{Y,k}^t\|^2 \geq 2\mu(F(y_k^t) - F^*)
        \end{aligned}
\end{equation}
By the proof in appendix \ref{sec: appendixB}, we know that
\begin{equation}
    \begin{aligned}
            \mathbb{E}[F(y_{k+1}^t)]    
                &\leq \mathbb{E}[F(y_k^t)] - (\frac{ m}{2\alpha_t} -\frac{L}{4})\mathbb{E}(\frac{\|y^t_{k+1} - x_t\|_2^2}{1+p}- \frac{\|y^t_{k} - x_t\|_2^2}{p}) - (\frac{ m\alpha_t}{4} -\frac{L\alpha_t^2}{8}) \mathbb{E}[\|g^t_{Y,k}\|_2^2] \\
                & \qquad  +   (\frac{3L^2 \alpha_t}{2b_t m}- \frac{\alpha_t^2 L^3}{4m^2b_t})\mathbb{E}[ \|y_k^t -  x_t\|^2] + (\frac{3\alpha_t}{2m}- \frac{\alpha_t^2L}{4m^2})\frac{ I(B_t<n)\sigma^2}{B_t}\\
               &\leq \mathbb{E}[F(y_k^t)] - (\frac{ m}{2\alpha_t} -\frac{L}{4})\mathbb{E}(\frac{\|y^t_{k+1} - x_t\|_2^2}{1+p}) - (\frac{ m\alpha_t}{2} -\frac{L\alpha_t^2}{4}) \mu(\mathbb{E}[F(y_k^t)] - F^*)\\
               & \qquad +   (\frac{3L^2 \alpha_t}{2b_t m}- \frac{\alpha_t^2 L^3}{4m^2b_t} + \frac{ m}{2\alpha_tp} -\frac{L}{4p})\mathbb{E}[ \|y_k^t -  x_t\|^2] + (\frac{3\alpha_t}{2m}- \frac{\alpha_t^2L}{4m^2})\frac{ I(B_t<n)\sigma^2}{B_t}\\
    \end{aligned}
\end{equation}
Therefore when $p = 2k-1$, define $\gamma := (1-(\frac{ m\alpha_t\mu}{2} -\frac{L\alpha_t^2\mu}{4}))$, we obtain
\begin{equation}
    \begin{aligned} 
        \frac{\mathbb{E}[F(y_{k+1}^t)] -F^*}{\gamma^{k+1}}
               &\leq \frac{(\mathbb{E}[F(y_k^t)] - F^*)}{\gamma^k} - (\frac{m}{2\alpha_t \gamma^{k+1}} -\frac{L}{4\gamma^{k+1}})\mathbb{E}(\frac{\|y^t_{k+1} - x_t\|_2^2}{2k}) \\
               & \qquad  +   \frac{1}{\gamma^{k+1}}(\frac{3L^2 \alpha_t}{2b_t m}- \frac{\alpha_t^2 L^3}{4m^2b_t} + \frac{ m}{2\alpha_t(2k-1)} -\frac{L}{4(2k-1)})\mathbb{E}[ \|y_k^t -  x_t\|^2] \\
               & \qquad  + \frac{1}{\gamma^{k+1}}(\frac{3\alpha_t}{2m}- \frac{\alpha_t^2L}{4m^2})\frac{ I(B_t<n)\sigma^2}{B_t}\\
    \end{aligned}
\end{equation}
Summing up with respect to the inner loop parameter $k$, we get that
\begin{equation}
    \begin{aligned} 
        {\mathbb{E}[F(x_{t+1})] -F^*} 
               &\leq \gamma^K{(\mathbb{E}[F(x_t)] - F^*)} - \gamma^{K+1}\sum_{k=1}^K (\frac{m}{2\alpha_t \gamma^{k+1}} -\frac{L}{4\gamma^{k+1}})\mathbb{E}(\frac{\|y^t_{k+1} - x_t\|_2^2}{2k}) \\
               & \qquad +  \gamma^{K+1}\sum_{k=1}^K \frac{1}{\gamma^{k+1}}(\frac{3L^2 \alpha_t}{2b_t m}- \frac{\alpha_t^2 L^3}{4m^2b_t} + \frac{ m}{2\alpha_t(2k-1)} -\frac{L}{4(2k-1)})\mathbb{E}[ \|y_k^t -  x_t\|^2] \\
               & \qquad +  \gamma^{K+1}\sum_{k=1}^K \frac{1}{\gamma^{k+1}}(\frac{3\alpha_t}{2m}- \frac{\alpha_t^2L}{4m^2})\frac{ I(B_t<n)\sigma^2}{B_t}\\
               &= \gamma^K{(\mathbb{E}[F(x_t)] - F^*)} - \gamma^{K+1}\sum_{k=1}^K (\frac{m}{2\alpha_t \gamma^{k+1}} -\frac{L}{4\gamma^{k+1}})\mathbb{E}(\frac{\|y^t_{k+1} - x_t\|_2^2}{2k}) \\
               & \qquad +  \gamma^{K+1}\sum_{k=1}^K \frac{1}{\gamma^{k+1}}(\frac{3L^2 \alpha_t}{2b_t m}- \frac{\alpha_t^2 L^3}{4m^2b_t} + \frac{ m}{2\alpha_t(2k-1)} -\frac{L}{4(2k-1)})\mathbb{E}[ \|y_k^t -  x_t\|^2] \\
               & \qquad +  \frac{1-\gamma^K}{1-\gamma} (\frac{3\alpha_t}{2m}- \frac{\alpha_t^2L}{4m^2})\frac{ I(B_t<n)\sigma^2}{B_t}\\
    \end{aligned}
\end{equation}
By the fact that $x_t = x^t_{1}$, and $\|x_{t+1} - x_t\| > 0$, we know that
\begin{equation}
    \begin{aligned} 
    \label{eqn: PL_main_inequality}
        &{\mathbb{E}[F(x_{t+1})] -F^*} \\
               &\leq \gamma^K{(\mathbb{E}[F(x_t)] - F^*)} - \gamma^{K+1}\sum_{k=1}^{K-1} (\frac{m}{2\alpha_t \gamma^{k+1}} -\frac{L}{4\gamma^{k+1}})\mathbb{E}(\frac{\|y^t_{k+1} - x_t\|_2^2}{2k}) \\
               & \qquad +  \gamma^{K+1}\sum_{k=2}^{K} \frac{1}{\gamma^{k+1}}(\frac{3L^2 \alpha_t}{2b_t m}- \frac{\alpha_t^2 L^3}{4m^2b_t} + \frac{ m}{2\alpha_t(2k-1)} -\frac{L}{4(2k-1)})\mathbb{E}[ \|y_k^t -  x_t\|^2] \\
               & \qquad +  \frac{1-\gamma^K}{1-\gamma}  (\frac{3\alpha_t}{2m}- \frac{\alpha_t^2L}{4m^2})\frac{ I(B_t<n)\sigma^2}{B_t}\\
               &= \gamma^K{(\mathbb{E}[F(x_t)] - F^*)} + \frac{1-\gamma^K}{1-\gamma} (\frac{3\alpha_t}{2m}- \frac{\alpha_t^2L}{4m^2})\frac{ I(B_t<n)\sigma^2}{B_t} \\
               &\qquad - \gamma^{K+1}\sum_{k=1}^{K-1} (\frac{m}{2\alpha_t \gamma^{k+1}} -\frac{L}{4\gamma^{k+1}})\mathbb{E}(\frac{\|y^t_{k+1} - x_t\|_2^2}{2k}) \\
               & \qquad +  \gamma^{K+1}\sum_{k=1}^{K-1} \frac{1}{\gamma^{k+1}}(\frac{3L^2 \alpha_t}{2b_t m\gamma}- \frac{\alpha_t^2 L^3}{4m^2b_t\gamma} + \frac{ m}{2\alpha_t(2k+1)\gamma} -\frac{L}{4(2k+1)\gamma})\mathbb{E}[ \|y_k^t -  x_t\|^2] \\
               &= \gamma^K{(\mathbb{E}[F(x_t)] - F^*)} + \frac{1-\gamma^K}{1-\gamma} (\frac{3\alpha_t}{2m}- \frac{\alpha_t^2L}{4m^2})\frac{ I(B_t<n)\sigma^2}{B_t} \\
               & \qquad +  \gamma^{K+1}\sum_{k=1}^{K-1} \frac{1}{\gamma^{k+2}}(\frac{3L^2 \alpha_t}{2b_t m}- \frac{\alpha_t^2 L^3}{4m^2b_t} + \frac{ m}{2\alpha_t(2k+1)} -\frac{L}{4(2k+1)} + \frac{L\gamma}{8k} - \frac{m\gamma}{4k\alpha_t })\mathbb{E}[ \|y_k^t -  x_t\|^2] \\
              &= \gamma^K{(\mathbb{E}[F(x_t)] - F^*)} + \frac{1-\gamma^K}{1-\gamma}  (\frac{3\alpha_t}{2m}- \frac{\alpha_t^2L}{4m^2})\frac{ I(B_t<n)\sigma^2}{B_t} \\
               & \qquad +  \gamma^{K+1}\sum_{k=1}^{K-1} \frac{1}{\gamma^{k+2}}(\frac{3L^2 \alpha_t}{2b_t m}- \frac{\alpha_t^2 L^3}{4m^2b_t} - (\frac{m}{2\alpha_t} - \frac{L}{4})( \frac{\gamma}{2k}- \frac{1}{2k+1}))\mathbb{E}[ \|y_k^t -  x_t\|^2] \\
    \end{aligned}
\end{equation}

By the definition $\gamma = 1- \frac{ m\alpha_t\mu}{2} +\frac{L\alpha_t^2\mu}{4}$, we know that

\begin{equation}
    \begin{aligned} 
              \frac{\gamma}{2k}- \frac{1}{2k+1} 
              &=\frac{1}{2k(2k+1)} - \frac{ m\alpha_t\mu}{4k} +\frac{L\alpha_t^2\mu}{8k}\\
              &=\frac{1}{2k(2k+1)}- \frac{\alpha_t^2 \mu }{2k}(\frac{m}{2\alpha_t} -\frac{L}{4})\\
    \end{aligned}
\end{equation}

Therefore when taking $\alpha_t = \frac{m}{L}$ and with the assumption $L/(\mu m^2)> \sqrt{n}$, the last term in the inequality (\ref{eqn: PL_main_inequality}) is

\begin{equation}
    \begin{aligned} 
               & \gamma^{K+1}\sum_{k=1}^{K-1} \frac{1}{\gamma^{k+2}}(\frac{3L^2 \alpha_t}{2b_t m}- \frac{\alpha_t^2 L^3}{4m^2b_t} - (\frac{m}{2\alpha_t} - \frac{L}{4})( \frac{\gamma}{2k}- \frac{1}{2k+1})\mathbb{E}[ \|y_k^t -  x_t\|^2] \\
               & =  \gamma^{K+1}\sum_{k=1}^{K-1} \frac{1}{\gamma^{k+2}}(\frac{3L^2 \alpha_t}{2b_t m}- \frac{\alpha_t^2 L^3}{4m^2b_t} - (\frac{m}{2\alpha_t} - \frac{L}{4})(\frac{1}{2k(2k+1)} - \frac{\alpha_t^2 \mu }{2k}(\frac{m}{2\alpha_t} -\frac{L}{4})))\mathbb{E}[ \|y_k^t -  x_t\|^2] \\
               &= \gamma^{K+1}\sum_{k=1}^{K-1} \frac{1}{\gamma^{k+2}}(\frac{3L^2 \alpha_t}{2b_t m}- \frac{\alpha_t^2 L^3}{4m^2b_t} - (\frac{m}{2\alpha_t} - \frac{L}{4})(\frac{1}{2k(2k+1)}) + \frac{\alpha_t^2 \mu }{2k}(\frac{m}{2\alpha_t} -\frac{L}{4})^2 ))\mathbb{E}[ \|y_k^t -  x_t\|^2] \\
               &\leq \gamma^{K+1}\sum_{k=1}^{K-1} \frac{1}{\gamma^{k+2}}( \frac{5L}{4b_t} - \frac{L}{4}(\frac{1}{2k(2k+1)}) + \frac{L }{32k\sqrt{n}})\mathbb{E}[ \|y_k^t -  x_t\|^2] \\
    \end{aligned}
\end{equation}

Define $H(x) := -\frac{1}{2x(2x+1)} + \frac{1}{8x\sqrt{n}} + \frac{5}{b_t}$, $H'(x) = \frac{8x+2}{4x^2(2x+1)^2} - \frac{1}{8x^2\sqrt{n}} = \frac{1}{4x^2}(\frac{8x+2}{4x^2 + 4x+1} - \frac{1}{2\sqrt{n}}) = \frac{1}{4x^2}(\frac{2(8x+2)\sqrt{n} - (4x^2+4x+1)}{2(4x^2 + 4x+1)\sqrt{n}}) $. When $x \leq K-1 < K < \sqrt{\frac{b_t}{16}} \leq \sqrt{\frac{n}{16}}$, $\frac{8x+2}{4x^2 + 4x+1} - \frac{1}{2\sqrt{n}}\geq \frac{8K+2}{4K^2 + 4K+1} - \frac{1}{2\sqrt{n}} \geq 0 $. Therefore $H(x) \leq H(K-1) \leq \frac{5}{b_t} - \frac{14K+1}{32K(K-1)(2K-1)} \leq 0$ when $K = \lfloor \sqrt{\frac{b_t}{32}}\rfloor$. which means the inequality above is smaller than zero. Hence

\begin{equation}
    \begin{aligned} 
        &{\mathbb{E}[F(x_{t+1})] -F^*} \leq \gamma^K{(\mathbb{E}[F(x_t)] - F^*)} + \frac{1-\gamma^K}{1-\gamma} \frac{5L}{4} \frac{ I(B_t<n)\sigma^2}{B_t} \\
    \end{aligned}
\end{equation}

Therefore 

\begin{equation}
    \begin{aligned} 
        &\frac{\mathbb{E}[F(x_{t+1})] -F(x_{t})}{\gamma^{K(t+1)}} \leq \frac{(\mathbb{E}[F(x_t)] - F^*)}{\gamma^{Kt}} + \frac{1-\gamma^K}{(1-\gamma)\gamma^{K(t+1)}} (\frac{5L}{4} \frac{ I(B_t<n)\sigma^2}{B_t}) \\
    \end{aligned}
\end{equation}

Now take sum with respect to the outer loop parameter $t$ and take $B_t$ as a constant, we can get

\begin{equation}
    \begin{aligned} 
        {\mathbb{E}[F(x_{T+1})] -F^*} 
        &\leq  \gamma^{KT} ({F(x_1) - F^*}) +  \gamma^{K(T+1)}
        \sum_{t=1}^T \frac{1-\gamma^K}{(1-\gamma)\gamma^{K(t+1)}}  \frac{5L}{4} \frac{ I(B_t<n)\sigma^2}{B_t} \\
        &\leq  \gamma^{KT} \Delta_F +  \gamma^{K(T+1)}  \frac{1-\gamma^K}{1-\gamma} 
        \sum_{t=1}^T \frac{1}{\gamma^{K(t+1)}}  \frac{5L}{4} \frac{ I(B_t<n)\sigma^2}{B_t} \\
        &= \gamma^{KT} \Delta_F +  \gamma^{K(T+1)}  \frac{1-\gamma^K}{1-\gamma} 
        \sum_{t=1}^T \frac{1}{\gamma^{K(t+1)}}  \frac{5L}{4} \frac{ I(B_t<n)\sigma^2}{B_t} \\
        &= \gamma^{KT} \Delta_F +    \frac{5L I(B_t<n)\sigma^2}{4B_t}  \frac{1-\gamma^K}{1-\gamma} 
         \frac{1-\gamma^{KT}}{1-\gamma^{K}}  \\
        &= \gamma^{KT} \Delta_F +    \frac{5L I(B_t<n)\sigma^2}{4B_t} 
         \frac{1-\gamma^{KT}}{1-\gamma}  \\
    \end{aligned}
\end{equation}

Since $1 - \gamma^{KT} < 1$, $\gamma = 1- \frac{ m\alpha_t\mu}{2} +\frac{L\alpha_t^2\mu}{4} = 1 - \frac{m^2\mu}{2L} +\frac{m^2\mu}{4L}  =  1 - \frac{m^2\mu}{4L}$, hence

\begin{equation}
    \begin{aligned} 
        {\mathbb{E}[F(x_{T+1})] -F^*} 
        &\leq \gamma^{KT} \Delta_F +    \frac{5L I(B_t<n)\sigma^2}{4B_t(1-\gamma)} \\
        &=  \gamma^{KT} \Delta_F +    \frac{5 I(B_t<n)\sigma^2}{B_tm^2\mu}  \\
    \end{aligned}
\end{equation}

Therefore when taking $ T = 1 \vee (\log \frac{2\Delta_F}{\epsilon})/({K \log \frac{1}{\gamma}}) = O( (\log \frac{2\Delta_F}{\epsilon})/(K \mu))$, $B_t = n \wedge \frac{10\sigma^2}{\epsilon m^2\mu}$. Then the total number of stochastic gradient computations is

\begin{equation}
    \begin{aligned}
    TB + TKb &= O \left((n \wedge \frac{\sigma^2}{\mu \epsilon} + b \sqrt{b})({ \frac{1}{\mu\sqrt{b}}\log \frac{1}{\epsilon}}) \right)  \\
            &=  O\left((n \wedge \frac{\sigma^2}{\mu \epsilon}){ \frac{1}{\mu\sqrt{b}}\log \frac{1}{\epsilon}} + { \frac{b}{\mu}\log \frac{1}{\epsilon}} \right) 
    \end{aligned}
\end{equation}

\section{Algorithm Implementation and More Experimental Details}
\label{sec: appendixD}
\subsection{Algorithm}

\begin{algorithm}[!ht]
   \caption{Variance Reduced AdaGrad Algorithm}
   \label{alg: AdaGrad+VR}
    \begin{algorithmic}[1]
   \STATE \textbf{Input:} Number of stages $T$, initial $x_1$, step sizes $\{\alpha_t\}_{t=1}^T$, batch sizes $\{B_t\}_{t=1}^T$, mini-batch sizes $\{b_t\}_{t=1}^T$, constant $m$
   \FOR{$ t= 1 $ \textbf{to} $T$}
   \STATE Randomly sample a batch $\mathcal{I}_t$ with size $B_t$
   \STATE $g_t = \nabla f_{\mathcal{I}_t}({x}_{t})$
   \STATE $y_1^t = x_{t}$
     \FOR{$ k= 1 $ \textbf{to} $K$}
        \STATE Randomly pick sample $\Tilde{\mathcal{I}}_t$ of size $b_t$ 
        \STATE $v_{k}^t = \nabla f_{\Tilde{\mathcal{I}}_t}(y_{k}^t) - \nabla f_{\Tilde{\mathcal{I}}_t}(y_{1}^t)+ g_t$
        \STATE $y_{k+1}^t = y_{k}^t - \alpha_t v_k^t/(\sqrt{\sum_{\tau=1}^{t-1}\sum_{i=1}^K v_i^{\tau 2} + \sum_{i=1}^k v_i^{t2}}+m)$
     \ENDFOR
   \STATE $x_{t+1} = y^t_{K+1}$
   \ENDFOR
    \STATE \textbf{Return} (Smooth case) Uniformly sample $x_{t^*}$ from $\{y_{k}^t\}_{k=1,t=1}^{K,T}$; (P-L case) $x_{t^*} = x_{T+1}$
    \end{algorithmic}
\end{algorithm}

We provide the implementation of Variance Reduced AdaGrad (VR-AdaGrad) in Algorithm \ref{alg: AdaGrad+VR} when $h(x) = 0$. Note that this implementation is actually a simple combination of the AdaGrad algorithm and the SVRAMD algorithm The implementation can be further extended to the case when $h(x) \neq 0$, but the form would depend on the regularization function $h(x)$. For example, the AdaGrad algorithm with $h(x) = \|x\|_1$, a nonsmooth regularization, can be found in \citet{Duchi2011Adaptive}. The VR-AdaGrad algorithm with $h(x)= \|x\|_1$ will therefore have a similar form. To change the algorithm into VR-RMSProp, one can simply replace the global sum design of the denominator with the exponential moving average in line 9. 

\subsection{Experiment Details}
\textbf{Datasets}. We used three datasets in our experiments. The MNIST \citep{MNIST} dataset has 50k training images and 10k testing images of handwritten digits. The images were normalized before fitting into the neural networks. The CIFAR10 dataset \citep{Cifar} also has 50k training images and 10k testing images of different objects in 10 classes. The images were normalized with respect to each channel (3 channels in total) before fitting into the network. The CIFAR-100 dataset splits the original CIFAR-10 dataset further into 100 classes, each with 500 training images and 100 testing images.

\textbf{Network Architecture}. For the MNIST dataset, we used a one-hidden layer fully connected neural network as the architecture. The hidden layer size was 64 and we used the Relu activation function \citep{Nair2010Rectified}. The logsoftmax activation function was applied to the final output. For CIFAR-10, we used the standard LeNet \citep{LeCun1998Gradient} with two layers of convolutions of size 5. The two layers have 6 and 16 channels respectively. Relu activation and max pooling are applied to the output of each convolutional layer. The output is then applied sequentially to three fully connected layers of size 120, 84 and 10 with Relu activation functions. For CIFAR-100, the ResNet-20 model follows the official implementation in \citet{He2016Deep} by  \citet{Li2020AdaX}.

\textbf{Parameter Tuning}. For the initial step size $\alpha_0$, we have tuned over $\{1, 0.5, 0.1, 0.01, 0.005, 0.002, 0.001\}$ for all the algorithms. For the batch sizes of the original algorithms, we have followed the choices by \citet{Zhou2018Stochastic} for ProxSGD and adaptive algorithms (\citet{Zhou2018Stochastic} used Adam but here we use AdaGrad and RMSProp). For the batch sizes and mini batch sizes $B_t$ and $b_t$ of the variance reduced algorithms, we have tuned over $b_t = \{64, 128, 256, 512, 1024\}$ and $r = \{4, 8, 16, 32, 64\}$. For the constant $m$ added to the denominator matrix $H_t$ in AdaGrad and RMSProp, we choose a reasonable value $m$=1e-3, which is slightly larger than the values set in the original implementations to guarantee strong convexity \citep{Kingma2015Adam}. The other parameters are set to be the default values. For example, the exponential moving average parameter $\beta$ in RMSProp (and VR-RMSProp) is set to be 0.999.  No weight decay is applied to any algorithm in our experiments. The parameters that generate the reported results are provided in Table \ref{tab: parameter_MNIST}, \ref{tab: parameter_CIFAR}, and \ref{tab: parameter_CIFAR100}.

\begin{table}[ht]
\caption{Parameter settings on the MNIST dataset. The batch size $B_t$ is equal to $b_t * r$.}
\vspace{3pt}
\centering
\small
    \begin{tabular}{ l c c c}
\hline
Algorithms & Step size & Mini batch size $b_t$ & Batch size ratio $r$ \\
 \hline 
ProxSGD &0.1 & 1024 & N.A. \\ 
\vspace{2pt}
AdaGrad &0.001 & 2048 & N.A. 
\\ \vspace{2pt}
RMSProp & 0.001& 1024 &N.A.
\\\vspace{2pt}
ProxSVRG+ & 0.1 &  256 & 32\\
\vspace{2pt}
VR-AdaGrad &  0.001 & 256 & 32\\
\vspace{2pt}
VR-RMSProp & 0.001 & 256 & 64 \\
\hline
\end{tabular}
\label{tab: parameter_MNIST}
\end{table}

\begin{table}[ht]
\caption{Parameter settings on the CIFAR-10 dataset. The batch size $B_t$ is equal to $b_t * r$.}
\vspace{3pt}
\centering
\small
    \begin{tabular}{ l c c c}
\hline
Algorithms & Step size & Mini batch size $b_t$ & Batch size ratio $r$ \\
 \hline
ProxSGD &0.1 & 1024 & N.A. \\ 
\vspace{2pt}
AdaGrad &0.001 & 1024 & N.A. \\ 
\vspace{2pt}
RMSProp & 0.001 & 1024 &N.A.\\
\vspace{2pt}
ProxSVRG+ & 0.1 &  512 & 64\\
\vspace{2pt}
VR-AdaGrad &  0.001 & 512 & 64\\
\vspace{2pt}
VR-RMSProp & 0.001 &  512 & 64\\
\hline
\end{tabular}
\label{tab: parameter_CIFAR}
\end{table}

\begin{table}[ht]
\caption{Parameter settings on the CIFAR-100 dataset. The batch size $B_t$ is equal to $b_t * r$.}
\vspace{3pt}
\centering
\small
    \begin{tabular}{ l c c c}
\hline
Algorithms & Step size & Mini batch size $b_t$ & Batch size ratio $r$ \\
 \hline
ProxSGD &0.1 & 1024 & N.A. \\ 
\vspace{2pt}
AdaGrad &0.001 & 1024 & N.A. \\ 
\vspace{2pt}
RMSProp & 0.001 & 1024 &N.A.\\
\vspace{2pt}
ProxSVRG+ & 0.1 &  512 & 64\\
\vspace{2pt}
VR-AdaGrad &  0.001 & 512 & 64\\
\vspace{2pt}
VR-RMSProp & 0.001 &  512 & 64\\
\hline
\end{tabular}
\label{tab: parameter_CIFAR100}
\end{table}

\subsection{Additional Experiments}

\begin{figure}[ht]
     \centering
     \begin{subfigure}[b]{0.4\linewidth}
        \centering
  \includegraphics[width=\linewidth]{ 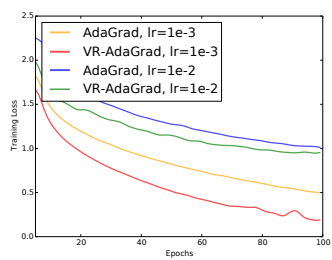}
  \caption{CIFAR10 Training Loss.}
    \label{fig: AdaGrad_LR_train}
     \end{subfigure}
     \hfill
     \begin{subfigure}[b]{0.4\linewidth}
         \centering
   \includegraphics[width=\linewidth]{ 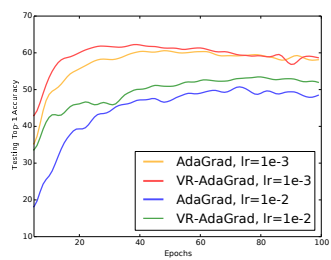}
  \caption{CIFAR10 Testing Acc.}
      \label{fig: AdaGrad_LR_test}
     \end{subfigure}
\caption{Comparison of AdaGrad and VR-AdaGrad on CIFAR-10 using different learning rates. The other parameters are the same as in Table \ref{tab: parameter_CIFAR}. ``lr'' stands for learning rate, which is a different name for step size. The results are averaged over 5 independent runs}
\label{fig: AdaGrad_LR}
\end{figure}

\begin{figure}[ht]
     \centering
     \begin{subfigure}[b]{0.4\linewidth}
        \centering
  \includegraphics[width=\linewidth]{ 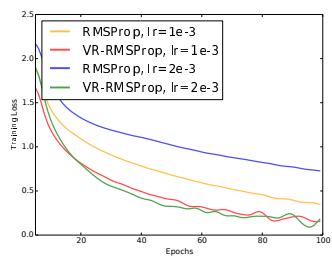}
  \caption{CIFAR10 Training Loss.}
  \label{fig: RMSProp_LR_train}
     \end{subfigure}
     \hfill
     \begin{subfigure}[b]{0.4\linewidth}
         \centering
    \includegraphics[width=\linewidth]{ 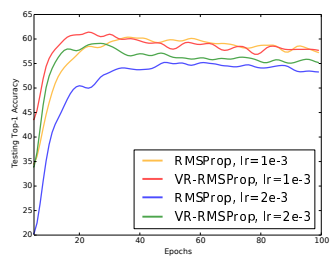}
  \caption{CIFAR10 Testing Acc.}
    \label{fig: RMSProp_LR_test}
     \end{subfigure}
\caption{Comparison of RMSProp and VR-RMSProp on CIFAR-10 using different learning rates. The other parameters are the same as in Table \ref{tab: parameter_CIFAR}. ``lr'' stands for learning rate, which is a different name for step size. lr=1e-2 is too large for RMSProp and the algorithm diverges. The results are averaged over 5 independent runs}
\label{fig: RMSProp_LR}
\end{figure}

\begin{figure}[ht]
     \centering
     \begin{subfigure}[b]{0.32\linewidth}
        \centering
  \includegraphics[width=\linewidth]{ 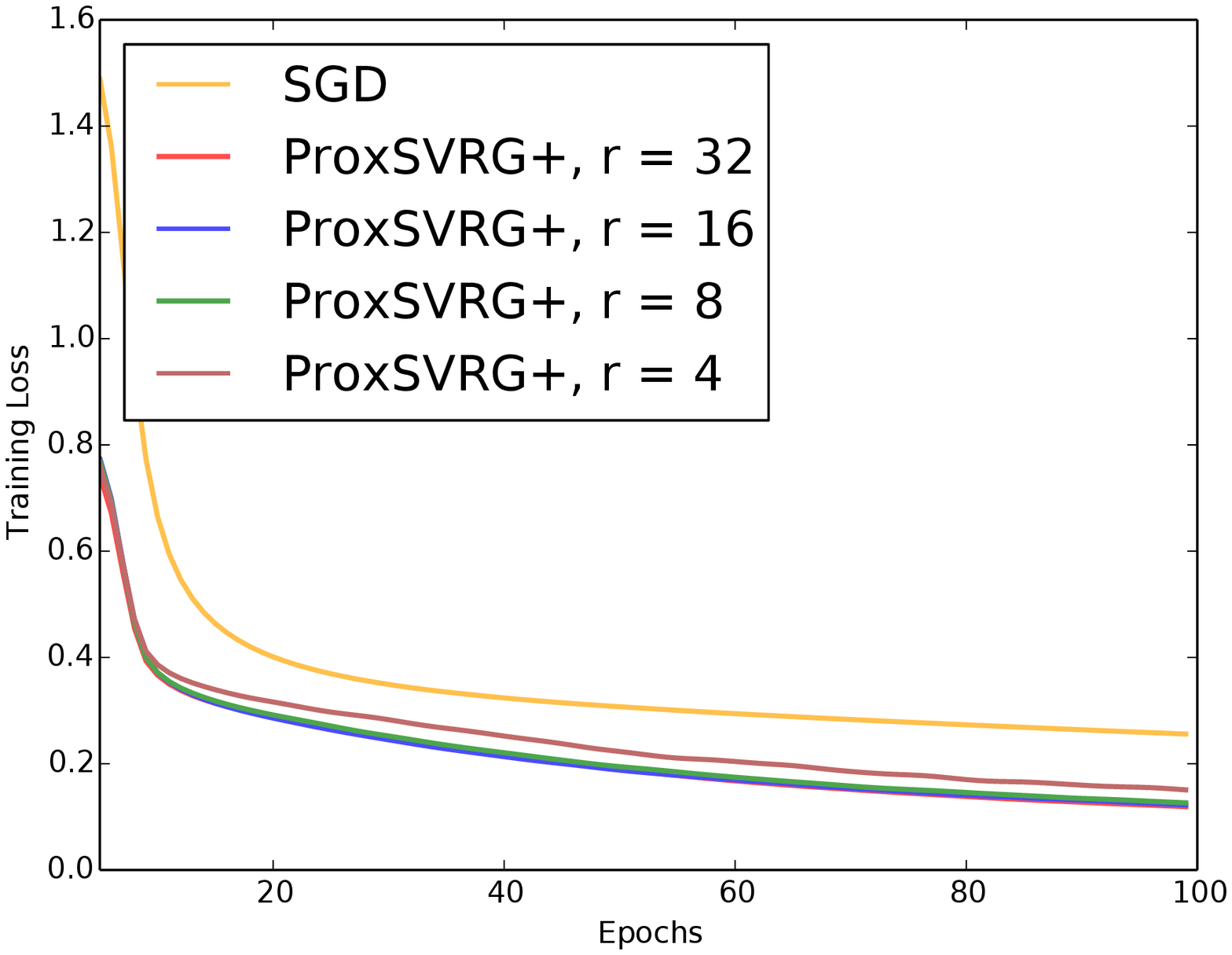}
  \caption{ProxSVRG+ on MNIST}
  \label{fig: SGD_MNIST_batch}
     \end{subfigure}
     \hfill
     \begin{subfigure}[b]{0.32\linewidth}
         \centering
     \includegraphics[width=\linewidth]{ 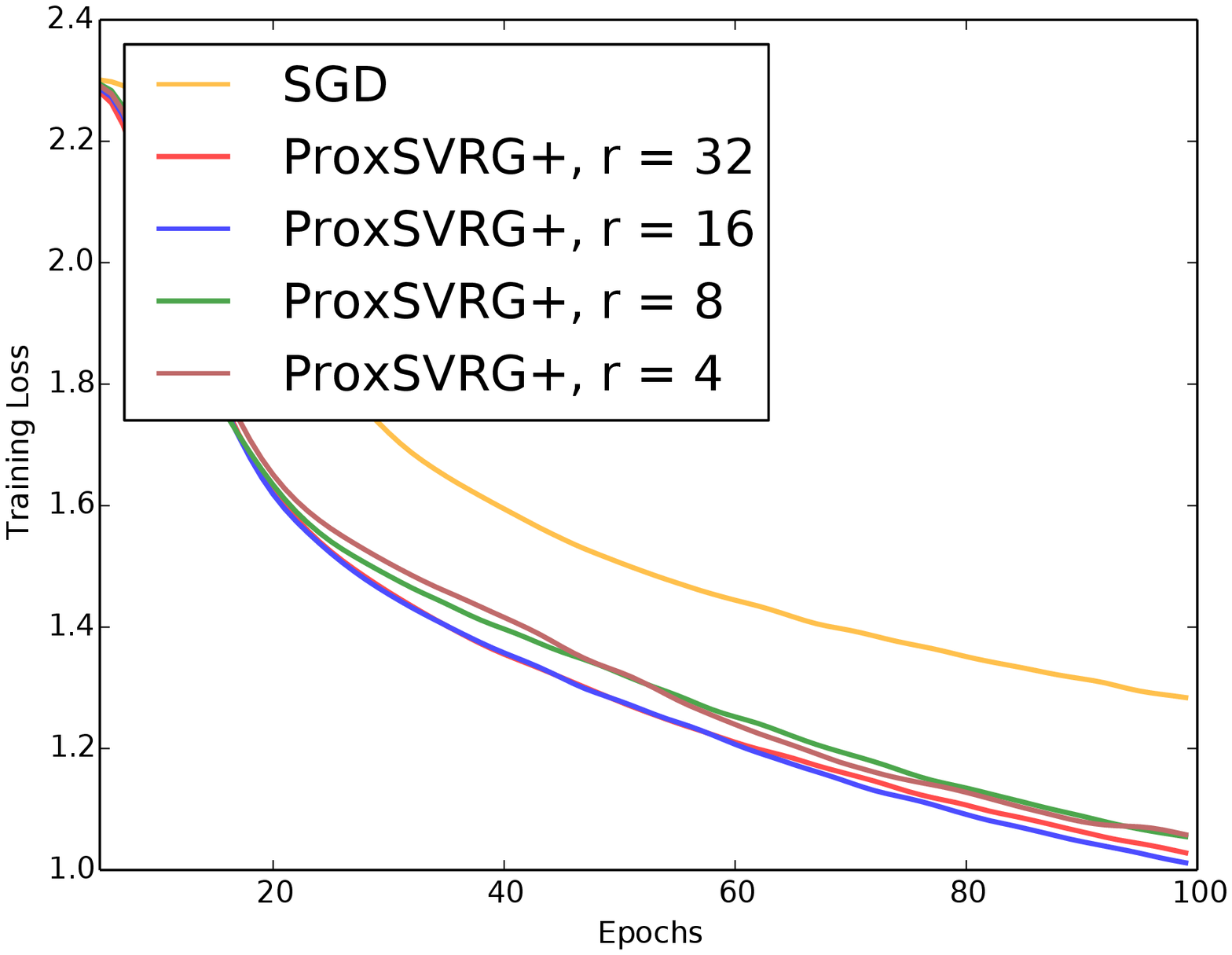}
  \caption{ProxSVRG+ on CIFAR10}
        \label{fig: SGD_CIFAR_batch}
     \end{subfigure}
          \begin{subfigure}[b]{0.32\linewidth}
         \centering
     \includegraphics[width=\linewidth]{ 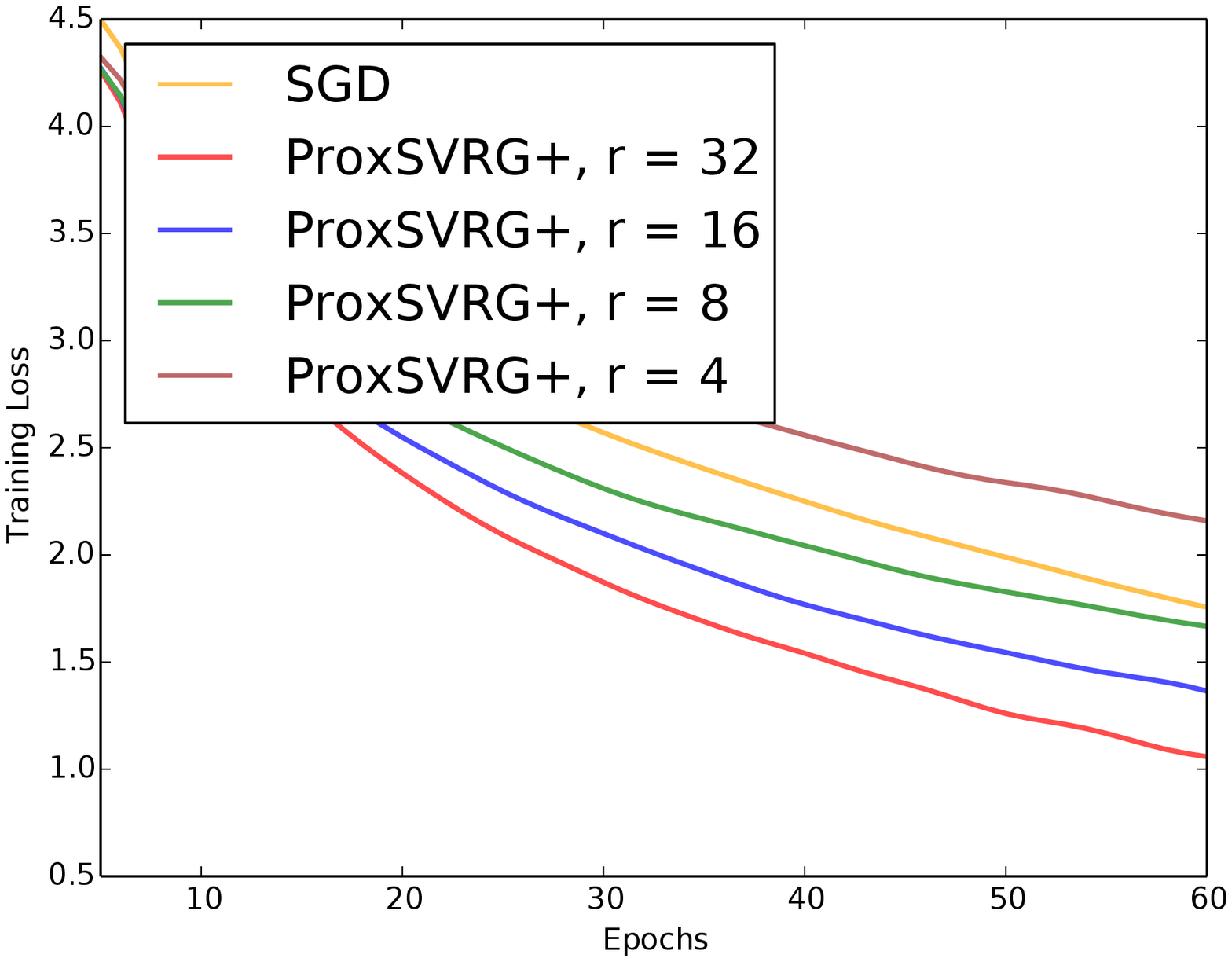}
  \caption{ProxSVRG+ on CIFAR100}
        \label{fig: SGD_CIFAR100_batch}
     \end{subfigure}
\caption{\ref{fig: SGD_MNIST_batch}. training loss of ProxSGD and ProxSVRG+ with different $r$ on MNIST. \ref{fig: SGD_CIFAR_batch}. training loss of ProxSGD and ProxSVRG+ with different $r$ on CIFAR-10.  \ref{fig: SGD_CIFAR100_batch}. training loss of ProxSGD and ProxSVRG+ with different $r$ on CIFAR-100. The other parameters are the same as in Table \ref{tab: parameter_MNIST}, \ref{tab: parameter_CIFAR}. The mini batch size is set to be the same as AdaGrad and RMSProp to ensure fair comparisons. The results were averaged over three independent runs. }
\label{fig: SGD_batch}
\end{figure}

We provide the performances of AdaGrad, RMSProp and their variance reduced variants with different step sizes in figure \ref{fig: AdaGrad_LR} and \ref{fig: RMSProp_LR}. A too-large step size actually makes the convergence of AdaGrad and RMSProp slower. Note that variance reduction always works well in these figures, and it results in faster convergence and better testing accuracy under both step size settings. Therefore for different step sizes, we can apply variance reduction to get faster training and better performances.

The baseline ratios of ProxSVRG+ and ProxSGD in different datasets are provided in Figure \ref{fig: SGD_batch}. Note that ProxSVRG+ only needs a small batch size ratio $(r=4)$ to be faster than ProxSGD.

\end{document}